\documentclass{article}

\usepackage{multirow}
\usepackage{pdflscape}
\usepackage{longtable}
\usepackage{listings}
\usepackage{pifont}
\usepackage{amsthm}
\usepackage{amsmath}
\usepackage{color}
\usepackage[table,usenames,dvipsnames]{xcolor}
\usepackage{paralist}
\usepackage{url}
\usepackage{fancyvrb}
\usepackage{alltt}
\usepackage{graphicx}
\usepackage{rotating}
\usepackage{array}
\usepackage{varwidth}
\usepackage{booktabs}
\usepackage[right]{eurosym}
\usepackage{subcaption}
\usepackage{tikz}
\usetikzlibrary{patterns}
\usetikzlibrary{arrows,shapes,decorations}
\usepackage{framed}
\usepackage[T1]{fontenc}
\usepackage{stmaryrd}

\colorlet{shadecolor}{gray!20}

\usepackage[margin=0.7in]{geometry}

\newcommand{\comments}[1]{}

\theoremstyle{definition}
\newtheorem{mydef}{Definition}{\bfseries}{\normalfont}

\theoremstyle{remark}
{\itshape}{\normalfont}

\newcommand{\HTN}{HTN}
\newcommand{\tHTN}{state-based HTN }
\newcommand{\THTN}{State-based HTN }
\newcommand{\pHTN}{plan-based HTN }
\newcommand{\PHTN}{Plan-based HTN }
\newcommand{\noah}{NOAH}
\newcommand{\nonlin}{Nonlin}
\newcommand{\oplan}{O-Plan2}
\newcommand{\sipe}{SIPE-2}
\newcommand{\umcp}{UMCP}
\newcommand{\shop}{SHOP2}
\newcommand{\siadex}{SIADEX}
\newcommand{\strips}{STRIPS}

\newcolumntype{C}[1]{>{\centering\let\newline\\\arraybackslash\hspace{0pt}}m{#1}}
\newcolumntype{L}[1]{>{\raggedright\let\newline\\\arraybackslash\hspace{0pt}}m{#1}}

\DeclareMathAlphabet{\mathpzc}{OT1}{pzc}{m}{it}

\begin{document}

\title{
{\bfseries \large An Overview of Hierarchical Task Network Planning}
}

\author{{\small Ilche Georgievski and Marco Aiello}\\
\\
{\small \em Distributed Systems Group}\\
{\small \em Johann Bernoulli Institute for Mathematics and Computer Science}\\
{\small \em University of Groningen}\\
\\
{\small e-mail: initial.surname@rug.nl }
\date{{\small March, 2014}}
}

\maketitle

\begin{abstract}
Hierarchies are the most common structure used to understand the world better. In galaxies, for instance, multiple-star systems are organised in a hierarchical system. Then, governmental and company organisations are structured using a hierarchy, while the Internet, which is used on a daily basis, has a space of domain names arranged hierarchically. Since Artificial Intelligence (AI) planning portrays information about the world and reasons to solve some of world's problems, Hierarchical Task Network (\HTN) planning has been introduced almost 40 years ago to represent and deal with hierarchies. Its requirement for rich domain knowledge to characterise the world enables \HTN~planning to be very useful, but also to perform well. However, the history of almost 40 years obfuscates the current understanding of \HTN~planning in terms of accomplishments, planning models, similarities and differences among hierarchical planners, and its current and objective image. On top of these issues, attention attracts the ability of hierarchical planning to truly cope with the requirements of applications from the real world. We propose a framework-based approach to remedy this situation. First, we provide a basis for defining different formal models of hierarchical planning, and define two models that comprise a large portion of \HTN~planners. Second, we provide a set of concepts that helps to interpret \HTN~planners from the aspect of their search space. Then, we analyse and compare the planners based on a variety of properties organised in five segments, namely domain authoring, expressiveness, competence, performance and applicability. Furthermore, we select Web service composition as a real-world and current application, and classify and compare the approaches that employ \HTN~planning to solve the problem of service composition. Finally, we conclude with our findings and present directions for future work.
\end{abstract}


\tableofcontents

\section{Introduction}\label{sec:intro}
Hierarchical Task Network (\HTN) planning is an Artificial Intelligence (AI) planning technique that breaks with the tradition of classical planning~\cite{nau2004:automated}. The basic idea behind this technique includes an initial state description, an initial task network as an objective to be achieved, and domain knowledge consisting of networks of primitive and compound tasks. A task network represents a hierarchy of tasks each of which can be executed, if the task is primitive, or decomposed into refined subtasks. The planning process starts by decomposing the initial task network and continues until all compound tasks are decomposed, that is, a solution is found. The solution is a plan which equals to a set of primitive tasks applicable to the initial world state. 

Beside being a tradition breaker, \HTN~planning appears to be controversial as well. The controversy lies in its requirement for well-conceived and well-structured domain knowledge. Such knowledge is likely to contain rich information and guidance on how to solve a planning problem, thus encoding more of the solution than it was envisioned for classical planning techniques. This structured and rich knowledge gives a primary advantage to \HTN~planners in terms of speed and scalability when applied to real-world problems and compared to their counterparts in classical world. 

The biggest contribution towards this kind of ``popular'' image of \HTN~planning has emerged after the proposal of the Simple Hierarchical Ordered Planner (SHOP)~\cite{nau1999:shop} and its successors. SHOP is an \HTN-based planner that shows efficient performance even on complex problems, but at the expense of providing well-written and possibly algorithmic-like domain knowledge. Several situations may confirm our observation, but the most well-known is the disqualification of SHOP from the International Planning Competition in 2000~\cite{bacchus2001:aips} with the motivation that the domain knowledge was not well written so that the planner produced plans that were not solutions to the competition problems~\cite{nau1999:shop}. Furthermore, the disqualification was followed by a dispute on whether providing such knowledge to a planner should be considered as cheating in the world of AI planning~\cite{nau2007:trends}.

SHOP's style of \HTN~planning was introduced by the end of 1990s, but \HTN~planning existed long before that. The initial idea of hierarchical planning was presented by the Nets of Action Hierarchies (\noah) planner~\cite{sacerdoti1975:structure} in 1975. It was followed by a series of studies on practical implementations and theoretical contributions on \HTN~planning up until today. We believe that the fruitful ideas and scientific contribution of nearly 40 years must not be easily reduced to controversy and antagonism towards \HTN~planning. On the other hand, we are faced with a situation full of fuzziness in terms of difficulty to understand what kind of planning style other \HTN~planners perform, how it is achieved and implemented, what are the similarities and differences among these planners, and finally, what is their actual contribution to the creation of the overall and possibly objective image of \HTN~planning. The situation cannot be effortlessly clarified because the current literature reports little or nothing at all on any of these issues, especially in a consolidated form. 

In addition to these issues, we observe the applicability of AI planning techniques as an ultimate goal of their development. We are especially interested in novel and real-world domains which may require reconsidering established techniques. The growing trend on other than classical and synthetic domains leads to the need for algorithms and systems that reflect planning better and more closely to the real world. This perspective gives another view to the abilities of \HTN~planners (and \HTN~planning in general) to cope with various properties of an application in the real world.

We aim to consolidate and synthesise a number of existing studies on \HTN~planning in a manner that will clarify, categorise and analyse \HTN~planners, and allow us to make statements that are not merely based on contributions of a single \HTN~planner. We also hope to rectify the perception of \HTN~planning as being controversial and antagonistic in the AI planning community. Finally, we choose a non-traditional, dynamic and uncertain application domain to ascertain \HTN~planning with respect to various domain characteristics.

\subsection{Approach}
\begin{figure} [!ht]
\centering
\includegraphics[width=.3\textwidth]{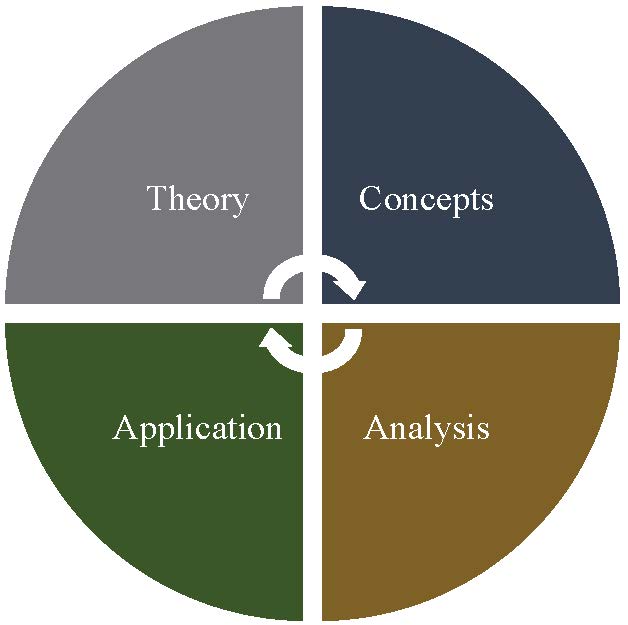}
\caption{Pie of Frameworks}\label{fig:pie}
\end{figure}

We take a framework-based approach to accomplish our objectives. We perceive a framework as an abstract and logical structure that we establish in need for support and guidance on the development of our study. Since we inspect \HTN~planning from four different perspectives, we provide a framework for each perspective. All four frameworks form the design depicted in Figure~\ref{fig:pie} to which we refer to as a {\em pie of frameworks}. The pie of frameworks serves as a central and unifying point of our study organisation and presentation flow. The first piece of the pie is a {\em theoretical framework} for \HTN~planning upon which we later build two models of \HTN~planning. The second slice is a {\em conceptual framework} for \HTN~planning that provides clarification of different concepts related to the search space, and context for interpretation of \HTN~planners. The next piece is an {\em analytical framework} for \HTN~planners that enables us to go deeper and beyond dry descriptions about \HTN~planners. The last piece might appear to taste differently than the other three, but it still has the flavour of \HTN~planning. The {\em application framework} concerns the application domain we choose to observe, and it helps us to analyse different studies in an organised and unified way, and possibly to identify points where \HTN~planning behaves as expected or can be further improved.

\subsection{Inclusion of \HTN~Planners and Studies}
We make use of two inclusion criteria for planners and studies. The {\em inclusion criterion of planners} relies on the inspection of existing literature for suggestions on \HTN~planners that have risen to some degree of prominence. For example, we accept the list of ``best-known domain-independent \HTN~planning systems'' as provided in~\cite{nau2004:automated}. In addition to those five suggested planners, we include two more. The complete list of \HTN~planners participants in our study is the following one:\footnote{Later in the study, we refer only to the most recent version of each planner.}

\begin{itemize}
\item Nets Of Action Hierarchies (\noah), the first \HTN~planner emerged in mid-1970s~\cite{sacerdoti1975:nonlinear, sacerdoti1975:structure},
\item \nonlin~that appeared one year later~\cite{tate1976:project, tate1977:generating}, 
\item System for Interactive Planning and Execution (SIPE) and its successor \sipe~introduced in 1984 and 1990, respectively~\cite{wilkins1991:practical},
\item Open Planning Architecture (O-Plan) and its successor \oplan~presented in 1984 and 1989, respectively~\cite{currie1991:oplan,tate1994:oplan2},
\item Universal Method Composition Planner (\umcp) introduced in 1994~\cite{erol1996:formalization}, 
\item Simple Hierarchical Ordered Planner (SHOP) and its successor \shop~that appeared in 1999 and 2003, respectively~\cite{nau1999:shop,nau2003:shop2}, and
\item \siadex~that emerged in 2005~\cite{castillo2005:temporal_enhancements}. 
\end{itemize}

The {\em inclusion criterion of studies} relies on the theoretical contribution of a study with respect to \HTN~planning in general, and theoretical and practical issues of each chosen planner separately. The criterion is based on the coverage a study gives, which may include information that ranges from a general discussion of techniques and approaches, peculiar matters, such as task interactions and condition types, relevant to our conceptual framework, to properties, such as domain authoring, expressiveness and competence, that may be a part of the analytical framework. Finally, we include an extensive number of studies that employ \HTN~planning for the purpose of our application domain, that is, Web service composition.

\subsection{Web Service Composition}
We choose the domain of Web services as a non-traditional and real-world application~\cite{papazoglou2003:soc}. {\em Web services} are software components that implement specific business logic, and are distributed over a network, typically the Web, to be used as Web resources for machine-to-machine interaction. For instance, travel agencies may provide a number of Web services, such as booking a flight ticket, reserving a hotel, renting a car, or organising sightseeing. The interaction is usually initiated by a client request which has to be satisfied by the functionalities that Web services offer. However, in cases when no single service can accomplish the request, a combination of several Web services might give a value-added functionality, and provide a way to request satisfaction. For example, a service to arrange a complete trip to some tourist destination might be of an exceptional use to the commercial travel agencies, and thus, it will not be offered as a Web service. {\em Web service composition} (WSC), especially when accomplished automatically, appears to be in-line with the objective of AI planning~\cite{lazovik2004:assertions, dustdar2005:surveywsc, kaldeli2013:dynamicuncertain}. That is, planning operators correspond to functionalities of Web services, while the goal, in the simplest example, is aggregated from the user request. In addition, the environment of Web services already shows some propensity to composite or hierarchical representation. For example, a composite service could describe a reservation of hotel by searching for a hotel, registering to the hotel, logging to the hotel and, finally, the actual booking of the hotel. Among AI planning techniques, \HTN~planning is well-suitable for domains in which some hierarchical representation is desirable or known in advance, domains that encourage complex and composed constructs, and domains of large size. These indicators suggest ideal conceptual matchmaking between \HTN~planning and Web service composition, but are also a computational challenge. Thus, continuing with our example, if the user objective is not only to reserve a hotel, but to arrange a complete trip, which includes also booking a flight, renting a car, and sightseeing, then, definitely, the complexity of services and their composition becomes an interesting and challenging task.

The environment of Web services offers more exciting challenges that make the effective selection and composition of services far from being plain and straightforward planning processes. In particular, Web services exist in a {\em dynamic} environment in which the availability of services is not guaranteed. This behaviour reflects the availability of information which, on the other hand, is assumed by planners to be complete and obtainable before the planning process is initiated. Furthermore, the environment of Web services favours techniques that are able to deal with {\em uncertainty} in terms of 1) incomplete information about the initial state; 2) uncertainty over the many possibilities for completion of missing information by invoking some sensing services at planning and/or execution time; 3) non-determinism caused by failed invocations of Web services (e.g., renting a car is not viable at the moment of invocation), a service not responding at all, a service yielding undesired outcome (e.g., booking a flight provides only business-class tickets); 4) services that show unexpected behaviour (e.g., Byzantine failure). Moreover, {\em complex goals} possibly in the form of a workflow or conditioned with some organisational regulations or augmented with user preferences are the norm rather than exception. Finally, the {\em high cardinality} of the set of Web services available on the Web implies a large space to be searched by a planner. 

\subsection{Running Example: Logistics Domain}\label{sec:exmpl}

\begin{figure} 
\centering
\includegraphics[width=.8\textwidth]{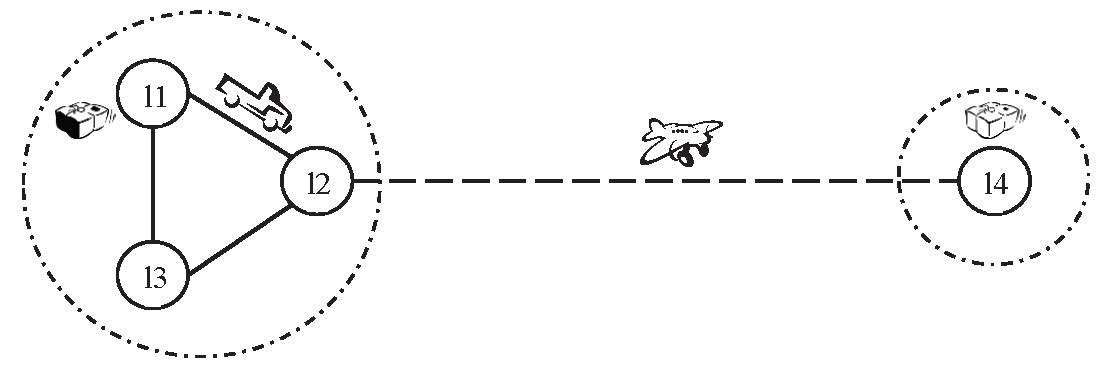}
\caption{An Example in the Logistics Domain}\label{fig:running-example}
\end{figure}

In order to illustrate some of the concepts and definitions discussed throughout the paper, we take as an example the logistics planning domain. Figure~\ref{fig:running-example} illustrates a scenario using the domain. Suppose a setting in the logistic world that contains several locations, such as \texttt{l1}, \texttt{l2} and \texttt{l3}, which can be adjacent (the solid line) and in a same city (the dash-dotted circle) or in different cities (the dashed line), a box \texttt{b} that should be delivered, and a truck \texttt{t} and a plane \texttt{p}, which are delivery vehicles for a box transported in same-city locations or different-city locations, respectively. The goal is to deliver a box from an initial location to a desired destination. For example, a goal could say to deliver a box from \texttt{l1} to \texttt{l4}. The box can be transported by a truck by loading the box into the truck at \texttt{l1} (denoted as \texttt{load-truck(t,b,l1)}), driving the truck to \texttt{l2} (denoted as \texttt{drive(t,l1,l2)}), and unloading the truck at \texttt{l2} (denoted as \texttt{unload-truck(t,b,l2)}). Finally, loading the box into a plane, flying the plane to \texttt{l4}, and unloading the plane at \texttt{l4} are denoted analogously to their counterparts.

\subsection{Organisation}
The remainder of the paper is organised as follows. Section~\ref{sec:theory-concepts} describes the theoretical and conceptual frameworks. Based on the ideas presented in these frameworks, two models of \HTN~planning are proposed and formalised. The planners corresponding to an \HTN~model are reviewed with respect to the conceptual framework. Section~\ref{sec:analysis} provides details on each \HTN~planner separately. Section~\ref{sec:wsc} goes deeper into the application area of Web service composition accomplished by \HTN~planning. Finally, Section~\ref{sec:conclusions} concludes the paper with some considerations and directions for future work.

\section{HTN Planning: Theory and Concepts}\label{sec:theory-concepts}

\HTN~planning has been formalised in several studies, such as~\cite{erol1994:umcp,nau1999:shop,nau2004:automated,geier2011:decidability}. These formalisations include similar definitions of \HTN~terms, appropriate to the needs for their model of \HTN~planning. Based on these existing theories, we describe the first piece of our framework pie, that is, the theoretical framework for \HTN~planning. In this framework, we keep the definitions of \HTN~terms high level. Further in the paper, we provide specific definitions of the terms that are characteristic for the model of \HTN~planning being analysed. The purpose of the theoretical framework is twofold. Firstly, it provides basic understanding of \HTN~planning. Secondly, it determines and defines the focus of categorisation of \HTN~planning that we propose.

\subsection{Theoretical Framework}\label{sec:tf}
The theoretical framework is composed of a planning language, tasks, operators, task networks, methods, planning problem and solution. The {\em \HTN~planning language} is a first-order language that contains several mutually disjoint sets of symbols. As usual, a {\em predicate}, which evaluates to true or false, consists of a predicate symbol $p \in P$, where $P$ is a finite set of predicate symbols, and a list of terms $\tau_{1},\dots,\tau_{k}$. A {\em term} is either a constant symbol $c \in C$, where $C$ is a finite set of constant symbols, or a variable symbol $v \in V$, where $V$ is an infinite set of variable symbols. We denote the set of predicates as $Q$. A predicate is {\em ground} if its terms contain no variable symbols. A {\em state} $s \in 2^{Q}$ is a set of ground predicates in which the {\em closed-world assumption} is adopted, that is, all and only the predicates that are true are specified in the state. We define a {\em primitive task} as an expression $t_{p}(\tau)$, where $t_{p} \in T_{p}$ and $T_{p}$ is a finite set of primitive-task symbols, and $\tau = \tau_{1},\dots,\tau_{k}$ are terms. A primitive task is represented by a planning operator.

\begin{mydef}[Operator]\label{def:o}
An {\em operator} $o$ is a triple $(p(o),pre(o),eff(o))$, where $p(o)$ is a primitive task, and $pre(o) \in 2^{Q}$ and $eff(o) \in 2^{Q}$ are preconditions and effects, respectively. The subsets $pre^{+}(o)$ and $pre^{-}(o)$ denote positive and negative preconditions of $o$, respectively.
\end{mydef}

A transition from one state to another is accomplished by an instance of an operator whose precondition is a logical consequence of the current state. An operator $o$ is {\em applicable} in state $s$, if $pre^{+}(o) \subseteq s$ and $pre^{-}(o) \cap s = \emptyset$. Applying $o$ to $s$ results in the state $s[o] = (s\setminus eff^{-}(o)) \cup eff^{+}(o)$,
where $eff^{-}(o)$ and $eff^{+}(o)$ are negative and positive effect of $o$, respectively. Notations $s[o]=s'$ and $s\underset{o}{\longrightarrow}s'$ are equivalent and we use them interchangeably.

A \textit{compound task} is an expression $t_{c}(\tau)$, where $t_{c} \in T_{c}$ and $T_{c}$ is a finite set of compound-task symbols, and $\tau = \tau_{1},\dots,\tau_{k}$ are terms. We refer to the union of the sets of primitive-task and compound-task symbols as a set of task names $T_{n}$. The following two definitions are further complemented for the respective model of \HTN~planning.

\begin{mydef}[Task network]\label{def:tn}
A \textit{task network} $tn$ is a pair $(T,\psi)$, where $T$ is a finite set of tasks, and $\psi$ is a set of constraints.
\end{mydef}

Constraints in $\psi$ specify restrictions over $T$ that must be satisfied during the planning process and by the solution. We refer to a task network over $T_{p}$ as a \textit{primitive task network}. The set of all task networks over $T_n$ is denoted as $TN$.

\begin{mydef}[Method]\label{def:m}
A {\em method} $m$ is a pair $(c(m),tn(m))$, where $c(m)$ is a compound task, and $tn(m)$ is a task network.
\end{mydef}

\begin{mydef}[Planning problem]\ignorespaces
A {\em planning problem} $\mathcal{P}$ is a tuple $(Q,T_{p},T_{c},O,M,tn_{0},s_{0})$, where
\begin{itemize}
\item $Q$ is a finite set of predicates
\item $T_{p}$ is a finite set of primitive task symbols
\item $T_{c}$ is a finite set of compound task symbols
\item $O \subseteq T_{p} \times 2^{Q} \times 2^{Q}$ is a finite set of operators
\item $M \subseteq T_{c} \times TN$ is a finite set of methods
\item $tn_{0}$ is the initial task network
\item $s_{0}$ is the initial state.
\end{itemize}
\end{mydef}

An operator sequence $o_{1},\dots,o_{n}$ is {\em executable} in $s$, if there are states $s_{0},\dots,s_{n}$ such that $s_{0}=s$ and $o_{i}$ is applicable in $s_{i-1}$ and $s_{i-1}[o_{i}]=s_{i}$ for all $a \leq i \leq n$. Given a problem $\mathcal{P}$, a {\em solution} to $\mathcal{P}$ is an operator sequence executable in $s_{0}$ by decomposing $tn_{0}$. The way of producing such sequence is defined in the following sections.

\subsection{Conceptual Framework}\label{sec:cf}
Literature reports vague information on \HTN~planning (and planners) especially in the early stages of hierarchical planning. First, it is difficult to understand the ideas and concepts used and how they are adapted for the purpose of \HTN~planning. The situation later improved with the evolution of \HTN~planners and the attempts at formalisation. Then, different models of \HTN~planning could be found at some point of the evolution, and at first glance, the model distinction seems not that obvious and comprehensible.

\begin{figure} [!ht]
\centering
\includegraphics[width=.7\textwidth]{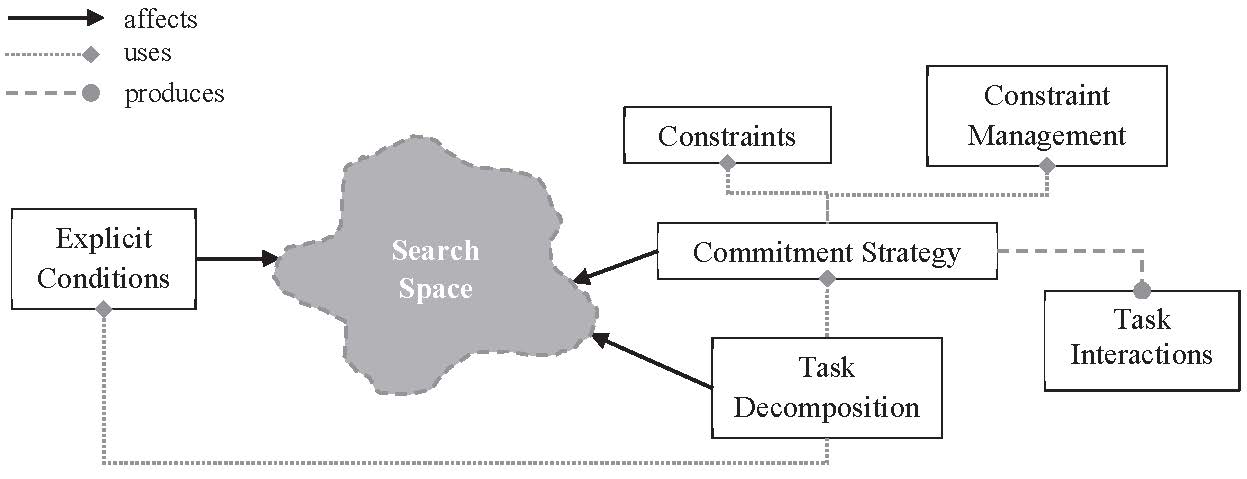}
\caption{Conceptual Framework}\label{fig:concepts}
\end{figure}

The motivation for the second slice of our framework pie lies exactly in these issues. We clarify them by designing and describing a conceptual framework shown in Figure~\ref{fig:concepts}. This framework is less formal (compared with the theoretical framework) and based on specific concepts derived from empirical observation. We start by providing basic and general enough descriptions of concepts that characterise different \HTN~planners and cover most of their important features. The concepts are placed within a logical and sequential design as much as possible. In our framework, the key concept is the search space to which other concepts are related and interconnected in various ways. The purpose of the conceptual framework is manifold. Firstly, it clarifies concepts and proposes relationships among them. Secondly, it provides context for interpreting the findings presented in the paper, and helps in explaining observations. Finally, the view of key concepts enables us to categorise \HTN~planning based on the space the search is performed in.

\subsubsection{Task Decomposition}\label{sec:td}
Given some task network, a {\em task decomposition} chooses a task from the task network and, if it is primitive and suitable for the current state, the task decomposition applies it to the state. Otherwise, all the methods are analysed that contain the chosen task as a part of their definition. Assuming that a set of methods is found, a non-deterministic choice of a method is made, and the task is replaced with the task network associated with the chosen method. Finally, the newly composed task network is checked against any constraint-related violation and modified, if necessary.

Task decompositions can be divided into three styles based on the representation of task networks in terms of task ordering, and the way of forming new task networks during decomposition. The first one is {\em totally ordered task decomposition} (TOTD). It follows the assumption of total order on task networks so as when a task is decomposed, the new task network is created in such a way that newly added tasks are totally ordered among each other and with respect to the tasks of the existing task network. Sometimes we refer to the \HTN~planning that uses this style as totally ordered \HTN~planning. The second style is {\em unordered task decomposition} (UTD) that relaxes the requirement of totally ordered task networks. That is, tasks can be totally ordered or unordered with respect to each other (but no tasks in parallel are allowed). When a task is decomposed, new task networks are created in such a way that newly added tasks are interleaved with the tasks of the existing task network until all permissible permutations are exhausted. Here as well, we refer to the \HTN~planning that embodies this style as unordered \HTN~planning. The last style is {\em partially ordered task decomposition} (POTD) that allows the existence of a partial order on tasks. When a task is decomposed, the tasks in the newly created network can be ordered in parallel whenever possible (with respect to the constraints). The \HTN~planning that uses this style is referenced as partially ordered \HTN~planning.

\subsubsection{Constraints}\label{sec:constraints}
Definition~\ref{def:tn} suggests that constraints are found in a task network, but constraints can be also added during the planning process in order to resolve inconsistencies. \HTN~planners deal with several types of constraints, and most of them can be interpreted as in~\cite{stefik1981:constraints}. Briefly, there are three interpretations. First, we meet a constraint that implies commitments about partial descriptions of state objects. Another type of constraint refines variable bindings if a certain variable binding does not satisfy some condition. Last, there is a constraint that expresses the relations between variables in different parts of a task network.

\paragraph{Commitment Strategy}
As with most of the other AI planners, \HTN~planners also need to make two decisions on constraints. The first one is on constraints for bindings variables, while the second decision is on constraints for ordering tasks in a task network. \HTN~planners use mainly two strategies for when and how to make these decisions. The first strategy manages constraints in compliance with the {\em least-commitment strategy} so that task ordering and variable bindings are deferred until a decision is forced~\cite{weld1994:least_commitment, tsuneto1996:commitment}. The second strategy handles constraints according to the {\em early-commitment strategy} so that variables are bound and operators in the plan are totally ordered at each step of the planning process. Planners employing this strategy greatly benefit from the possibility of adopting forward chaining where chaining of actions is achieved by imposing a total order on (some) plan actions. The total ordering ensures that neither the current action to be added to the plan can interfere with some earlier action's preconditions or effects, nor a later action can interfere with current action's preconditions or effects. 

\begin{figure} 
\centering
	\begin{subfigure}[b]{\textwidth}
		\includegraphics[width=\textwidth]{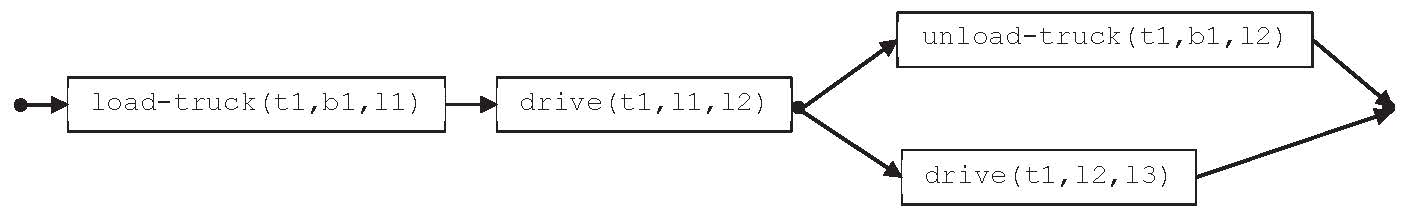}
		\caption{Deleted-Conditions Interaction}
		\label{fig:deleted-condition-example}
	\end{subfigure}
	~
	\begin{subfigure}[b]{\textwidth}
		\includegraphics[width=0.7\textwidth]{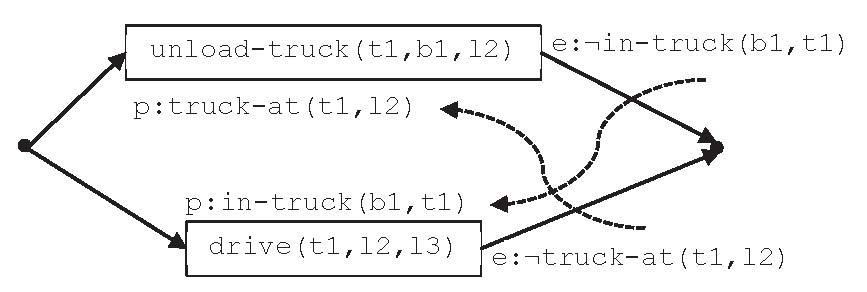}
		\caption{Double-Cross Interaction}
		\label{fig:double-cross-example}
	\end{subfigure}
	~ 
\caption{Examples of Deleted-Condition and Double-Cross Interactions}\label{fig:interaction-examples}
\end{figure}

\paragraph{Task Interaction}
Depending on the commitment strategy chosen, and especially in the case of the least-commitment strategy, an inevitable consequence is the interaction among tasks in a given task network. Generally, an {\em interaction} is a connection between two tasks (or parts) of a task network in which these tasks (or parts) have some effect on each other. Based on this effect, we divide interactions into two categories. The first category, called {\em harmful interactions} (also threats or flaws), introduces conflicts among different parts of a task network that threaten its correctness. \HTN~planners consider harmful interactions individually, and provide rather intuitive descriptions. In the following list, we provide a general description of harmful interactions found in \HTN~planners.
\begin{itemize}
\item {\em Deleted-condition interaction} - appears when a primitive task in one part of a task network deletes an expression that is a precondition to a primitive task in another part of that task network. For example, consider the following situation from our running-example domain. The state contains three possible locations \texttt{l1}, \texttt{l2} and \texttt{l3}, one truck \texttt{t1}, and one box \texttt{b} to be transported. The idea is to move \texttt{b1} from \texttt{l1} to \texttt{l2} and to move \texttt{t1} from \texttt{l2} to \texttt{l3}. Then, the task network depicted in Figure~\ref{fig:deleted-condition-example} has a deleted-condition interaction. The conflict arises when the action \texttt{drive(t1,l2,l3)} deletes the fact \texttt{truck-at(t1,l2)}, which is added by \texttt{drive(t1,l1,l2)} and is a precondition of \texttt{unload-truck(t1,b1,l2)}.
\item {\em Double-cross interaction} - appears when an effect of each of two conjunctive primitive tasks deletes a precondition for the other. That is, an effect of the first task deletes a precondition of the second primitive task, and an effect of the second task deletes a precondition of the first task. For example, consider the situation shown in Figure~\ref{fig:double-cross-example}, where a truck \texttt{t1} is at location \texttt{l2} and a box \texttt{b1} is loaded into the \texttt{t1}. The idea is to unload the truck at \texttt{l2} and move the truck to \texttt{l3}, and in addition, for the purpose of illustration, consider an expanded description of the \texttt{drive(t,b,l)} operator that requires a box to be in the truck while moving. Then, tasks are in a double-cross interaction, that is, \texttt{unload-truck(t1,b1,l2)} deletes \texttt{in-truck(b1,t1)}, which is a precondition for \texttt{drive(t1,l2,l3)} in order to move to \texttt{l3}, and \texttt{drive(t1,l2,l3)} deletes \texttt{truck-at(t1,l2)}, which is a precondition for the execution of \texttt{unload-truck(t1,b1,l2)}.
\item {\em Resource interaction} - appears in two situations, and it is subdivided accordingly.  A \textit{resource-resource interaction} is similar to the deleted-condition interaction, while a \textit{resource-argument interaction} occurs when a resource in one part of a task network is used as an argument in another part of that task network.
\end{itemize}

The second category, called {\em helpful interactions}, refers to situations when one part of a task network can make use of information associated with another part in the same task network. The detection of these interactions implies the possibility for a planner to generate better-quality task networks and solutions. In fact, some tasks can be merged together, which eliminates task  redundancy and potentially optimises the cost of the solution~\cite{foulser1992:merging}. The following list contains general descriptions of helpful interactions. 

\begin{itemize}
\item {\em Placeholder replacement} - appears when a real value already exists for a particular formal object. \HTN~planners allow tasks with variables to be inserted into a task network. If there is no specific value to be chosen for a particular variable choice, a so-called formal object is created to bind the variable~\cite{sacerdoti1975:nonlinear}. The formal object is simply a placeholder for some entity unspecified at that point.
\item {\em Phantomisation} - appears when some goal is already true at the point in a task network where it occurs. In the descriptions of some \HTN~planners, the term `goal' is interchangeably used with the term `precondition'. In fact, if some task precondition is not satisfied, it is inserted as a goal to be achieved.
\item {\em Disjunct optimisation} - appears in disjunctive goals when one disjunctive goal is ``superior to the others by the nature of its interaction'' with the other tasks in a task network~\cite{sacerdoti1975:nonlinear}.
\end{itemize}

\paragraph{Constraint Management}
Task interactions can be solved by posting various types of constraints onto a task network. This {\em constraint posting} is also known as conflict resolution~\cite{yang1992:conflict} or critics~\cite{tate1976:project,wilkins1988:extending}. Similarly as for the representation, \HTN~planners do not provide a general approach for handling interactions, thus each of the above interactions has its own resolution method. However, those methods are based on well-known operations on constraints generally described elsewhere, e.g.,~\cite{stefik1981:constraints}. We briefly describe some operations in the context of \HTN~planning. 

\begin{figure} 
		\centering
		\includegraphics[width=\textwidth]{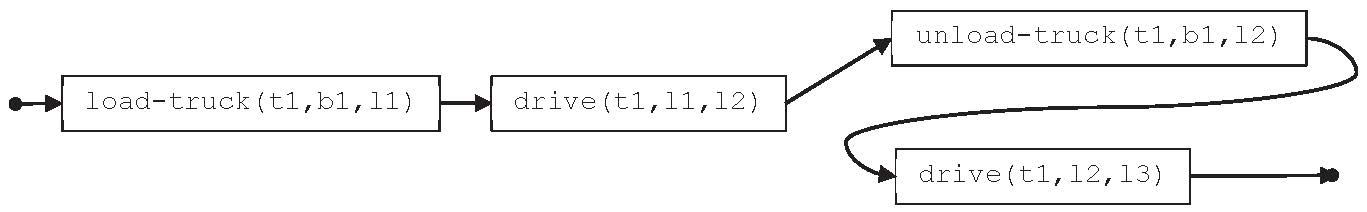}
		\caption{A Conflict-Free Portion of a Task Network as a solution to the interaction in Figure~\ref{fig:deleted-condition-example}}
		\label{fig:deleted-conditions-resolution}
\end{figure}

The most basic operation is {\em constraint satisfaction} which happens when an \HTN~planner searches for a variable binding that satisfies the given constraints, and guarantees the consistency of, for example, a set of ordering constraints over a task network. {\em Constraint propagation} enables adding or retracting constraints to and from a task network. Variable constraints in one part of a task network can be propagated based on variable constraints in another part of that task network. With respect to ordering constraints, propagation is used when the linking process is performed. When some task interferes with another task, the {\em linking process} records a causal link, that is, a three-element structure of two pointers to tasks $t_{e}$ and $t_{p}$, and a predicate $q$ which is both an effect of $t_{e}$ and a precondition of $t_{p}$. For example, the deleted-condition interaction from Figure~\ref{fig:deleted-condition-example} is resolved by adding a constraint to execute action \texttt{unload-truck(t1,b1,l2)} before \texttt{drive(t1,l2,l3)}, which results in a conflict-free task network as shown in Figure~\ref{fig:deleted-conditions-resolution}. 
The solution to the phantomisation interaction is practically a linking process. Phantomisation of a task $t$ with an effect $e$ is accomplished by treating $e$ as achieved, and finding an existing task $t'$ in the task network that achieves the same effect $e$. If task $t'$ is found, a constraint $(t',e,t)$ is added to the task network to record the causal relation.

The last operation is different in that it does not happen during the planning process. {\em Constraint formulation} can be taken into account when modelling \HTN~domain knowledge, especially when the domain author is aware in advance of some possible impasse situations. By posting constraints as control information into the domain knowledge, the planner can gain on efficiency by refining the search space~\cite{erol1994:semanticshtn, nareyek2005:constraints}. In some \HTN~planners, the phantomisation of a task is achieved by an explicit encoding in the domain knowledge. The planners handle the phantomisation of a rather recursive task by taking into account an alternative method which explicitly encodes a `do-nothing' operation. Therefore, one could say that the planners lack the ability to infer such situations. This issue is addressed in~\cite{georgievski2011:phantomisation}, where a support for reasoning about the conditions that are already achieved without the need of explicit specification into the domain knowledge is provided.

\subsubsection{Explicit Conditions}\label{sec:ec}
\HTN~planners depend on the quality of the domain knowledge to restrict and guide the search. The domain author is the one who has the responsibility of giving the information about the guidance in the search space. One way to represent such information is by using explicit conditions. We provide a general description of conditions found in \HTN~planners.

\begin{itemize}
\item {\em Supervised condition} - accomplished within a compound task. The condition may be satisfied either by an intentional insertion of a relevant effect earlier in the task network, or by an explicit introduction of a primitive task that will achieve the required effect. Generally, only this condition should allow further decompositions to be made and, since it may be included for the achievement of the condition invocation of another task, this condition corresponds to preconditions in \strips-like planning systems.\footnote{With the ``\strips-like`` term, we refer to planning languages that use representations similar to the \strips~one~\cite{fikes1971:strips}, that is, an operator is composed of preconditions, a delete list and an add list.} 
\item {\em External condition} - must be accomplished at the required task, but under the assumption that it is satisfied by some other task from the task network. An external condition can be seen as a sequencing constraint.
\item {\em Filter condition} - decides on task relevance to a particular situation. In the case of method relevance to a certain task decomposition, this condition reduces the branching factor by eliminating inappropriate methods.
\item {\em Query condition} - accomplishes queries about variable bindings or restrictions at some required point in a task network.
\item {\em Compute condition} - requires satisfaction by information coming only from external systems, such as a database.
\item {\em Achieve condition} - allows expressing goals that can be achieved by any means available to a planner.
\end{itemize}

\paragraph*{Note on Filter Conditions}
According to Erol et al.~\cite{erol1996:formalization}, filter conditions can be used to prune the search space. On the contrary, Kambhampati discusses that filter conditions can be seen as selection heuristics rather than pruning mechanisms~\cite{kambhampati1995:comparative}. In particular, preferring task networks with already established filter conditions is not the same as pruning. Kambhampati argues that filter conditions should be seen as an integral part of the fashion of allowing the domain author greater control over the types of generated solutions, that is, filter conditions enable the domain author to disallow certain types of solutions. In Kambhampati's view, the loss of completeness using filter conditions, a point made by~\cite{collins1992:filter}, is a ramification of the erroneous interpretation and implementation of filter conditions.  

\subsubsection{Search Space}\label{sec:ss}
So far we described concepts that affect the structure of the space to be searched. Next, we intuitively describe two structures of search spaces created by \HTN~planners. The first type of space consists of task networks and task decompositions as evolutions from one task network to another. Given some planning problem, at the beginning of the search, a task decomposition is imposed on the initial task network, and the process continues by repeatedly decomposing tasks from a newly created task network until a primitive task network is produced. A linearisation of this primitive task network executable in the initial state represents a solution to the planning problem.

The second type of search space is a subset of the state space. The subset consists of explicitly described states restricted by task decompositions. As in the classical state space, the search begins in the initial state with an empty plan, but instead of searching for a state that will satisfy the goal, the search is for a state that will accomplish the initial task network. In particular, if a task from the task network is compound, a task decomposition is performed and the search continues on the next decomposition level, but in the same state. If the task is primitive, it is executed and the search continues into a successor state. This task is then added to the plan. When there are no more tasks in the task network to be decomposed, the search is finished. The solution to the planning problem is the plan containing a sequence of totally ordered actions.

\paragraph{Categorisation of \HTN~planners}
In the first type of search space, the initial task network is reduced to a primitive task network that constitutes a solution to the planning problem. At each point in the space, the task network can be seen as a partially specified plan until the search reaches the point where the task network is primitive and represents a solution plan. Thus, we employ the term {\em plan space} to refer to this type of search space. We refer to \HTN~planners that search in this plan space as {\em \pHTN planners}, and to the model of \HTN~planning as {\em \pHTN planning}. For the obvious reasons, we employ the term {\em state space} to refer to the second type of search space. Thus, we refer to \HTN~planners searching in this space as {\em \tHTN planners}, and to the model of \HTN~planning as {\em \tHTN planning}.

\subsection{Plan-Based HTN Planning}\label{sec:plan-htn}
We draw the formalism of \pHTN planning upon the work of Geier and Pascal~\cite{geier2011:decidability}. With respect to the  Definition~\ref{def:tn} of the theoretical framework, we complement a task network as follows.

\begin{mydef}[Task network]
A \textit{task network} $tn$ is a triple $(T,\varphi,\psi)$, where 
\begin{itemize}
\item $T$ is a finite and non-empty set of tasks
\item $\varphi: T \to T_{n}$ labels a task with a task name
\item $\psi$ is a formula composed by conjunction, disjunction or negation of the following sets of constraints:
	\begin{itemize}
	\item $\prec\: \subseteq T \times T$ is a strict partial order on T (irreflexive, transitive, asymmetric)
	\item $\mapsto\: \subseteq V \times V \cup V \times C$ is a restriction on bindings of task network variables
	\item $\vdash_{\prec}\: \subseteq T \times Q \cup Q \times T \cup T \times Q \times T$ is a partial order on tasks and state predicates.
	\end{itemize}
\end{itemize}
\end{mydef}

Since some task name can occur many times in one task network, task labelling enables identifying uniquely many occurrences of that task name. For example, $tn=(\{t_{1},t_{2},t_{3}\},\{(t_{1},t'),(t_{2},t''),(t_{3},t')\},\emptyset)$ denotes that the task network consists of two tasks associated with task name $t'$ and one task associated with $t''$.

A task network $tn = (T,\varphi,\psi)$ is {\em isomorphic} to $tn' = (T',\varphi',\psi')$, denoted as $tn \equiv tn'$, if and only if there exists a bijection $\beta: T \rightarrow T'$, such that
\begin{itemize}
\item for all $t,t' \in T$ it holds $(t,t') \in\: \prec$ if and only if $(\beta(t),\beta(t')) \in\: \prec'$
\item for all $v_{1},v_{2} \in V$ and $c \in C$ it holds $(v_{1},v_{2}) \in\: \mapsto$ or $(v_{1},c) \in\: \mapsto$ if and only if there exist $v_{1}',v_{2}' \in V$ and $c' \in C$ such that $v_{1}=v_{1}'$, $v_{2}=v_{2}'$ and $(v_{1}',v_{2}') \in\: \mapsto'$ or $v_{1}=v_{1}'$, $c=c'$ and $(v_{1}',c) \in\: \mapsto'$
\item for all $t,t' \in T$ and $p \in Q$ it holds $(t,p) \in\: \vdash_{\prec}$ or $(p,t) \in\: \vdash_{\prec}$ or $(t,p,t') \in\: \vdash_{\prec}$ if and only if $(\beta(t),p) \in\: \vdash_{\prec}'$ or $(p,\beta(t)) \in\: \vdash_{\prec}'$ or $(\beta(t),p,\beta(t')) \in\: \vdash^{'}_{\prec}$
\end{itemize}
and $\varphi(t) = \varphi'(\beta(t))$.

\begin{mydef}[Decomposition]
Let $m$ be a method and $tn_{c} = (T_{c}, \varphi_{c}, \psi_{c})$ be a task network. Method $m$ {\em decomposes} $tn_{c}$ into a new task network $tn_{n}$ by replacing task $t$, denoted as $tn_{c} \underset{t,m}{\longrightarrow_{D}}tn_{n}$, if and only if $t \in T_{c}$, $\varphi_{c}(t) = c(m)$, and there exists a task network $tn' = (T', \varphi', \psi')$ such that $tn' \equiv tn(m)$ and $T' \cap T \neq 0$, and
\begin{align}
\begin{split}
tn_{n}:=&((T_{c} \setminus \{t\}) \cup T', \varphi_{c} \cup \varphi', \psi_{c} \cup \psi' \cup \psi_{D})\; \text{where}\\
\psi_{D}:=&\{(t_{1},t_{2}) \in T_{c} \times T'\:|\:(t_{1},t) \in\: \prec_{c}\}\: \cup \{(t_{1},t_{2}) \in T' \times T_{c}\:|\:(t,t_{2}) \in\: \prec_{c}\}\: \cup \\
&\{(p,t_{1}) \in Q \times T'\:|\:(p,t) \in\: \vdash_{\prec c}\}\: \cup \{(t_{1},p) \in T' \times Q\:|\:(t,p) \in\: \vdash_{\prec c}\}\: \cup \\ 
&\{(t_{1},p,t_{2}) \in T' \times Q \times T'\:|\:(t,p,t_{2}) \in\: \vdash_{\prec c}\} \nonumber
\end{split}
\end{align}
\end{mydef}

Given a planning problem $\mathcal{P}$, $tn_{c}\rightarrow^{*}_{D} tn_{n}$ indicates that $tn_{n}$ results from $tn_{c}$ by an arbitrary number of decompositions using methods from $M$.

\begin{mydef}[Executable Task Network]
Given a planning problem $\mathcal{P}$, $tn=(T,\varphi,\psi)$ is {\em executable} in state $s$, if and only if it is primitive and there exists linearisation of its tasks $t_{1},\dots,t_{n}$ that is compatible with $\psi$ and the corresponding sequence fo operators $\varphi(t_{1}),\dots,\varphi(t_{n})$ is executable in $s$.
\end{mydef}

\begin{mydef}[Solution]
A task network $tn_{s}$ is a solution to a planning problem $\mathcal{P}$, if and only if $tn_{s}$ is executable in $s_{0}$, and $tn_{0}\rightarrow^{*}_{D}tn_{s}$ for $tn_{s}$ being a solution to $\mathcal{P}$.
\end{mydef}

Our definition of the plan space is similar to the definition of the decomposition problem space in~\cite{alford2012:htnspaces}. Intuitively, a problem space is a directed graph in which task networks are vertices, and a decomposition of one task network into another task network by some method is an outgoing edge, under the condition that the initial task network belongs to the graph. More precise definition follows.

\begin{mydef}[Plan space]
Given a plan-based \HTN~planning problem $\mathcal{P}$, a {\em plan space} $PG$ is a directed graph $(V,E)$ if and only if $tn_{0} \in V$, and for each $tn\rightarrow_{D}tn'$: $tn,tn' \in V$ and $(tn,tn') \in E$.
\end{mydef}

\begin{figure*} [!ht] 
\centering
\includegraphics[width=.4\textwidth]{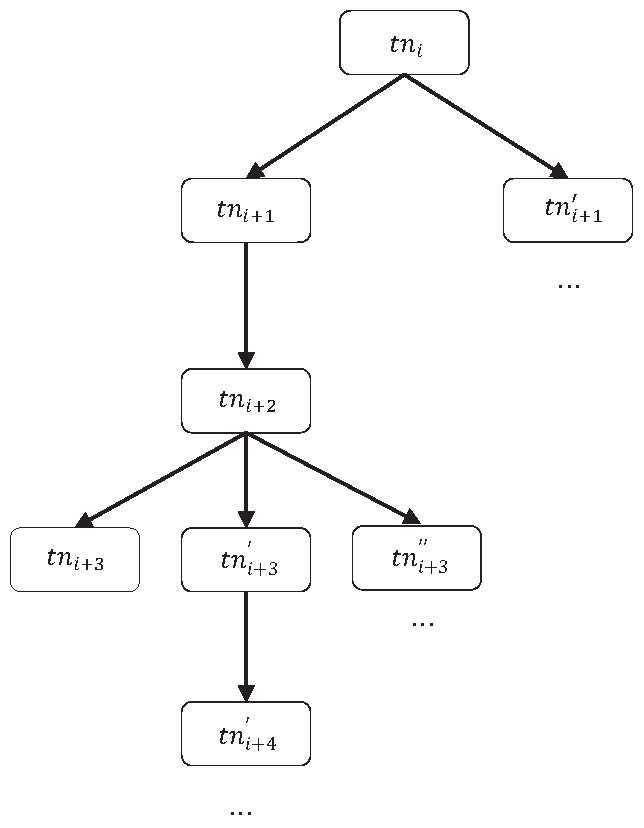}
\caption{Plan Space}\label{fig:plan-space}
\end{figure*}

An example of a \HTN~plan space is illustrated in Figure~\ref{fig:plan-space}.

\subsubsection{Review of Plan-Based HTN Planners}\label{sec:plan-review}

\subsubsection*{Task Decomposition} 
Most \pHTN planners perform task decomposition in a slightly different way than the process described in Section~\ref{sec:td}. The main reason lies in the approach that these planners use to distinct and represent tasks. In fact, with the exception of \umcp, the rest of the planners support only a single structure to encode both primitive and compound tasks. Although it is not always clear what is the purpose of a respective structure or how task decomposition is accomplished, we try to describe the approach that each planner takes to represent a task, and consequently, a high-level explanation of what does the main idea behind task decomposition consist of in each planner.

\noah~considers a network of partially ordered nodes, where each node represents one partially specified task. The planner does not have a clear distinction between primitive and compound tasks, but instead, each task contains some procedural code, declarative code, and links to other tasks. The procedural code denotes the body and the declarative code denotes the effects of a task. A task decomposition consists of evaluating a task body in terms of inserting new tasks in the current task network and applying the task effects. The newly created task network is then checked for potential interactions. In \nonlin, a task network is a collection of nodes, where a node can be considered as a task represented completely in declarative form. A task is represented by a single structure (or schema). This schema describes how a particular task can be accomplished, that is, it defines a special construct (the `expansion' tag) that can contain either a set of tasks, each of which can be further decomposed, or a so-called `null expansion' when the schema represents a primitive task. Given a task network, \nonlin~performs a task decomposition such that if a task schema introduces a `null expansion', the task is applied and the process continues with the next task of the task network, and if the  schema introduces a list of tasks, the planner expands the task network with the list of tasks and checks for interaction corrections. 

\sipe~also uses only one structure (called `Operator') to refer to its primitive and compound tasks. Inside this structure, a specific construct (the 'Plot' tag) is used to specify, both, an action to be applied or a network of tasks to be further decomposed. If the structure represents a primitive task, the construct contains one task (called `PROCESS' node). If the structure represents a compound task, the construct contains a set of tasks, each of which can require a particular predicate to be achieved (a `GOAL' node), or a task to be performed (a `PROCESS' or a `CHOICEPROCESS' node). A task decomposition consists of the following. Each compound task is decomposed by evaluating a given structure and replacing the task with the set of tasks. The task network is controlled against any newly introduced interactions and corrected accordingly. Each primitive task is copied to the current task network, and task's (deductive) effects are recalculated based on the context introduced with the modified task network. 

\oplan~represents and handles tasks similarly as in \nonlin, that is, the planner uses a schema to describe primitive and compound tasks, and given a task network, the planner examines and determines whether there is a schema to be decomposed or a schema to be applied in order to bring the task network into a more detailed level. Finally, \umcp~uses explicit notations about primitive (an `operator' tag) and compound (a `declare-method' tag) tasks. The description for the latter type of task includes a list of tasks (the `expansion' tag) and a set of variable binding and ordering constraints associated with the list of tasks (the `formula' tag). Given a network of compound tasks, a task decomposition retrieves all methods associated with a chosen compound task, expands the task network by applying each method to the chosen task, and returns the resulting set of task networks. Each task network is examined for potential conflicts and appropriate resolution steps are initiated.

\subsubsection*{Constraints} 
We analyse this concept with respect to other constraint-related concepts, such as the commitment strategy and constraint management.

\subparagraph{Commitment Strategy.} \PHTN planners take advantage of the least-commitment strategy. However, two deficiencies can be observed. First, except for \umcp, the rest of the planners take a rigid approach of incorporating the commitment strategy into the search mechanism, resulting in tightly coupled and inflexible planning systems. Second, only few planners backtrack on poor decisions, thus not implementing completely the concept of the least-commitment strategy. We confirm the latter observation by examining at which point during the search these planners make decisions and whether the planners backtrack on different choices with respect to variable bindings and task orderings. 

\noah~takes a simple but limited approach. It uses the commitment strategy in such a way that when a task is decomposed, task ordering and variable binding decisions are delayed. \noah~does not backtrack on its choices. The commitment strategy in \nonlin~is applied when the planner decomposes tasks in a breath-first manner in the way that variables are bound immediately after they are inserted into the task network, and all constraints are applied before the next decomposition. The planner keeps choices for alternative orderings among tasks and alternative variable bindings. \sipe~follows the least-commitment strategy in a depth-first manner by firstly phantomising a goal, which may instantiate variables, then decomposing a task, delaying task ordering decisions and employing a more `intelligent' (with respect to its predecessors and according to the author in ~\cite{wilkins1988:extending}) choice of variable bindings. In many cases, the planner avoids backtracking for alternative task orderings and alternative variable bindings. \oplan~uses the least-commitment strategy similarly as described in MOLGEN~\cite{stefik1981:constraints}. In particular, the planner deals with a list of entries (called `agenda') that needs to be work out in order the task network to be considered as accomplished. An entry could be decomposing a task, adding ordering or variable binding constraints, etc. In other words, these are pending decisions. For each entry, a priority is calculated and attached to it. Generally, the priority could be expressed as a heuristic value, but in the case of \oplan~(as described in~\cite{tate1994:oplan2}) the priority values are fixed by the designer. With this priority, a candidate from the list is chosen and an appropriate modification to the task network is applied. While doing this, the planner takes into consideration alternative variable bindings. \umcp~uses the least-commitment strategy to defer task ordering and variable bindings. From the variable bindings perspective, refinement of variable or state constraints either prunes possible values or records variable distinctions. The planner is loosely coupled with the commitment strategy, which enables other commitment approaches to be plugged in. Thus, \umcp~is tested with two more commitment strategies, namely an eager-commitment strategy and a dynamic-commitment strategy for which detailed discussions can be found in~\cite{tsuneto1996:commitment}. 

The least-commitment strategy is not the only approach that \pHTN planners take to manipulate commitments. \sipe, \oplan~and \umcp~provide users with the ability to control commitments interactively. In \umcp, for example, at each decision point, the user is provided a process step to be applied to a task network. The user may accept the suggested step, or can select another step from a list of steps allowed to be applied to the current task network. Another example of interactive commitments is provided below in the description of the phantomisation process in \sipe.

\subparagraph{Constraint Management: Harmful Interactions.} \noah~uses the `Resolve Conflicts' method to identify and resolve deleted-condition interactions. In order to identify these interactions, the planner uses a special table (called 'Table Of Multiple Effects') that stores an entry for each expression and its value that is asserted or threatened by more than one task in a task network. For example, the statement \texttt{(truck-at t1 l2)=true} is a valid entry in the table. By having this information, the planner can recognise a conflict whenever an expression, which is asserted by some task, is threatened by a task that is not the asserting one. As described in Section~\ref{sec:cf}, the resolution consists in propagating a constraint stating that the threatened task must be performed first. A double-cross interaction is recognised and resolved by the `Resolve Double Cross' method. \noah~resolves this conflict by identifying which variable is related to the assertion or denial of expressions that led to the conflict. Consequently, the planner finds and inserts other appropriate expressions and/or tasks into the task network ensuring that variables are bound differently at the time of the assertion or denial. The task network with the new expressions and/or tasks must be decomposed again, and in the case of a newly appeared conflict, the planner fails to find a solution.

In contrast to \noah, \nonlin~introduced a rather simple mechanism to deal with interactions. The planner does not treat interactions individually, but provides a unified solution. A causal representation of a task network (called 'Goal structure') is used to store information about the task network that would be difficult to extract from the task network itself. With this explicit representation, which equals to the causal link described in Section~\ref{sec:cf}, track of task conditions with the points in the network where these conditions are or could be achieved is kept. In the case of interactions, the planner uses its linking process and a table to suggests corrections (similarly to \noah). These corrections are in the form of linearisations of the network. Linearisations make a particular expression to have correct value for any point in the network to have tasks performed. 

In \sipe, harmful interactions are detected and resolved by the `Solving Harmful Interactions' method. In fact, the method recognises three cases of harmful interactions. The first case relates to the deleted-condition interaction. As usual, the solution is to order the task network to first accomplish the segment with the precondition and then the corresponding goal. The second case is when a side effect conflicts with a goal (in \sipe, a task can have expected and side effects; an expected effect is a fact the planner expects to be true within a task network; a side effect happens otherwise). The solution is to order the task network to first apply the side effect and then the goal. The last case relates to a double-cross interaction. Since there is no ordering that will achieve both goals, \sipe~tries to re-achieve one of the goals later, and if it fails, it backtracks and uses another task at a higher level. Resource interactions are solved by the `Resource Conflicts' method. If a resource-resource interaction is detected, the planner adds ordering constraints as a resolution step. If a resource-argument interaction is identified, a heuristic is used to first order the task (or a part of the task network) using the object as a resource and then the task (or a part of the task network) using the same object as an argument. \sipe~uses an additional method to check interactions between planning variables. The `Solving a Constraint Network' method is called at each planning level. In particular, a depth-first search is used to find an appropriate binding for each variable such that no constraints are violated. If no solution is found, the planner prunes the current task network from the search. The planner saves the last found solution for the global constraint network and at each point tries the choice that worked in the previous solution.

In \oplan, interactions are handled with the help of constraint managers. A constraint manager is a component that provides efficient support to a higher-level component and does not make any decisions by itself. As such, the constraint manager must maintain information which can be used to prune the search space (if a task network is found to be invalid) or to order search alternatives according to some heuristics~\cite{tate1994:oplan2}. Three constraint managers are used to resolve interactions. The first manager (called `TOME/GOST Manager') enables insertion of conflict-free effects and conditions into a task network. The potential conflicts are resolved with support of the Question-Answering (QA) mechanism. The QA provides answers about the value of some proposition at a particular point in a task network. The second manager (called `Plan State Variable Manager') is used to handle interactions and dependencies among values and variables. This manager maintains the consistency of constraints over these values and variables. Finally, the last manager (called `Resource Utilisation Manager') is used to monitor and provide information about resource levels and resource utilisation.

\umcp~uses one resolution method to deal with interactions in a task network. Given a task network to be corrected, the method outputs a set of task networks each of which potentially resolves some of the conflicts in the given task network. \umcp~is the only \pHTN planner that clearly defines how resolution is accomplished. Since this planner is closest to our definition of \pHTN planning, we describe in more details the inclusion of constraint operations in this method. The method incorporates four operations. The first operation (called `constraint selection') determines the constraints to deal with. Given a task network $tn$, the operation returns $k$ constraint lists that are used to transform the task network into $k$ branches. In a branch $i$, the $i$-th constraint list of the form ($c_{i1}c_{i2}\dots$) is enforced. For each constraint type $c_{i1}$ in the $i$-th constraint list, the second operation (called `constraint enforcement') produces a set of task networks that make $c_{i1}$ true. Each of the produced task network is further refined by calling this operation associated with $c_{i2}$, etc., until a set of task networks is produced that make ($c_{i1}c_{i2}\dots$) true. The next operation is constraint propagation and is adapted as follows. The input to it is a set of task networks from the previous operation. For each of these task networks, the list of constraints the planner has committed to make true (but has not done yet) is examined. It is checked whether some constraints can be enforced at the current level of details in the task network. If such a constraint is found, the previous operation (constraint enforcement) associated with that constraint is invoked. If no constraint is found, the task network is passed to the last operation. Finally, the last operation (called `constraint simplification') consists of evaluation and simplification steps for every type of constraints. If the current level of details in the task network is sufficient to evaluate a constraint, then the task network whose constraints evaluate to false is pruned. Remaining task networks form the output of the resolution method. If the current level of details is not sufficient, then the evaluation step returns either the constraint itself or a simplified version of it~\cite{erol1996:formalization}.

\subparagraph{Constraint Management: Helpful Interactions.} \noah~employs the `Use Existing Objects' method to identify placeholder-replacement interaction, and thus to replace a formal object with an actual value. In \noah, the use of this method may imply merging some tasks in a task network in terms of task reordering and partial linearisation of the task network. The planner uses the `Optimise Disjuncts' method to handle situations when disjunct optimisation is possible. The method builds planner's special tables (a table refers to `Table of Multiple Effects') for each choice among disjuncts, examines these tables and looks for a choice that will substantially minimise the number of tasks in the overall task network. If such disjunct exists, it is selected and the rest of the disjunction is ignored. Furthermore, \noah~identifies phantomisation by using the `Eliminate Redundant Preconditions' method. When redundant goals are found, the method simply eliminates them. The rationale behind this action is to avoid redundant processing in further planning levels, and to save memory space. In \nonlin, the phantomisation is handled by the linking process exactly as described in Section~\ref{sec:constraints}. Furthermore, \sipe~recognises phantomisation by using the `Goal Phantomisation' method. Since the planner is not able to solve this situation correctly by itself, it provides the user with two options to phantomise goals, particularly by using variable bindings or linearisations~\cite{wilkins1991:practical}. In the case of variable bindings, the user has three choices, namely: bind a variable whenever possible, never post constraints, or bind a variable only when there is one possible binding to accomplish the goal. In the case of linearisations, the user has also three choices: never add ordering constraints for the purpose of phantomisation, add ordering constraints when no other constraints are required to accomplish the phantomisation, or add ordering constraints when there is a parallel branch with an effect that has only one possible instantiation for accomplishing the phantomisation. \oplan~uses existing tasks in the network to satisfy some state constraints at any point the network. \oplan~is supposed to handle all situations that need to be phantomised in \nonlin. 

In \umcp, the phantomisation is performed in a different way. If some predicate $q$ of a (goal) task $t$ is already true, an empty task network represents its accomplishment. For that purpose, for each such task, there is an explicit method that includes a dummy primitive task $t_{d}$ with no effects, and the condition that the predicate $q$ should be true immediately before the $t_{d}$.

\subparagraph{Summary}
Table~\ref{tab:resolution-methods-plan} summarises and classifies resolution methods with respect to the task interaction they solve. If a cell contains `\ding{53}', it means that the planner does not need to and does not handle the respective interaction. If a cell is empty, then it means that the information was not available from the literature.

\begin{table*}\footnotesize 
\centering
\caption{Resolution Methods for Task Interactions in Plan-Based \HTN~Planners}\label{tab:resolution-methods-plan}
\begin{tabular}{L{2.1cm}|L{2.7cm}L{1.9cm}L{2.5cm}L{2.4cm}L{2.2cm}}
\toprule
{\bf Interaction} & {\bf \noah} & {\bf \nonlin} & {\bf \sipe} & {\bf \oplan} & {\bf \umcp} \\ \hline
Deleted-condition & Resolve Conflicts & Linking Process & Solving Harmful Int. & TOME/GOST Man. & Resolution Method \\
Double-cross & Resolve Double Cross & \ding{53} & Solving Harmful Int. & \ding{53} & \ding{53} \\
Resource & \ding{53} & \ding{53} & Resource Conflicts & Resource Util. Man. & \ding{53} \\ \hline
Placeholder replacement & Use Existing Objects & & & & \\ 
Phantomisation & Eliminate Redun. Prec. & Linking Process & Goal Phantomisation & Question Ans.
 & Domain Method \\
Disjunct optimisation & Optimise Disjuncts & & & & \ding{53} \\
\bottomrule
\end{tabular}
\end{table*}

Constraint management involves heavy computation and makes \pHTN~planners face computational difficulties. It is hard to determine whether a given predicate is true or not at a given point in a task network. \nonlin, for instance, uses its QA mechanism to determine the value of such a predicate. However, the answer might not be deterministic due to the partial order of the tasks, that is, it may answer with `yes', when the predicate can have a particular value, `no', when the predicate can not have the particular value, or `maybe', when it may have the particular value, some other value or undefined. \sipe~employs a truth criterion to reason over the truthfulness of a formula. As the formula is formed by a conjunction of predicates, the formula truth criterion is further reduced to a rather complex predicate truth criterion. Moreover, \sipe~restricts the partial orderings (nonlinearity in \sipe's term) for the purpose of circumventing the NP-completeness of the truth criterion~\cite{wilkins1991:practical}. \oplan's truth criterion is an extended version of \nonlin's QA. For example, if the answer is `maybe', QA provides an alternative set of strategies to ensure the truth of a predicate. \umcp~extends Chapman's Modal Truth Criterion (MTC)~\cite{chapman1987:conjunctive}, which is similar to QA, to also consider compound tasks. \umcp~ uses the extended MTC to evaluate the state constraints and it runs in quadratic time~\cite{erol1996:formalization}.

\subsubsection*{Explicit Conditions}
We summarise and classify explicit conditions that \pHTN planners employ in Table~\ref{tab:conditions-plan}. Similarly to Table~\ref{tab:resolution-methods-plan}, a cell containing `\ding{53}' denotes that a planner does not support the respective condition. We observe that the idea of explicit conditions is initiated by \nonlin. The planner supports four types of conditions. A supervised condition is always inserted explicitly in a network (through a `GOAL' node). External conditions are considered by the planner only when all ``compound'' tasks are decomposed into ``primitive'' ones. In \sipe, a supervised condition ensures that the main effect of a task remains true until the condition holds. An external condition defines that the goal will be achieved by a possibly parallel action and phantomised by an ordering link~\cite{wilkins2000:sipe2_manual}. Conditions play a special role in \oplan, since the planner does not consider any notion of a goal (a `GOAL' node)~\cite{tate1994:oplan2}. Instead, \oplan~takes advantage of the achieve condition as the only one in which insertion of new tasks is allowed. In fact, this condition can be expressed in two ways. Firstly, the  `achieve at N' condition supports insertion of tasks in a task network without temporal restrictions. Secondly, the `achieve at N after $<$time point$>$' poses temporal restrictions on the parts of the task network that could satisfy this condition. The latter condition is a generalisation of the achieve condition in \noah, \nonlin~and \sipe~\cite{tate1994:conditions}. In contrast to \nonlin, \umcp~handles external conditions at higher planning levels to prune the search space. External conditions are represented as state constraints which means that the planner will look for variable bindings or task ordering that make the conditions true, but the planner does not allow an establishment of the conditions by insertion of new tasks into a task network, or by task decomposition. Filter conditions are represented as state constraints as well, and used by the resolution method to prune inconsistent task networks. If filter conditions are not affected by any task, then it suffices to check the initial state to evaluate them. The flexibility of \umcp~has been proven again by extending the planner to reason about implicit external conditions~\cite{tsuneto1998:external_conditions}. Thus, instead of being specified explicitly, external conditions occur as a result of the structure of the domain knowledge, and are detected by examining the domain knowledge. 

\begin{table*} \footnotesize 
\centering
\caption{Condition Types in Plan-Based HTN Planners}\label{tab:conditions-plan}
\begin{tabular}{l|lllll}
\toprule 
{\bf Condition}  & {\bf \noah} & {\bf \nonlin} & {\bf \sipe} & {\bf \oplan} & {\bf \umcp} \\ \hline
Supervised & precondition & supervised & protect-until & supervised & goal task\\ 
External & \ding{53} & unsupervised & external-condition & unsupervised & state constraints/external\\ 
Filter & \ding{53} & use-when & precondition & only\_use\_if & filter\\ 
Query & \ding{53} & \ding{53} & \ding{53} & only\_use\_for\_query & \ding{53}\\ 
Compute & \ding{53} & \ding{53} & \ding{53} & compute & \ding{53}\\ 
\multirow{2}{*}{Achieve} & \ding{53} & \ding{53} & \ding{53} & achieve at N & \ding{53}\\ 
& goal & goal & goal & achieve at N after $<$time point$>$ & goal task \\ 
\bottomrule
\end{tabular}
\end{table*}

\subsection{State-Based HTN Planning}\label{sec:state-htn}
We complement Definition~\ref{def:tn} and Definition~\ref{def:m} of the theoretical framework as follows.

\begin{mydef}[Task network]\label{def:tns}
A \textit{task network} $tn$ is a pair $(T,\prec)$, where
\begin{itemize}
\item $T$ is a finite set of tasks
\item $\prec$ is a strict partial order on $T$ (irreflexive, transitive, and asymmetric)
\end{itemize}
\end{mydef}

\begin{mydef}[Method]\label{def:ms}
A \textit{method} $m$ is a triple $(c(m),pre(m),tn(m))$, where $c(m)$ is a compound task, $pre(m) \in 2^{Q}$ is a precondition, and $tn(m)$ is a task network. The subsets $pre^{+}(m)$ and $pre^{-}(m)$ denote positive and negative precondition of $m$, respectively.
\end{mydef}

A method $m$ is applicable in state $s$, if and only if $pre^{+}(m) \subseteq s$ and $pre^{-}(m) \cap s = \emptyset$. Applying $m$ to $s$ results in a new task network.

\begin{mydef}[Decomposition]
Let $m$ be an applicable method in $s$ and $tn_{c} = (T_{c},\prec_{c})$ be a task network. Method $m$ decomposes $tn_{c}$ into a new task network $tn_{n}$ by replacing task $t$, written $tn_{c}\underset{s,t,m}{\longrightarrow_{D}}tn_{n}$, if and only if $t \in T_{c}$, $t=c(m)$ and
\begin{align}
\begin{split}
tn_{n}:=&((T_{c} \setminus \{t\}) \cup T_{m}, \prec_{c} \cup \prec_{m} \cup \prec_{D})\; \text{where} \\
\prec_{D}:=&\{(t_{1},t_{2}) \in T_{c} \times T_{m}\:|\:(t_{1},t) \in\: \prec_{c}\}\: \cup \{(t_{1},t_{2}) \in T_{m} \times T_{c}\:|\:(t,t_{2}) \in\: \prec_{c}\} \nonumber
\end{split}
\end{align}
\end{mydef}

\begin{mydef}[Decomposition]
Let $m$ be an applicable method in $s$ and $tn_{c} = (T_{c},\prec_{c})$ be a task network. Method $m$ decomposes $tn_{c}$ into a new task network $tn_{n}$ by replacing task $t$, written $tn_{c}\underset{s,t,m}{\longrightarrow_{D}}tn_{n}$, if and only if $t \in T_{c}$, $t=c(m)$ and
\begin{align}
\begin{split}
tn_{n}:=&((T_{c} \setminus \{t\}) \cup T_{m}, \prec_{c} \cup \prec_{m} \cup \prec_{D})\; \text{where} \\
\prec_{D}:=&\{(t_{1},t_{2}) \in T_{c} \times T_{m}\:|\:(t_{1},t) \in\: \prec_{c}\}\: \cup \{(t_{1},t_{2}) \in T_{m} \times T_{c}\:|\:(t,t_{2}) \in\: \prec_{c}\} \nonumber
\end{split}
\end{align}
\end{mydef}

A similar model of \HTN~planning as this one is presented in~\cite{alford2012:htnspaces} where the authors define a progression problem space. The space is a directed graph in which pairs of state and task network are vertices, and a progression from one pair to another is an outgoing edge. We take a slightly different approach in which a state is a vertex, and a task decomposition maps to the same state where the corresponding method is applicable, and operator application leads to a successor state.

\begin{mydef}[State space]
Given a state-based \HTN~planning problem $\mathcal{P}$, a {\em state space} $SG$ is a directed graph $(V,E)$ if and only if $s_{0} \in V$, and there is a state $s_{i}$ and $t_{k} \in tn$ such that
\begin{itemize}
\item if $t_{k}$ is primitive, then $s_{i}\underset{t_{k}}{\longrightarrow}s_{i+1}$ such that $k=i+1$, $s_{i},s_{i+1} \in V$ and $(s_{i},s_{i+1}) \in E$; or
\item if $t_{k}$ is compound, then $tn\rightarrow_{D}tn'$ is a self-transition such that $s_{i} \in V$ and $(s_{i},s_{i}) \in E$.
\end{itemize}
\end{mydef}

\begin{figure*} 
\centering
\includegraphics[width=.6\textwidth]{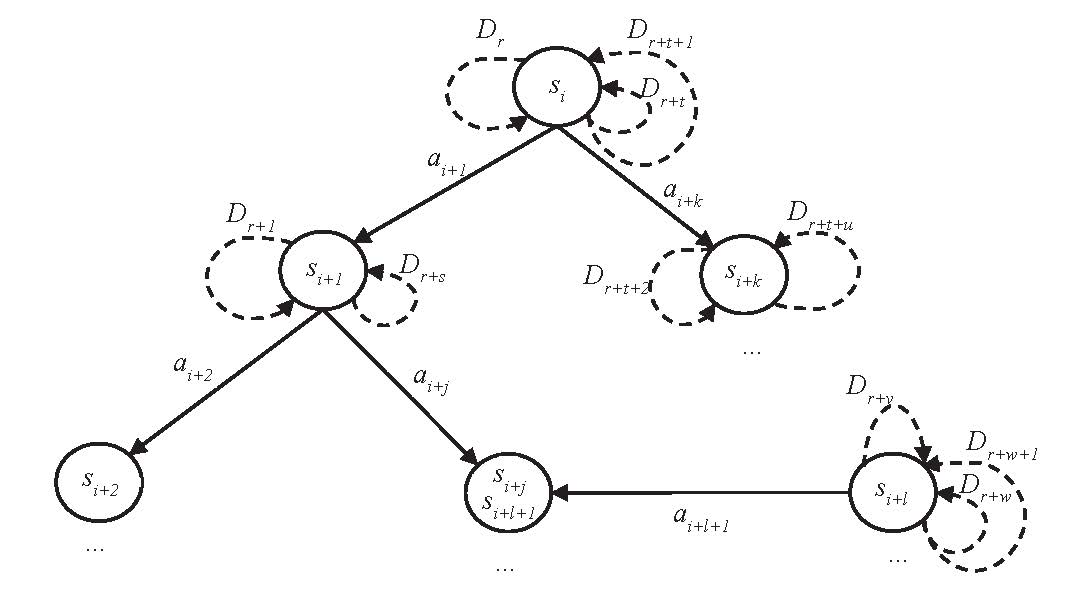}
\caption{State Space}\label{fig:state-space}
\end{figure*}

The \HTN~state space is illustrated in Figure~\ref{fig:state-space}.

\subsubsection{Review of State-Based HTN Planners}\label{sec:state-review}
We explore chronologically \tHTN planners in terms of the concepts described in Section~\ref{sec:cf}. We show that these planners employ a simpler approach to planning compared to \pHTN planners. The concept choices are easy to understand and indeed influence the structure of the space.

\subsubsection*{Task Decomposition}
\shop~and \siadex~use a specific construct (`method' and `task', respectively) to represent a compound task amenable to several {\em ways} of accomplishment. Each way (`method' in \siadex) contains applicability conditions and a task network. In both planners, the set of ways can be seen as an if-then-else representation, that is, the planners select the first way whose if-statement (preconditions) holds in the current state. Thus, given a compound task, a task decomposition evaluates the preconditions of task's associated ways, and chooses the first way applicable in the current state to expand the existing task network. Two observations are in order. First, recall that in Section~\ref{sec:td} we stated that the task decomposition makes a non-deterministic choice of which method to use for the decomposition. However, in the case of both planners, the choice is controlled, that is, the first method from the if-then-else representation that is applicable in the state is chosen. Second, the newly composed network does not need to be checked for corrections.

Among \tHTN planners, SHOP uses the ordered task decomposition, while \shop~the unordered task decomposition (by using `ordered' and `unordered' tags to denote totally and unordered task networks, respectively). In \shop, effort is made to bring the planner closer to the partially ordered task decomposition by allowing partial restrictions on the order of unordered task networks. In particular, if some method with a task network $tn$ has a task $t_i$ that begins with the `immediate' tag, then the planner knows that it must do $t_i$ immediately after $t_{i-1}$ finishes, without trying to interleave other tasks between $t_{i-1}$ and $t_i$. \siadex, on the other hand, follows the partially ordered task decomposition. Tasks in a network can be ordered, unordered and in parallel.

\subsubsection*{Constraints} Similarly to the previous section, we analyse this concept with respect to the commitment strategy and constraint management of \tHTN planners.

\subparagraph{Commitment Strategy.} \THTN planners employ the early-commitment strategy. Thus, they know the current state at each step in the planning process. In fact, the state is transformed only by applying an action to it. With respect to task ordering, \shop~and \siadex~greatly benefit from adopting the forward chaining approach in which chaining of actions is achieved by imposing total order on (some) plan actions. The total ordering ensures that neither the current action to be added to the plan can interfere with some earlier action's preconditions or effects, nor a later action can interfere with current action's preconditions or effects. With respect to variable bindings, all variables are instantiated when task preconditions are evaluated and an action is applied. Thus, all actions in a plan are fully specified. The evaluation of preconditions implies a list of all possible sets of variable bindings that satisfy these preconditions in the current state. If some task later fails, both planners backtrack on other alternatives according to the variable bindings left in the list, or maybe to some criterion specified in the definition of the task. Considering the truth value of the preconditions, \shop~and \siadex~prune inappropriate tasks from the search space.

\subparagraph{Constraint Management.} By taking the early-commitment strategy into consideration, we could conclude that \tHTN planners avoid task interactions altogether. However, this statement is not entirely correct. Table~\ref{tab:resolution-methods-state} suggests that \tHTN planner avoid most interactions. In \shop, for example, a deleted-condition interaction may arise due to the process of interleaving tasks. The planner is able to solve this situation under a rather restricting assumption, that is, it requires a specification of `protection' conditions in the effects of operators. A protection request enforces the planner from deleting conditions, and a protection cancellation allows the planner to delete these conditions. For example, consider our running-example domain. To give knowledge to \shop~that after driving the truck \texttt{t} from location \texttt{l\_f} to location \texttt{l\_t}, the truck should stay at \texttt{l\_t} until a box is loaded into \texttt{t}, we need to modify the operator \texttt{drive} by adding a protection request, denoted by \textit{:protection}, for \texttt{truck-at(t,l\_t)}, i.e., \texttt{(:protection truck-at(t,l\_t))}. A protection cancellation can be added into the negative effects of the \texttt{load-truck(t,b,l)} operator in a similar manner.
\siadex~needs a more powerful mechanism to accomplish planning and handle interactions that may arise in partially ordered task networks. The planner uses a causal structure of tasks and task networks. Constraint satisfaction checks the consistency of task networks (and the solution) based on the causal structure, and constraint propagation is used to post constraints, if necessary.

\THTN planners include a certain form of phantomisation. In \shop~and \siadex, similarly as in \umcp, the phantomisation of a task is explicitly encoded in the domain knowledge. The planners handle the phantomisation of a rather recursive task by taking into account an alternative decomposition which explicitly encodes a `do-nothing' operation. Therefore, one could say that the planners lack the ability to infer such situations by themselves. This issue is addressed in~\cite{georgievski2011:phantomisation}, where authors extend the JSHOP2 planner~\cite{shop}, that is, a Java implementation of \shop, to reason about the conditions that are already achieved without the need of explicit specification into the domain knowledge at the expense of spending negligible additional planning time.

For example, consider our running-example domain. Let say that the truck \texttt{t1} should take the box \texttt{b1} from a location \texttt{l3} to a location \texttt{l1}, and from there to a location \texttt{l2}, and from \texttt{l4} the box should be transported by a plane \texttt{p} to the final location \texttt{l4}, under the assumption that the locations are adjacent as described, and that \texttt{l1}, \texttt{l2} and \texttt{l3} are in a same city, while \texttt{l4} is in another city. For that purpose, in the scope of the \texttt{deliver} task, we can write a method with several branches. One decomposition could describe how to transport a box by a plane, another decomposition could describe how to transport a box with a truck, and, finally, the last decomposition describes explicitly that the transportation of the box is finished since the box is already at the desired location.
\begin{alltt}
task:         deliver(b,l_f,l_t)
way 1:        transport a box with a plane
precondition: plane-at(p,l_f),adjacent(l_f,l_n),box-at(b,l_f),
              in-city(l_f,c_f),in-city(l_n,c_n),different-city(c_f,c_t)
task network: \(\langle\)load-plane(b,p,l_f),fly(p,l_f,l_n),
              unload-plane(b,p,l_n),deliver(b,l_n,l_t)\(\rangle\)

way 2:        transport a box with a truck
precondition: truck-at(t,l_f),adjacent(l_f,l_n),box-at(b,l_f),
              in-city(l_f,c_f),in-city(l_n,c_n),same-city(c_f,c_n)
task network: \(\langle\)load-truck(b,t,l_f),drive(t,l_f,l_n),
              unload-truck(b,t,l_n),deliver(b,l_n,l_t)\(\rangle\)

way 3:        no transport needed
precondition: box-at(t,l_t)
task network: \(\langle\)\(\rangle\)
\end{alltt}

\subsubsection*{Explicit Conditions}
\THTN planners do not share the strong need for explicit conditions with \pHTN planners, as shown in Table~\ref{tab:conditions-state}. The whole reasoning power of \shop~and \siadex~is encapsulated in the preconditions of both primitive and compound tasks, thus they do not require other explicit domain knowledge. In the scope of preconditions, however, \shop~enables several types of computations, such as invocations of external knowledge resources by using the 'Call' condition. \siadex~also supports complex computations by incorporating complete (Python-based~\cite{python}) procedures in the domain. External conditions are modelled in a similar fashion.

\begin{table*}\footnotesize
\parbox{.4\textwidth}{
	\centering
	\caption{Resolution Methods for Task Interactions in \THTN planners}\label{tab:resolution-methods-state}
	\begin{tabular}{l|ll}
	\toprule
	{\bf Interaction} & {\bf \shop} & {\bf \siadex} \\ \hline
	Deleted-condition & Protection condition & Constraint propagation \\
	Double-cross & \ding{53} & \ding{53} \\
	Resource & \ding{53} & \ding{53} \\ \hline
	Placeholder replacement & \ding{53} & \ding{53} \\ 
	Phantomisation & Domain method & Domain method \\
	Disjunct optimisation & \ding{53} & \ding{53} \\
	\bottomrule
	\end{tabular}
}
\hfill
\parbox{.4\textwidth}{
	\centering
	\caption{Condition Types in \THTN Planners}\label{tab:conditions-state}
	\begin{tabular}{l|ll}
		\toprule 
		{\bf Condition}  & {\bf \shop} & {\bf \siadex} \\ \hline
		Supervised & \ding{53} & \ding{53} \\
		External & Call & Arbitrary code \\
		Filter & precondition & precondition \\
		Query & \ding{53} & \ding{53} \\
		Compute & Call & function \\
		Achieve & \ding{53} & \ding{53} \\
		\bottomrule
	\end{tabular}
}
\end{table*}

\section{Analysis of \HTN~Planning and \HTN~Planners}\label{sec:analysis}

So far we categorised \HTN~planners based on the space they search in. This categorisation highlights the common search-related features among planners, though we have two more reasons for analysing the planners and \HTN~planning. First, this categorisation does not cover the capabilities of \HTN~planners with respect to the level od knowledge these planners require, which expressiveness constructs the planners support, and what is the performance of these planners. The second reason lies in some of the implicit assumptions made about \HTN~planners, that is, claims and beliefs accepted for granted and without evidence. These include the ``sophistication'' of domain knowledge provided to \HTN~planners, the expressive power of \HTN~planning in theory and practice, \HTN~planners being fast and scalable, and being \HTN~planning very suitable for and most applied to real-world problems.

To this end, we provide the third piece of our framework pie, that is, we develop an analytical framework to collect and organise studies on \HTN~planning and planners. Then, we apply exploratory research to examine diversity and similarity of \HTN~planning within their category and between categories, and comparative research to make sense of a range of cases. In this way, we believe that statements about domain knowledge, practical expressiveness, performance, and applicability can be made in a neutral and evidence-oriented way. 

\subsection{Analytical Framework}\label{sec:af}
We develop an analytical framework to collect and organise data about \HTN~planners. The framework directs us on where to look and what kind of properties to look for but without making specific hypotheses about relationships among properties. The framework consists of five main elements, namely, domain authoring, expressiveness, competence, performance, and applicability. Each element, the motivation for its inclusion in the framework, and its corresponding properties are defined in the following subsections.

\subsubsection{Domain Authoring}
An interesting perspective on \HTN~planners says:

\begin{quote}
``[Compared with classical planners,] the primary advantage of HTN planners is their sophisticated knowledge representation [and reasoning capabilities]~\cite{nau2004:automated}.''
\end{quote}

\noindent Two remarks are in order. First, there is uncertainty in the meaning of ``sophisticated''. Does it refer to the complexity, richness or some other attribute of the representation? For now, let us assume that it refers to the so-called ``knowledge-rich'' representation~\cite{wilkins2001:call}. The second remark is on \HTN~planners taking advantage of the use of knowledge-rich encoding. On the one hand, this could be correct, if we consider that these planners improve their performance (over classical planners) thanks to their domain knowledge~\cite{long2003:ipc3}. On the other hand, why are \HTN~planners in advantage if we do not know at what expense, in terms of encoding effort, we obtain that improvement?

We define {\em domain authoring} the process of domain knowledge performed by a domain author. What we are interested in, in this process, is the relative effort needed to formulate such domain knowledge for an \HTN~planner. However, the community has not yet found a way or measures to provide an objective answer to this type of questions, thus the answer remains rather anecdotal. It is ambiguous and difficult to define an answer especially because it directly depends on the capabilities and experience of the domain author with respect to the understanding of the underlying planning system and the expertise for the respective domain. 

With the observation that the knowledge-rich representation is a requirement for \HTN~planners, we give a flavour of \HTN~planners' requirements by taking a model of the well-known and overused domain of block world as described for each planner, and inspecting each model from two aspects. First, we take the same task of each domain model and analyse closely what needs to be encoded. Second, we give a broader view of each domain model by quantifying its content with respect to knowledge symbols, keyword symbols, and domain elements.

\subsubsection{Expressiveness}
We tackle expressiveness from two perspectives. The first one is the formal expressiveness of \HTN~planning language, and requires formal semantics of the language that completely determines what the language can express. This issue has been a subject of discussion for some time, resulting in a number of studies on expressiveness of \HTN~planning, such as~\cite{erol1994:expressivity, kambhampati1995:comparative, nau1998:control, wilkins2001:call, lekavy2007:expressivity,erol1996:formalization}. The most comprehensive analysis is provided in~\cite{erol1995:complexity}, where the expressiveness of \HTN~planning language is analysed from a model-theoretic, operational and computational aspect. In each aspect, the expressiveness of \HTN~planning is compared to the one of \strips-like planning. Thus, we gain a perspective in theoretical expressiveness by summarising the findings in~\cite{erol1995:complexity}.\footnote{We assume that the reader is familiar with model-theoretic, operational and computational-based expressiveness.}

The second perspective is more practical, thus suggesting a relaxed definition of the expressive power. In this case, {\em expressiveness} of a language is determined by the breadth of what the language can represent and communicate. The breadth includes assessment of the description languages of \HTN~planners with respect to their formal system, preference support, etc. Unfortunately, there is no common planning language for \HTN~planners. The idea of standardising a planning language is introduced with the Planning Domain Definition Language (PDDL)~\cite{ghallab1998:pddl} in 1998 for the purpose of the International Planning Competition (IPC), rather late with respect to the history of, above all, \pHTN planners. Although in the first version of PDDL there was an attempt to formalise a common syntax compatible to \HTN~planners, the idea was discarded with version 2.1 of PDDL~\cite{fox2003:pddl21} due to the immense differences between planners. 

Since we are still interested in what \HTN~planners can express in practice, we explore the language expressiveness of each \HTN~planner from three aspects. First is the use of first-order logic, where we want to determine the support for different logical operators. The second aspect is on the use of quality constraints, where we want to determine the support for sort hierarchy, extended goals, and preferences. We define each as follows. A {\em sort hierarchy} expresses characteristics of an object in a type hierarchy (similar to typing in PDDL). The basic idea of {\em extended goals} is to express a planning objective in a way that its satisfaction could be on a part or on the whole trajectory of the plan, and not just in the final state. A {\em preference} is a condition on the plan trajectory that some user would prefer satisfied rather than not satisfied, but would accept if the condition might not be satisfied~\cite{gerevini2006:pddl3}.

\begin{figure} [!ht] 
\centering
\includegraphics[width=.5\textwidth]{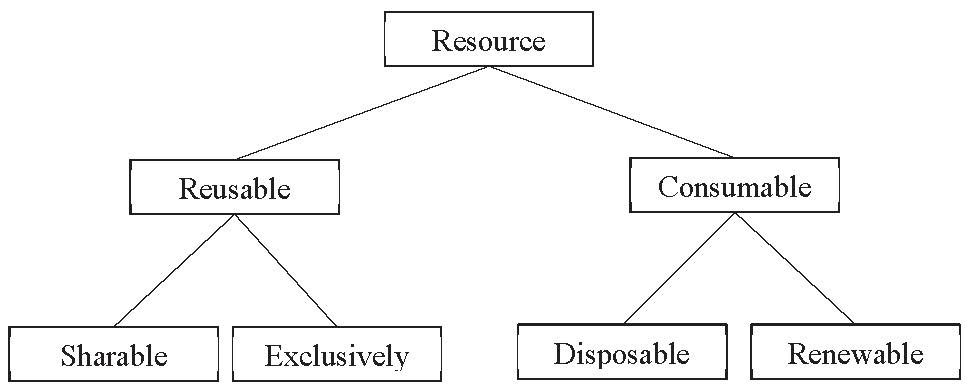}
\caption{Resource Hierarchy}\label{fig:resources}
\end{figure}

The third aspect is on the use of scheduling, where we want to determine the support for resource and time management. By {\em resource} we mean an object 
of limited availability within a planning problem. We define and consider when analysing the planners a resource hierarchy, as shown in Figure~\ref{fig:resources}. Each element of the hierarchy is defined as follows. A reusable resource is a resource that can be used again after its initial use. A shared reusable resource can be shared among several tasks at the same time, while an exclusive reusable resource cannot be used by two tasks in parallel. A consumable resource is a resource that is usable only a limited number of times. A consumable resource can be replenished or not. If the resource cannot be restored after the use of the set amount, it is called disposable consumable resource. Otherwise, if the resource amount can be topped up, it is called renewable consumable resource. Time is considered as usual, that is, a consumable resource that cannot be reproduced. We are interested in how \HTN~planners acquire and handle temporal information. In the simplest case, we are curious about three explicit temporal forms of a task, that is, the start of a task, the end of a task, and the duration of a task.

\subsubsection{Competence} 
We define {\em competence} as the ability of an \HTN~planner to accomplish or address a specific feature, such as domain dependence, fault tolerance, completeness, and soundness:

\begin{itemize}
\item {\em Domain dependence} defines the class of a planner based on whether (and in what way) the planner can be configured to work in different domains~\cite{nau2007:trends}. The following classes exist: 1) domain-specific planners are composed for particular domain and probably unsuitable for any other domain; 2) domain-independent planners have a general planning engine that works in any domain, while the input is a planning problem to be solved; 3) domain-configurable planners are provided with a domain-independent planning engine, but the input is a domain-specific knowledge.
\item {\em Solution flexibility} defines the ordering of actions in the solution plan. We say that a solution is flexible if it is partially ordered.
\item {\em Complexity of the search mechanism} defines the parts that form the mechanism, and the easiness to adapt or modify the mechanism. This includes an algorithm with or without backtracking points (e.g., a choice among few tasks, a choice on variable bindings and task ordering), use of heuristics to help the search, search control interactively performed by a user, evaluation of preconditions, and flexibility to include other automatic algorithms.
\item {\em Fault tolerance} defines the property of a planner to recognise, react appropriately, and continue operating in the event of a failure. We are interested in monitoring plan executions in order to identify problems, and replanning by modifying the existing or generating a new plan in order to solve such problems.
\item Given an \HTN~planning problem, an \HTN~planner is {\em complete} if it always finds a solution to the planning problem when such a solution exists.
\item Given an \HTN~planning problem, an \HTN~planner is {\em sound} if every answer it gives is a correct solution to the planning problem. 
\end{itemize}

\subsubsection{Performance}
Similarly to expressiveness, we are interested in performance from two perspectives. The first perspective refers to the theoretical boundaries of \HTN~planning. This aspect is already addressed in~\cite{erol1995:complexity}, where time and space complexity of \HTN~planning are analysed. The computational problem, called plan existence, poses the following question: \textit{given an HTN planning problem, is there a plan that solves the problem?} In order to properly understand the results of analysis, the following possible settings should be taken into account.
\begin{itemize}
\item The sets of operators and methods can be provided in two ways. The sets can be a part of the {\em input}, or they can be {\em fixed} in advance, that is, the tasks are allowed to contain decompositions corresponding only to predicates in the initial state.
\item A compound task can be defined in several ways. A compound task is without any restriction ({\em yes}); a {\em regular} task in task networks, that is, at most one compound task followed by all primitive tasks; an {\em acyclic} task, that is, a task can be decomposed to only a finite depth; and, finally, compound tasks are not allowed at all ({\em no}).
\item A task network containing primitive and compound tasks as defined in the previous point can be {\em totally ordered} or {\em partially ordered}, whenever there is a total or partial order among those tasks, respectively.
\item Variables can be allowed or not in the planning problem.
\end{itemize}

The second perspective gives insights into the practice of testing of \HTN~planners. We are interested in the runtime and scalability results of each \HTN~planner. We say that a planning system is {\em scalable} if it is capable to cope and suitably perform under a varying size of planning problems. Anything but easy is to define dimensions that could measure the size of a problem, nevertheless, scalability is highly desirable in practical settings with an increasing and large number of facts about the state, a large number of users, or a large number of tasks. We are interested in how good planners scale relative to one another assuming increasingly difficult problems. As for {\em runtime}, we are interested in pairwise comparisons between \HTN~planners with respect to the amount of time they spend on the same sets of problems.

\subsubsection{Applicability} 
{\em Applicability} concerns the use of planners in real application domains. It appears to be orthogonal to previous categories. We have two reasons for the inclusion of this element in the framework. First, we strongly believe that the ultimate objective of research on automated planning must be application of planners in a variety of real domains: oil spills~\cite{agosta1996:oilspill}, spacecraft assembly~\cite{aarup1995:optimum}, microwave manufacturing~\cite{smith1997:electrical}, smart spaces~\cite{kaldeli2012:coorindating}, and Web service composition~\cite{kuter2005:gathering}, to name a few prominent examples. Second, \HTN~planning is promoted as the most applied automated planning technique of real-world problems~\cite{nau2005:applications}, however, mostly referring to the applications of \shop. Thus, we want to see 1) whether \HTN~planning contributes towards the aforementioned objective, and 2) what is the status of applicability of other \HTN~planners.

\subsection{Outcome}
For each element of the analytical framework, we explore the literature and apply descriptive research to understand the respective property. In two cases, we provide theoretical and practical perspectives of a framework element. Where possible, we also show comparison of \HTN~planners. In some cases, we aggregate the data on planners in tabular form. We use the following common notation for all tables.  The `\ding{51}' denotes that a planner supports the respective property, the `\ding{53}' indicates that a planner does not support a particular property, and an empty cell denotes that it is unknown from the literature whether the planner is able to deal with a given element. There are rated properties, where the rate ranges from `\ding{72}', denoting limited support for the given property, to `\ding{72}\ding{72}\ding{72}', indicating extended support.

\subsubsection{Domain Authoring}
Figure~\ref{fig:puton} shows the description of the `put-on' task provided to each \HTN~planner except \siadex~for which we were not able to find neither a task nor a domain description. We start by describing \noah's task which specifies that three statements should be evaluated in sequence (line~\ref{lst:noah-1},~\ref{lst:noah-4} and~\ref{lst:noah-5}). The first statement is handled by evaluating lines~\ref{lst:noah-2} and~\ref{lst:noah-3}, which cause new tasks to be added. If a predicate (e.g., `(cleartop \$x)') is not true, we should expect that a task would be added to achieve the predicate, otherwise some form of a `do-nothing' is assumed. The statement in line~\ref{lst:noah-4} is evaluated analogously. For the last statement in line~\ref{lst:noah-5}, we need to be aware that it deletes a predicate from the (global) state, but also inserts that predicate in the add list of some dummy task at the current level (of planning). 

\begin{figure*} [!ht]
\centering
\small
\begin{tabular}{ll|ll}
\toprule 
{\bf Planner} & {\bf Description} & {\bf Planner} & {\bf Description} \\ \hline
{\bf \noah} & \begin{lstlisting}
(puton
  (qlambda
    (on <-X <-Y)
@\label{lst:noah-1}@     (pand
@\label{lst:noah-2}@         (pgoal (clear $x) (cleartop $x)
              apply (clear))
@\label{lst:noah-3}@         (pgoal (clear $y) (cleartop $y)
              apply (clear))
      )
@\label{lst:noah-4}@     (pgoal (put $x on top of $y)
             (on $X $y) apply nil)
@\label{lst:noah-5}@     (pdeny (cleartop $y))
  )
) 
\end{lstlisting} & {\bf \nonlin} & \begin{lstlisting}
actschema puton
  pattern <<put $*x on top of $*y>>
@\label{lst:nonlin-1}@  conditions 
    holds <<cleartop $*x>> at self
    holds <<cleartop $*y>> at self
@\label{lst:nonlin-2}@    holds <<on $*x $*z>> at self
@\label{lst:nonlin-3}@  effects 
    + <<on $*x $*y>>
    - <<cleartop $*y>>
    - <<on $*x $*z>>
@\label{lst:nonlin-4}@    + <<cleartop $*z>>
  vars x undef y undef z undef;
end;
\end{lstlisting} \\ \hline
{\bf \sipe} & \begin{lstlisting}
operator: puton
arguments: block1, object1 is not block1;
@\label{lst:sipe-1}@purpose: (on block1 object1);
@\label{lst:sipe-2}@plot:
@\label{lst:sipe-3}@parallel
  branch 1: goals: (clear object1);
  branch 2: goals: (clear block1);
end parallel
@\label{lst:sipe-4}@process
action: puton.primitive;
arguments: block1, object1;
resources: block1;
effects: (on block1 object1);
end plot end operator
\end{lstlisting} & {\bf \oplan} & \begin{lstlisting}
schema puton;
 vars ?x=undef, ?y=undef, ?z=undef;
@\label{lst:oplan-1}@ expands {put ?x on top of ?y};
@\label{lst:oplan-2}@ only_use_for_effects
    {on ?x ?y}=true,
    {cleartop ?y}=false,
    {on ?x ?z}=false,
    {cleartop ?z}=true;
@\label{lst:oplan-3}@ conditions
    only_use_for_query {on ?x ?y}=true,
@\label{lst:oplan-4}@    achievable {cleartop ?y}=true,
@\label{lst:oplan-5}@    achievable {cleartop ?x}=true;
endschema;
\end{lstlisting} \\ \hline
{\bf \umcp} & \begin{lstlisting}
(operator puton (x y)
   :pre ((clear x)(on x table)(clear y))
   :post ((~on x table)(on x y)(~clear y))
)
\end{lstlisting} & {\bf \shop} & \begin{lstlisting}
(:operator (!puton ?x ?y)
@\label{lst:shop-1}@  ((clear ?x) (on-table ?x) (clear ?y))
@\label{lst:shop-2}@  ((clear ?y) (on-table ?x))
@\label{lst:shop-3}@  ((on ?x ?y))
)
\end{lstlisting}\\ \hline
{\bf \siadex} & \begin{lstlisting}
(:action puton
@\label{lst:siadex-1}@ :parameters (?x ?y - block)
@\label{lst:siadex-2}@ :precondition (and (grasping ?x)(clear ?y))
@\label{lst:siadex-3}@ :effect (and (not (grasping ?x))
         (not (clear ?y))(clear ?x)
         (on ?x ?y)(handempty))
)
\end{lstlisting} & & \\
\bottomrule
\end{tabular}
\caption{The `put-on' Block-World Task in \HTN~Planners}\label{fig:puton}
\end{figure*}

In contrast to \noah, the encoding of \nonlin~is more clear. In \nonlin, the `put-on' task contains three filter conditions (line~\ref{lst:nonlin-1} to~\ref{lst:nonlin-2}). The first two filter conditions state that before applying the task two blocks  must be clear. The third filter condition is used to bind some variable with what is on top of the variable from line~\ref{lst:nonlin-1}. If all of these conditions hold, the effects of the task are applied (line~\ref{lst:nonlin-3} to~\ref{lst:nonlin-4}). With \sipe~things are getting complicated again. Its task contains the following expressions of interest. Line~\ref{lst:sipe-1} specifies the goal that this task can achieve. In the element of line~\ref{lst:sipe-3}, a task network of two parallel tasks is contained with the purpose of achieving some predicates. The element in line~\ref{lst:sipe-4} specifies the action that should be used in order for the `put-on' task to be accomplished. The task contains no information about when a block is clear or not, or when a block is not on another block. This information would be inferred by planner's deductive theory. \oplan's task is very similar to the one of \nonlin. Line~\ref{lst:oplan-1} specifies the task network used to accomplish the task, that is, a single action. The expressions in line~\ref{lst:oplan-2} to~\ref{lst:oplan-3} are the actual effects of the action. On the other hand, three conditions must be satisfied in order for these effects to be applied. The conditions in line~\ref{lst:oplan-3} and~\ref{lst:oplan-4} might be achieved by other tasks during the planning process. The condition in line~\ref{lst:oplan-5} is used to bind a variable for the purpose of specifying the effects.

The task descriptions for \umcp, \shop~and \siadex~differ only in the notation, but specify the same meaning. All tasks, that is, operators contain simple applicability preconditions (line~\ref{lst:shop-1} in \shop), and effects (line~\ref{lst:siadex-3}) as a postcondition in \umcp, and as a delete and an add list in \shop~(line~\ref{lst:shop-2} and line~\ref{lst:shop-3}, respectively). However, beside the representational simplicity, the power of these operators is much weaker than the tasks of the previously discussed planners. The operators cannot handle situations where some block is above another one or when a block is not clear. The solution to achieve fairly equal functionality is to write methods that describe all possible situations. 

\begin{figure*} [!ht]
\centering
	\begin{subfigure}[b]{0.3\textwidth}
	\centering
	\includegraphics[width=\textwidth]{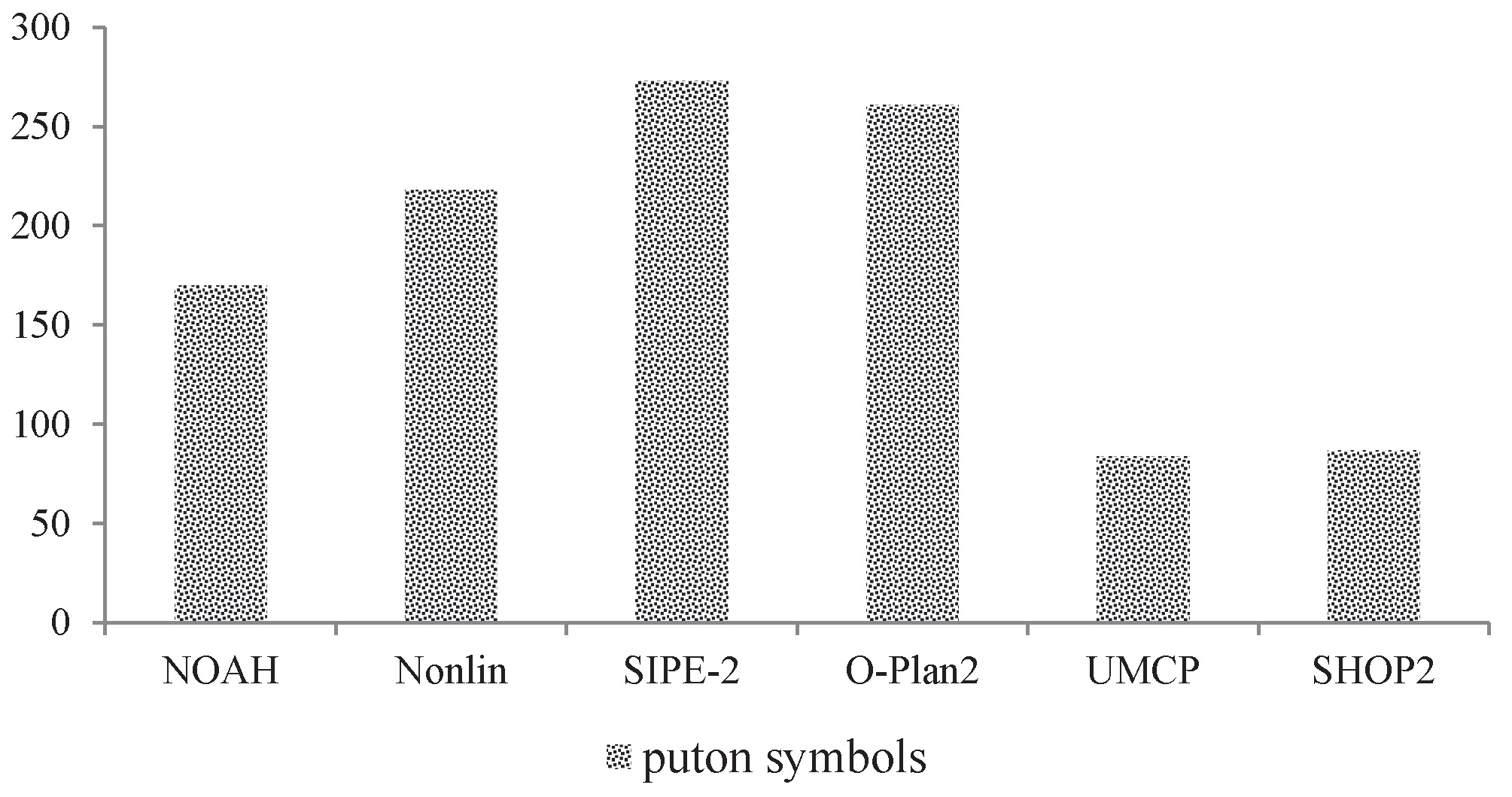}
	\caption{Number of \textit{put-on} Symbols}
	\label{fig:size-puton}
	\end{subfigure}
	~
	\begin{subfigure}[b]{0.3\textwidth}
		\centering
		\includegraphics[width=\textwidth]{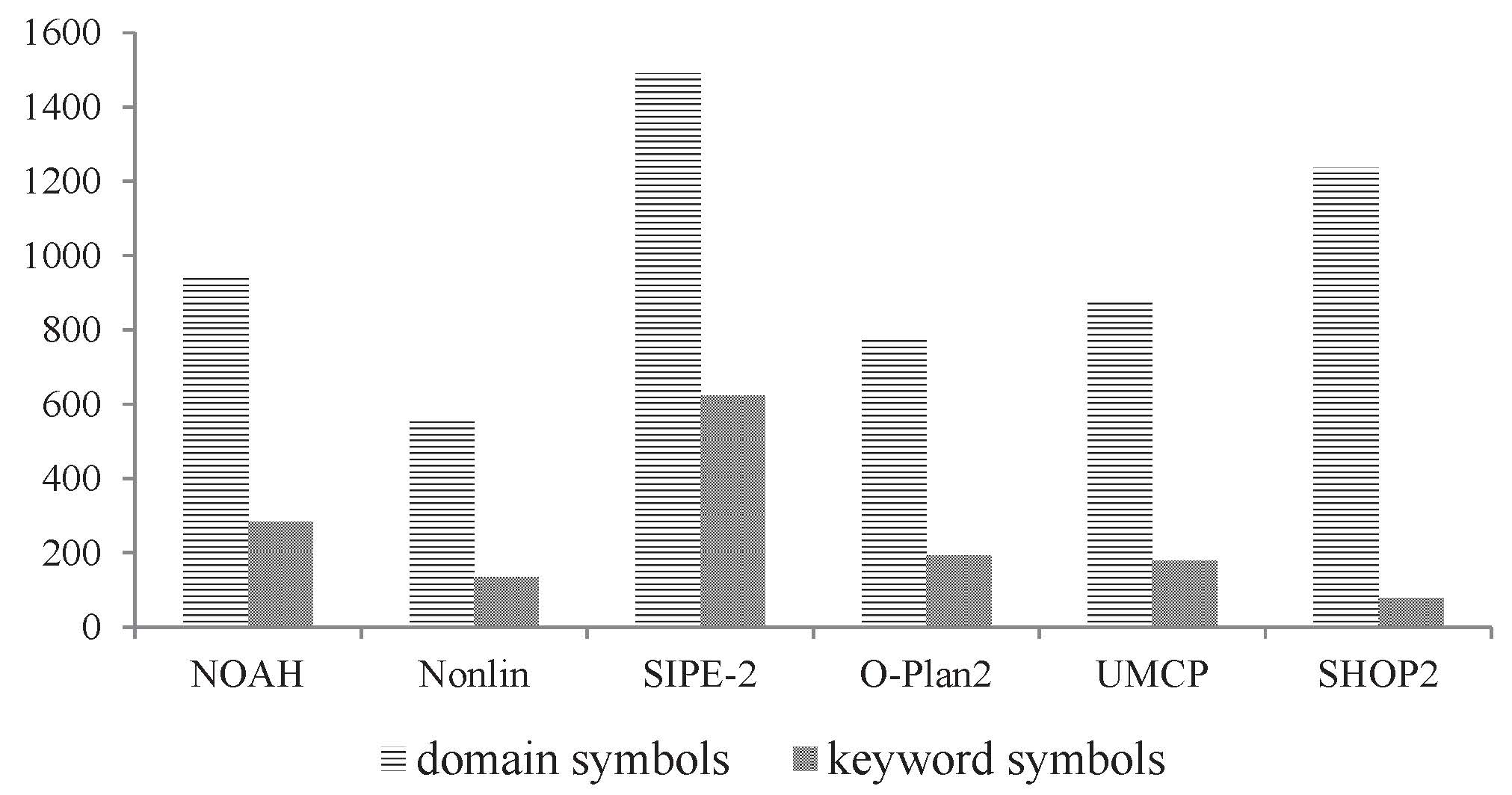}
		\caption{Number of Domain Symbols}
		\label{fig:size-domain}
	\end{subfigure}
	~	
	\begin{subfigure}[b]{0.3\textwidth}
	\centering
	\includegraphics[width=\textwidth]{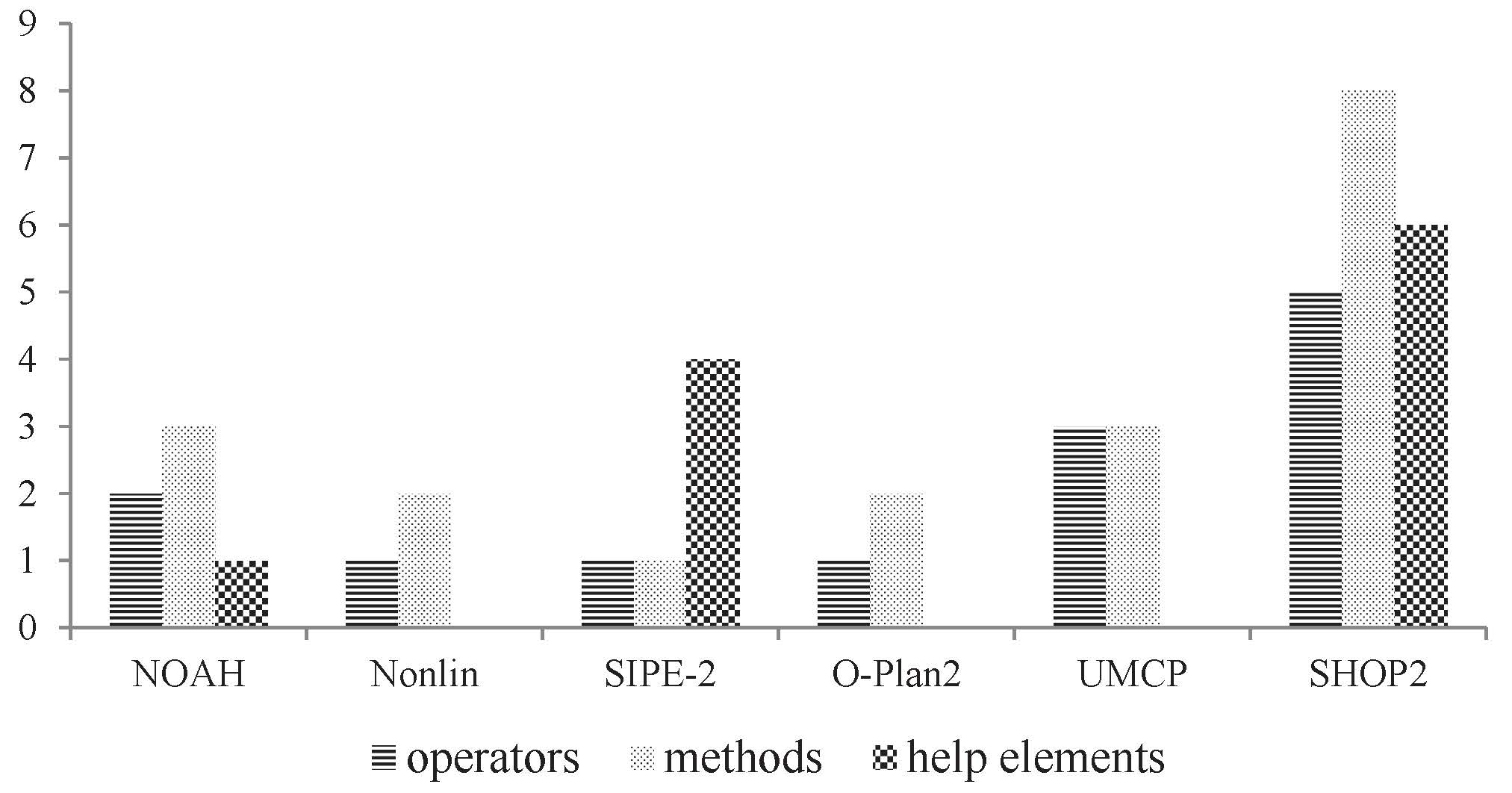}
	\caption{Number of Domain Elements}
	\label{fig:number-domain}
	\end{subfigure}
\caption{Quantitative Perception of the `put-on' Task Description and Block-World Domain}\label{fig:sizes}
\end{figure*}

Figure~\ref{fig:size-puton} gives a quantitative perception of the size of the `put-on' task descriptions from Figure~\ref{fig:puton} in terms of the number of symbols. At first glance, \umcp~and \shop~seem to require smaller `put-on' descriptions, but looking at the number of symbols at the domain level, as shown in Figure~\ref{fig:size-domain}, \sipe~has the largest domain description, however, almost half of it belongs to keyword symbols. On the other hand, \shop~has slightly smaller domain description than \sipe, but the number of keyword symbols is negligible, which means that the rest of the symbols represent the actual domain knowledge. Figure~\ref{fig:number-domain} makes it even more clear. \umcp~and \shop~need more tasks in total than the rest of \HTN~planners in order to successfully perform planning in the domain. In fact, \shop~needs 13 operators and methods in total, and 6 axioms to do planning, while \oplan, for example, needs only three tasks in total. 

\subsubsection*{Interpretation}
The analysis of tasks shows, on the one hand, that a prerequisite to author domain knowledge for most \pHTN planners is the comprehension of their underlying systems, such as expectations of what the system would do in a particular situation. On the other hand, a domain author does not need to have a priori familiarity with a \tHTN planner, but the author must provide quite powerful and elaborate domain configuration. This is supported by the analysis of results shown in Figure~\ref{fig:sizes}. Additional evidence to the latter observation is the criticism of \shop~planner that it is a problem-solving programming language rather than a planner~\cite{schattenberg2009:hybrid}. Finally, Figure~\ref{fig:size-domain} indicates the richness of domain descriptions with knowledge.

\subsubsection{Expressiveness}\label{sec:expressiveness}

\subsubsection*{Expressiveness in Theory}
\begin{figure} [!ht]
\centering
	\begin{subfigure}[b]{0.5\textwidth}
	\centering
	\includegraphics[width=0.5\textwidth]{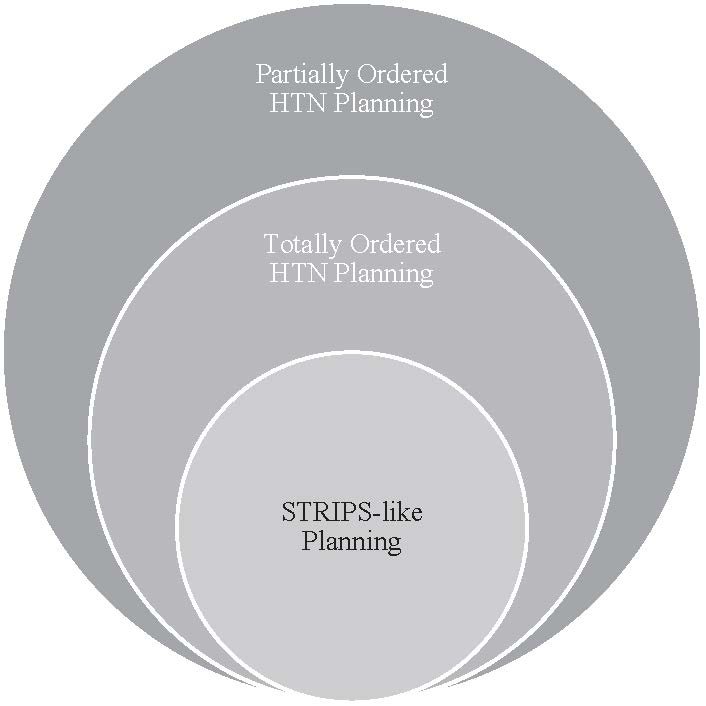}
	\caption{Hierarchy of model-theoretical and operational expressiveness}
	\label{fig:mte}
	\end{subfigure}
	\qquad
	\begin{subfigure}[b]{0.45\textwidth}
	\centering
	\includegraphics[width=0.5\textwidth]{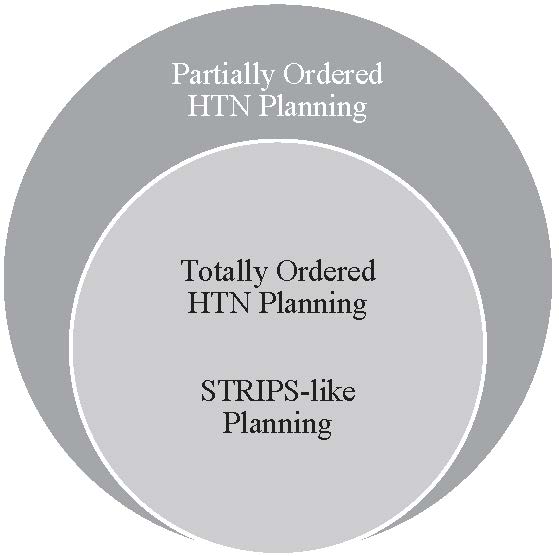}
	\caption{Hierarchy of computational-based expressiveness}
	\label{fig:cce}
	\end{subfigure}
\caption{Hierarchies of expressiveness}\label{fig:expr-hierarchy}
\end{figure}

Figure~\ref{fig:mte} depicts the model-theoretic expressiveness. From this aspect, the \HTN~language is strictly more expressive than the \strips~language, but totally ordered \HTN~planning is less expressive than partially ordered \HTN~planning, and strictly more expressive than \strips-like planning~\cite{nau1998:control}. An \HTN~planning problem (with totally or partially ordered task networks) can be transformed into a \strips-like planning problem, but the converse is not true. On the other hand, a totally ordered \HTN~planning problem can be transformed into a partially ordered \HTN~planning problem, but the converse is not true.

Figure~\ref{fig:mte} shows that the operational aspect has the same expressiveness hierarchy as the model-theoretic aspect. That is, \HTN~planning is strictly more expressive than \strips-style planning, and totally ordered \HTN~planning is strictly between \strips-like planning and partially ordered \HTN~planning.

Figure~\ref{fig:cce} depicts the computational-based hierarchy. Once more, \HTN~planning is strictly more expressive than \strips-like planning. In particular, there is a (polynomial) transformation from \strips-like planning to \HTN~planning, but there is no computable transformation from \HTN~planning to \strips-like planning. Intuitively, \HTN~elements can represent computationally more complex problems than \strips-like operators. However, these results are true when partially ordered task networks are allowed. In fact, totally ordered \HTN~planning is at the same level of expressiveness as \strips-like planning, but significantly less expressive than partially ordered \HTN~planning. This is because totally ordered \HTN~planning avoids interleaving of tasks from different compound tasks.

\subsubsection*{Expressiveness in Practice}
Table~\ref{tab:expr} illustrates the expressiveness properties of the planning languages of \HTN~planners. \sipe, \umcp, \shop~and \siadex~employ first-order logic in their preconditions with some restrictions, while disjunction is not allowed in the effects of tasks. Thus, task preconditions are more expressive than the task effects. Except \nonlin~and \shop, which use deletion of a predicate, other \HTN~planners use negation of a predicate in the effects. \shop's language supports about the level 2 of the PDDL version 2.1, and allows declarations of implications in the task preconditions and universally quantified in the task preconditions and effects.

\begin{table*} \footnotesize
\centering
\caption{Practical Expressiveness of \HTN~Planners}\label{tab:expr}
\begin{tabular}{lccccccc}
\toprule 
{\bf Property} & {\bf \noah} & {\bf \nonlin} & {\bf \sipe} & {\bf \oplan} & {\bf \umcp} & {\bf \shop} & {\bf \siadex} \\ \hline \hline
{\bf Conjunction} & \ding{51} & \ding{51} & \ding{51} & \ding{51} & \ding{51} & \ding{51} & \ding{51} \\ 
{\bf Disjunction} & preconditions & \ding{53} & preconditions & constraints & preconditions & preconditions & preconditions \\ \cline{2-8} 
\multirow{3}{*}{\bf Negation} &  &  &  &  & preconditions & & \\ 
& effects & & \parbox{3cm}{\centering
preconditions \\ effects} & effects & constraints & preconditions & preconditions \\
& & &  & & effects & & \\ \cline{2-8}
{\bf Implication} & \ding{53} & \ding{53} & \ding{53} & \ding{53} & \ding{53} & preconditions & \\ \cline{2-8} 
\multirow{2}{*}{\bf Existential Q.} &  &  & preconditions &  & \multirow{2}{*}{constraints} & & \\ 
 & & & conditions & & & & \\ \cline{2-8}
\multirow{2}{*}{\bf Universal Q.} & \multirow{2}{*}{\ding{53}} & \multirow{2}{*}{\ding{53}} & \multirow{2}{*}{effects} & \multirow{2}{*}{\ding{53}} & \multirow{2}{*}{\ding{53}} & preconditions & preconditions \\
 & & & & & & effects & effects \\
 \hline \hline
{\bf Sort Hierarchy} & \ding{53} & \ding{53} & \ding{72}\ding{72}\ding{72} & \ding{72} 
& \ding{53} & \ding{53} & \\ 
{\bf Extended Goals} & \ding{53} & \ding{53} & \ding{53} & \ding{53} & \ding{53} & \ding{53} & \ding{72} \\ 
{\bf Preferences} & \ding{53} & \ding{72} & \ding{72}\ding{72} & \ding{72} & \ding{72} & \ding{72}\ding{72}\ding{72} & \ding{53} \\ \hline \hline
{\bf Resource} & \ding{53} & \ding{53} & \ding{72}\ding{72} & \ding{72}\ding{72}\ding{72} & \ding{53} & \ding{53} & \ding{53} \\
{\bf Time} & \ding{53} & \ding{53} & \ding{72} & \ding{72}\ding{72}\ding{72} & \ding{53} & \ding{72} & \ding{72}\ding{72}\ding{72} \\ 
\hline 
\bottomrule
\end{tabular}
\end{table*}

The planning language of \sipe~supports definition of sort hierarchy. Some variable can be of a specific class in the sort hierarchy, for example, `object1 of type light'. A constraint can define that some variable must not be a member of a given class, for example, `light1 is not of type dimmable'. \oplan~also employs a certain level of sort hierarchy. A specification of one or more classes of objects is allowed, which are then used within a variable declaration to determine the possible set of variable bindings. For example, the classes `types object=(b1 b2 truck plane), box=(b1 b2), vehicle=(truck plane)' could be defined to further constrain the type of the variable `?b=?\{type box\}'.

Preferences in \pHTN planners are introduced by Myers in~\cite{myers1996:advice, myers2000:conflicting}. The author provided a formal language and semantics for preferences, and an algorithm implemented on top of \sipe. In particular, the planner accepts a preference that may condition the desired solution and the decisions steps taken during the planning process. A preference can impose prescriptions and prohibitions on characteristics of interest for some task, on the capacity in which some object is to be used in the task, and on objects. What is interesting in this approach is that there are two forms of expressing preferences. The first form prescribes or prohibits the use of some objects when filling certain capacities in some task, while the second one prescribes or prohibits the use of a particular task when accomplishing some objective. For example, the following form expressed in natural language, ``stay in a 5-star hotel while on holiday in The Netherlands'', introduces preferences on accommodation during holiday in a given country. On the other hand, the following form also expressed in natural language, ``find a package bike tour starting in Groningen for the holiday in The Netherlands'', indicates that the approach taken to satisfy a particular portion of the trip should have certain characteristics, such as `bike' and `package', and capacity, such as `start location is Groningen'.

\oplan~allows expressing simple preferences on plans and tasks~\cite{drabble1995:tfmanual}. Plan preferences provide heuristic information to the planner, but it is not clear how is this information actually used, while task preferences provide information about the order in which some tasks can be decomposed, or used to handle an `achieve' condition.

From a \tHTN perspective, \shop~explicitly supports only hard constraints in preconditions, that is, constraints that must be satisfied by all found plans. However, there are studies that propose languages for specifying preferences and techniques that can take these constraints into account. In particular, the language and approach proposed in~\cite{sohrabi2008:onpreferences} is applied over \shop. The language supports three types of preferences. The first type are basic constructs of Linear Temporal Logic (LTL). The second type are soft constrains, such as the precedence constraint \texttt{before(t,t')} and the state constraints \texttt{holdBefore(t,q)}, where \texttt{t} and \texttt{t'} are tasks and \texttt{q} is a predicate. The third and most interesting type are the preferences over how tasks are decomposed into task networks (e.g., prefer to apply a certain method over another), preferences over the parametrisations of decomposed tasks (e.g., preferring one task grounding to other\footnote{For the sake of completeness, \shop~supports a similar feature called `sort-by' that sorts variable bindings according to some criteria.}), and a variety of temporal and non-temporal preferences over task networks. This approach is later adjusted to comply with the syntax of preferences in the version 3 of PDDL~\cite{gerevini2006:pddl3,sohrabi2009:preferences}. Preferences over task occurrences and task decompositions are enabled by using several modal operators, which are allowed within simple and temporally extended preferences and constraints, but not within precondition preferences. A modal operator can indicate that 1) some primitive task may occur in the current state; 2) some task (or task decomposition) is initiated in the current state; or 3) some task (or task decomposition) is terminated in the current state.

\sipe~and \oplan~support our resource hierarchy completely. In addition to this hierarchy, \oplan~allows sharing reusable resources unary, where a sharable resource cannot be shared among many tasks at the same time, or simultaneously, where a resource can be shared among many tasks without any specific control. In \sipe, renewable resources can be topped up by tasks, while in \oplan, it can be done by tasks, some so-called off-line processes or a combination of both.  

\sipe~offers a limited mechanism for temporal reasoning, but full support is enabled by using an external temporal reasoning system called Tachyon~\cite{stillman1993:tachyon}. Tasks may include temporal constraints in three ways~\cite{wilkins2000:sipe2_manual}. First, a task may implicitly define simple ordering constraints between tasks in its task network. These constraints indicate `before' or `meets' relationships in terms of relations defined independently by Allen~\cite{allen1983:temporal} and van Benthem~\cite{benthem1991:logictime}. Second, explicit temporal constraints can be added to the description of a task to express any of the thirteen Allen relations between its tasks. Last, temporal constraints can be exposed on any task by using some of the temporal classes, such as `earliest-start', `latest-end', or `shortest-duration', and some integer value to denote the time of the task. For example, if a task has assigned the expression `earliest-start.3', then the ``global'' task network containing this task must respect the given earliest start time (that is 3) for this task.

\oplan~maintains a time point as a numerical pair (min,max) which denotes the upper and lower bound of any time~\cite{drabble1995:tfmanual}. A time point can record the start and end time of an action, the duration of an action, and delays between actions. The actions of the solution take a temporal class similar to those used by \sipe.

\shop~does not explicitly reason about time. However, there are three attempts to extend the planner to handle durative actions. The first attempt tries to produce plans with parallel actions based on concurrent update rules for numeric state variables~\cite{yaman2002:timeline}. The value of such a variable can be decreased or increased by some amount. This approach is restricted in terms of the number of relations in Allen's algebra, and becomes inefficient when the problem size and concurrency level increases. The second approach uses a preprocessing technique that translates durative actions of level 3 in version 2.1 of PDDL~\cite{fox2003:pddl21} into \shop~operators to bookkeep the temporal information~\cite{nau2003:shop2}. Parallelism is enabled only in situations where the actions begin at the same moment. The proposed algorithm is incomplete too because it avoids certain effects, axioms, and violates some ordering constraints. The last approach also enables encoding PDDL durative actions for \shop, where every durative action corresponds to a method composed of a `start' and an `end' operator~\cite{goldman2006:durative}. The parallelism of tasks is achieved by interleaving the start and end operators. Unfortunately, this approach also shows disappointing performance when reasoning about time in large problems.

\siadex~uses three sources of temporal information~\cite{castillo2006:handling}. The first one comes from the ordering of tasks in each task network. The planner supports three types of such ordering constraints: 1) two tasks must be executed in the total order, 2) tasks can occur in parallel, and 3) tasks may occur in any of the total orders given by their permutation. The second source is the causal structure of actions in which every action is associated with two time points, `start' and `end'. Since the state is also temporally annotated, it is known which action produced some predicate and at what time it was accomplished. This information is then used to propagate constraints between tasks during the planning process and to produce flexible (sub)plan. The last source are deadline goals and complex synchronisations. The former expresses a goal that must be achieved at a particular time point, while the latter expresses constraints that allow actions to interact along a period of time.

\subsubsection*{Interpretation}
From the theoretical perspective, we can conclude that \HTN~planning is able to express a broader and more complex set of planning problems than \strips-like planning. However, this statement is controversial since the assumption is that the theoretical model of \HTN~planning uses an infinite set of symbols to represent tasks. But in reality, this model cannot be supported by any planner unless some restrictions are imposed~\cite{lekavy2007:expressivity}.

From the practical perspective, we may say that both categories of \HTN~planners are able to address similar expressiveness. It appears that planners and their corresponding category still have some challenges to address. For example, first-order logic is not fully supported by most planners, and the quality constraints are only partially implemented.

\subsubsection{Competence}
Table~\ref{tab:competence} summarises the properties related to the competence of \HTN~planners. We begin with the property of domain dependence, where all state-of-the-art \HTN~planners fit into the category of domain-configurable planners. The next property is on the flexibility of plans and their execution. The result of the planning process in \pHTN planners is a partially ordered plan which is in compliance with the definition of flexibility. An exception to this outcome is the \umcp~planner which restricts the actions of the solution to be totally ordered. With respect to \tHTN planners, there are two cases as well. \shop~produces a totally ordered plan, while \siadex~is able to plan for a more flexible output.

\begin{table*}\footnotesize 
\caption{Competence of \HTN~Planners}\label{tab:competence}
\centering
\begin{tabular}{llccccccc}
\toprule 
\multicolumn{2}{l}{\bf Property} & {\bf \noah} & {\bf \nonlin} & {\bf \sipe} & {\bf \oplan} & {\bf \umcp} & {\bf \shop} & {\bf \siadex} \\ \hline \hline
\multicolumn{2}{l}{\bf Domain Dependence} & \multicolumn{7}{|c|}{Domain-configurable}\\ \cline{3-9}
\multicolumn{2}{l}{\bf Solution Flexibility} & \multicolumn{4}{|c|}{Partial} & \multicolumn{2}{|c|}{Total} & \multicolumn{1}{c|}{Partial} \\ \cline{3-9}
\multirow{7}{*}{\bf Search} & {\bf Automatic Algorithm} & DFS & BFS & DFS & HS & DFS & DFS & DFS \\ \cline{5-9} 
& {\bf Backtracking} & \ding{53} & DD & \multicolumn{5}{|c|}{Chronological}\\ \cline{5-9}
& {\bf Heuristics} & \ding{51} & \ding{51} & \ding{51} & \ding{51} & \ding{53} & \ding{53} & \ding{53} \\
& {\bf Domain-Specific Control} & \ding{53} & conditions & preconditions & conditions & conditions & preconditions & preconditions \\ 
& {\bf Interactive Control} & \ding{53} & \ding{53} & \ding{51} & \ding{51} & \ding{51} & \ding{53} & \ding{51} \\ 
& {\bf Other} & CM & CM & CM & CM & CM & \ding{53} & CM \\ \cline{2-9}
& {\bf Alternative Algorithms} & \ding{53} & \ding{53} & \ding{53} & \ding{53} & BrFS, BFS & IDS & \ding{53} \\ \hline \hline
\multicolumn{2}{l}{\bf Execution Monitoring} & \ding{72} & \ding{53} & \ding{72}\ding{72}\ding{72} & \ding{72}\ding{72}\ding{72} & \ding{53} & \ding{53} & \ding{72}\ding{72}\ding{72} \\
\multicolumn{2}{l}{\bf Replanning} & \ding{72} & \ding{53} & \ding{72}\ding{72}\ding{72} & \ding{72}\ding{72}\ding{72} & \ding{53} & \ding{53} & \ding{72}\ding{72}\ding{72} \\ \hline \hline
\multicolumn{2}{l}{\bf Completeness} & \ding{53} & \ding{51} & \ding{53} & \ding{51} & \ding{51} & \ding{51} & \\ 
\multicolumn{2}{l}{\bf Soundness} & \ding{53} & \ding{53} & \ding{53} & \ding{53} & \ding{51} & \ding{51} & \\
\bottomrule
\end{tabular}
\end{table*}

The next observation is on the search mechanism of planners. Except for \nonlin~and \oplan, all planners implement depth-first search (DFS) as their main algorithm, however, not all of them backtrack to all alternative points. Since backtracking and decisions on variable bindings and task ordering is covered in Section~\ref{sec:plan-review} and Section~\ref{sec:state-review}, we do not go in details here. \noah~does not backtrack at all, but relies on its heuristics and choosing the ``correct'' task, binding or ordering among few possibilities, and ignores the rest of them. \nonlin~uses breadth-first search (BrFS) with dependency-directed (DD) backtracking, that is, it backtracks on choices of variable bindings and choices of task orderings. \sipe~backtracks chronologically and uses heuristics to limit backtracking points to alternative tasks, allows variable binding choices only on two places during the planning process, and does not backtrack the addition of ordering constraints. \oplan~uses heuristic search over its choices of tasks in the plan space, where an evaluation function based on the opportunistic merit of the state is used. \oplan~uses an alternative manager to manage alternatives it is provided with and to seek alternatives when no results are provided by other components. It appears that both planners, \oplan~and \sipe, use resource reasoning to improve their search, but another observation says that resource reasoning is used to prune the search space in \oplan, while it is used to enhance the efficiency of \sipe~\cite{kartam1989:foundation}. Almost all planners use their constraint management (CM) to handle constraints during the search for a solution.

Beside depth-first search, \umcp~offers a choice of two more algorithms, namely breadth-first and best-first search (BFS). The planner is able to backtrack to all decision points, and along the chosen algorithm, it uses its resolution method, and domain-specific knowledge to successfully search for a solution. \shop~tries all possible alternatives, either variable bindings or method choices. With respect to algorithm flexibility, the planner is provided with iterative-deepening search (IDS) as well. The authors claim that incorporation of any other search techniques is simple enough too, meaning that the structure of the planner is fairly flexible~\cite{nau2003:shop2}. As \shop, \siadex~uses depth-first search too, but this planner is heavily dependent on constraint-based operations in order to successfully search for a plan.

Interaction with users throughout the planning process can be used to help the search. User interaction addresses some issues that are beyond the capabilities of the algorithms. Among \HTN~planners, \sipe, \oplan~and \umcp~provide user interfaces for guiding purposes, while SHOP2 (in fact, its Java version) offers an interaction interface only for debugging purposes.

Backtracking points ensure completeness. Since \noah~does not backtrack to any alternative point, the planner is obviously incomplete. \nonlin~is complete under the constraints defined in the domain description. Moreover, if all conditions in the domain description are set to `achieve', then the planner might be complete.\footnote{Austin Tate, personal communication, November 23, 2012.} \oplan~shares the property of \nonlin, whereby the planner is complete under the requirements provided in the domain description. In other words, the planner guarantees to produce at least one valid solution if such a solution exists within the provided domain description and within the capabilities of the constraint managers. Due to extensive use of inadmissible and vaguely-defined heuristics, \sipe~does not guarantee that a solution will be found, although a solution may exist. This statement is even practically confirmed in~\cite{gomes1994:comparison}, where the results of experiments show that \sipe~failed to solve feasible problems, while \oplan~solved all of them. Finally, \umcp~and \shop~are provably complete and sound planners~\cite{erol1994:umcp}. The completeness of these planners is defined with respect to both the operators, as well as the set of compound tasks.

\begin{figure} [!ht]
\centering
	\begin{subfigure}[b]{0.3\textwidth}
	\centering
	\includegraphics[width=\textwidth]{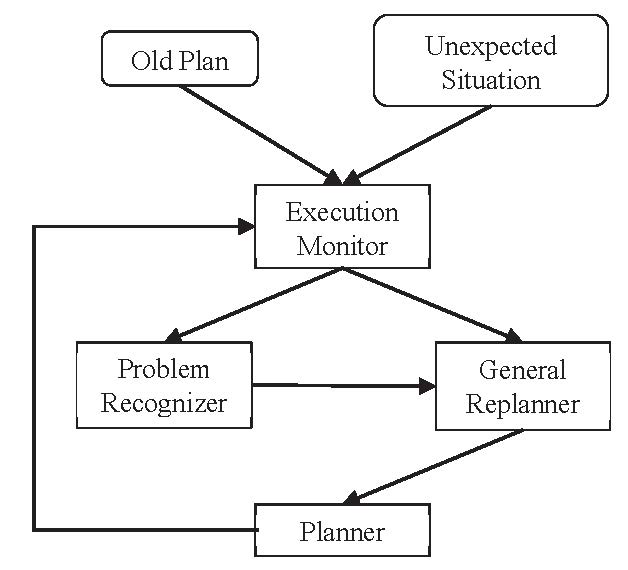}
	\caption{System in \sipe}
	\label{fig:em-sipe}
	\end{subfigure}
	~ ~
	\begin{subfigure}[b]{0.4\textwidth}
	\centering
	\includegraphics[width=\textwidth]{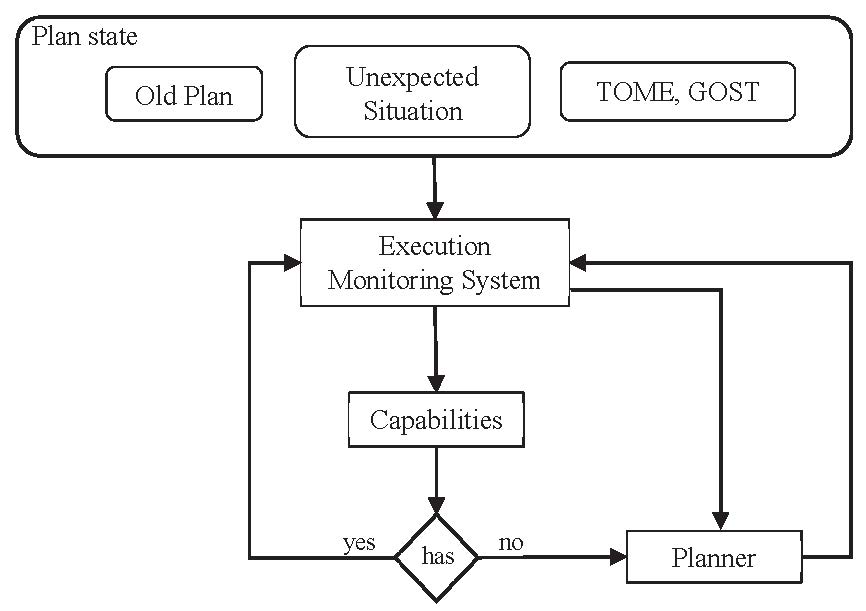}
	\caption{System in \oplan}
	\label{fig:em-oplan}
	\end{subfigure}
	~ ~
	\begin{subfigure}[b]{0.25\textwidth}
	\centering
	\includegraphics[width=\textwidth]{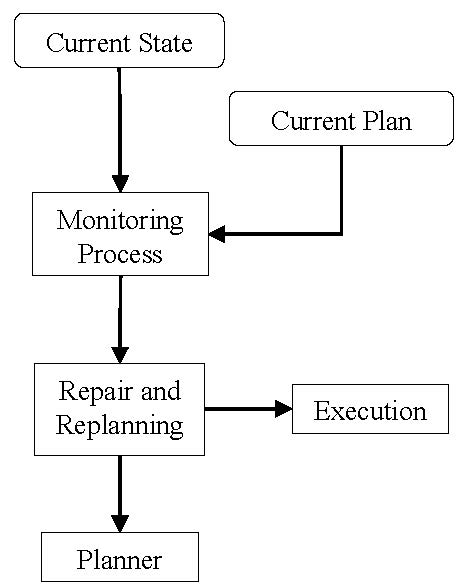}
	\caption{System in \siadex}
	\label{fig:em-siadex}
	\end{subfigure}
\caption{Execution Monitoring and Replanning in some \HTN~planners}\label{fig:em}
\end{figure}

Figure~\ref{fig:em} shows the monitoring and replanning systems of \sipe, \oplan~and \siadex, the only planners capable of advanced replanning and execution monitoring.  The monitoring and replanning system of \sipe~is depicted in Figure~\ref{fig:em-sipe}, and  requires an input of arbitrary descriptions of the current plan and state in which the failure happened. This input is represented in form of predicates whose truth values should be checked. If the values of predicates differ from the expected ones, \sipe~uses its deductive theory to infer the changes. So far these steps are performed by the `Execution Monitor'. The next step is executed by the `Problem Recognizer' which checks the current plan against some predefined problems that can possibly occur, and returns a list of all detected problems. Consequently, \sipe~through its `General Replanner' tries to find so-called replanning actions that will modify the current plan. Some of these actions will insert new and unsolved goals into the plan. After having some actions found, \sipe~searches a solution for the modified plan which in fact represents a task network with unsolved goals. When  a new solution is found, it is proceeded for execution. 

The execution monitoring and replanning system of \oplan~is shown in Figure~\ref{fig:em-oplan}. The input to the system consists of the current plan, some tables containing the causal structure of the plan (`TOME' and `GOST'), and some monitoring strategies. A strategy may specify to monitor all actions and report the outcome of their execution; to monitor particular actions whether they will succeed or fail during execution; to monitor specified start and end time of an action relative to a some reference point; and to report only when the whole solution has completed execution. If the information derived from the monitoring contains a failure, then there are two ways to handle that failure. First, if \oplan~is ignorant about the failure, it initiates a planning process with all the necessary input about the unexpected situation, and waits for a plan. Second, if \oplan~knows how to handle the failure (the system posses a  `Capability'), it tries to revise the plan, and, in case of successful repair, the system continues with execution. In case of failed attempt, the information is passed to be planned for.

\siadex~implements an execution monitoring and replanning system as shown in Figure~\ref{fig:em-siadex}. It requires the state with exceptions and the current plan together with its casual structure as an input~\cite{fdez-olivares2011:tnmodeling}. The monitoring process first checks whether preconditions of an action are satisfied before and whether the effects of the action are correctly applied after the action execution in the real-world state. Second, the process ensures that actions are initialised and terminated according to their temporal constraints. Furthermore, the monitoring process checks the consistency of the casual structure of the plan. If there is some link failure, this check will detect exceptions as soon as they occur. Finally, this process supports interactive suggestions, such as decisions by experts made during run time or exceptions detected by experts. The next step performed by \siadex~is repair and replanning of the exception caught. For this step, a multilevel strategy according to criticality -- expresses the number of actions affected by the exception -- and the nature of exception is used. The first two levels are fully autonomous and repair the plan by applying some local changes or some repair rule a priori defined. If this is not effective, a replanning is initiated to create a new plan by using the domain knowledge. The last level is the last alternative. It provides lowest degree of autonomy and it depends on an intervention of a human expert. 

\subsubsection*{Interpretation}
\begin{figure} [!ht] 
\centering
\includegraphics[width=.6\textwidth]{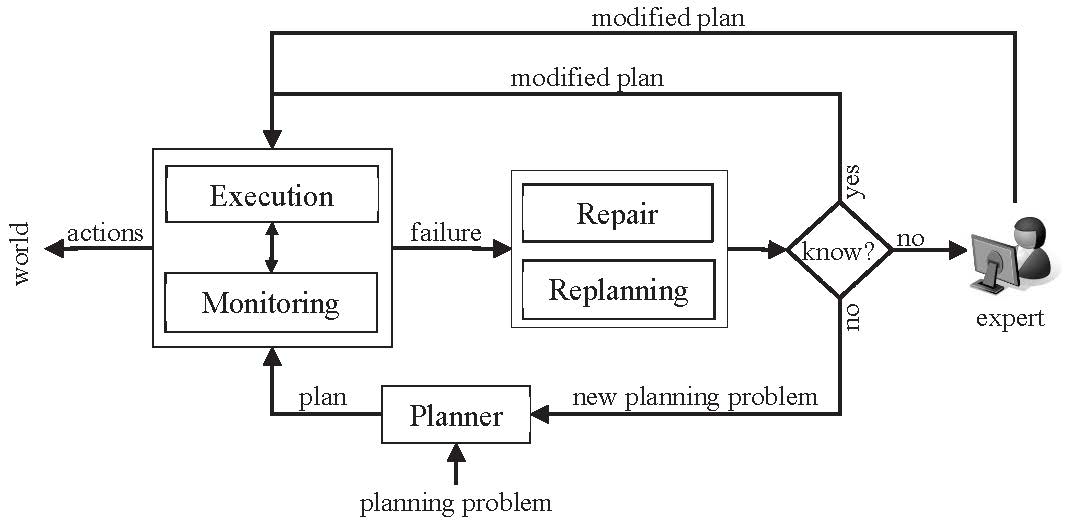}
\caption{General Architecture for Fault Tolerance}\label{fig:em-general}
\end{figure}

Table~\ref{tab:competence} suggests that \tHTN planners have much simpler search mechanisms than \pHTN~planners. This also might confirm the statement that the underlying mechanisms of \pHTN~are much more difficult to grasp. Furthermore, almost all planners adopt depth-first search. It is clear that the planners producing totally ordered plans are unable to support execution monitoring and replanning. With respect to execution monitoring and replanning, we provide a general architecture in Figure~\ref{fig:em-general}. This architecture aims to comprise functionality of \HTN~planners that support monitoring of plan execution and handling of failures.

\subsubsection{Performance}

\subsubsection*{Performance in Theory} 
The complexity results are summarised in Table~\ref{tab:complx}. When no restrictions on compound tasks are imposed and task networks are partially ordered, then giving the operators and methods in the input or fixing them in advance, or allowing variables or not, does not affect the outcome and the existence of a plan is undecidable. However, given the operators and methods in the input, and being every task acyclic and every task network partially ordered, the plan existence becomes decidable. 

\begin{table*} [!ht] \footnotesize
\caption{Time and Space Complexity of \HTN~Planning}\label{tab:complx}
\centering
\begin{tabular}{llllr}
\toprule 
{\bf Operators and methods} & {\bf Compound task} & {\bf Task network} & {\bf Variables} & {\bf Plan existence}\\ \hline
fixed & \multirow{7}{*}{yes} & \multirow{3}{*}{partially ordered} & \multirow{2}{*}{yes} & undecidable\\ \cline{1-1} \cline{5-5}
\multirow{14}{*}{input} & & & & undecidable\\ \cline{4-5}
& & & no & undecidable \\ \cline{3-5}
& & \multirow{6}{*}{totally ordered} & \multirow{2}{*}{yes} & 2-EXPTIME \\
& & & & EXPSPACE-hard \\ \cline{4-5}
& & & \multirow{2}{*}{no} & EXPTIME \\ 
& & & & PSPACE-hard \\ \cline{2-2} \cline{4-5}
& \multirow{5}{*}{no} & & yes & NP-complete \\ \cline{4-5}
& & & no & P \\ \cline{3-5}
& & \multirow{4}{*}{partially ordered} & yes & NP-complete\\ \cline{4-5}
& & & no & NP-complete \\ \cline{2-2} \cline{4-5}
& \multirow{2}{*}{acyclic} & & yes & decidable \\ \cline{4-5}
& & & no & decidable \\ \cline{2-5}
& \multirow{4}{*}{regular} & \multirow{4}{*}{unimportant} & yes & EXPSPACE-complete\\ \cline{4-5}
& & & no & PSPACE-complete \\ \cline{1-1} \cline{4-5}
\multirow{2}{*}{fixed} & & & \multirow{2}{*}{yes} & PSPACE\\
& & & & PSPACE-complete\\ 
\bottomrule
\end{tabular}
\end{table*}

The plan existence is decidable when task networks are totally ordered. In particular, when unrestricted compound tasks and variables are allowed, the existence is EXPSPACE-hard in double exponential time (2-EXPTIME), or, if no variable is allowed, the existence of a plan is PSPACE-hard in exponential time. When only primitive tasks and variables are allowed, the existence is NP-complete, irrespective of the ordering of task networks. Furthermore, forbidding the use of variables makes the existence to be in P. However, disallowing variables when task networks are partially ordered tasks does not change the outcome and the existence of a plan remains NP-complete.

Regardless of the ordering of task networks, when compound tasks are regular, there are two outcomes. When operators and methods are given in the input, and if variables are allowed, then the plan existence is EXPSPACE-complete, otherwise the plan existence is PSPACE-complete. When operators and methods are fixed in advance, and variables are allowed, the plan existence is PSPACE-complete in PSPACE.

\subsubsection*{Performance in Practice}
Unfortunately, for most of \HTN~planners the performance is unknown. To the best of our knowledge, two pieces of evidence report on performance and pairwise comparison results. The first evidence compares \umcp~and SHOP under loads of different problems~\cite{nau2000:mshop}. The experiments are based on the UM Translog~\cite{andrews1995:umtranslog}, a domain similar, but quite larger than the standard logistics domain. For this domain, a set of problems with increasing number of boxes to be delivered is randomly created. The results show that the run time for \umcp~is several orders of magnitude larger than the run time for SHOP. Only in first ten problems \umcp~appears to perform better than SHOP, as depicted in Table~\ref{tab:translog-perf}. Additionally, \umcp~faced some difficulties when trying to find solutions to the problems. The planner tries to solve only 37\% of total number of problems, and failed 45\% of those 37\%. The reasons for such behaviour are due to running out of memory, inability to find an answer within a specific time frame, or returning a failure. The second evidence compares the performances of \shop~and \siadex~\cite{castillo2005:temporal_enhancements,castillo2006:handling}. The planners are tested on the Zeno travel domain~\cite{long2003:ipc3} under a set of different temporal problems. In all cases, \siadex~outperforms the temporal version of \shop~\cite{nau2003:shop2}. Table~\ref{tab:zenotravel-perf} summarises the results of this comparison.

\begin{table*}\footnotesize [!ht] 
\caption{Performance of some of \HTN~Planners}\label{tab:performance}
\centering
	\begin{subtable}{0.45\textwidth}
			\centering
			\caption{Run Time and Scalability of \umcp~and SHOP }\label{tab:translog-perf}
			\begin{tabular}{ll|ll}
			\toprule
			{\bf Domain} & {\bf Property} & {\bf \umcp} & {\bf SHOP} \\ \hline \hline
			\multirow{3}{*}{UM Translog} & \multirow{2}{*}{Run Time} & 10 problems & 100 problems \\ 
			& & $>$ SHOP & $>$ \umcp \\ \cline{2-4}
			& Scalability &  & $>$\umcp \\
			\bottomrule
			\end{tabular}
	\end{subtable}
	\quad \qquad
	\begin{subtable}{0.45\textwidth}
			\centering
			\caption{Run Time and Scalability of \shop~and \siadex}\label{tab:zenotravel-perf}
			\begin{tabular}{ll|ll}
			\toprule
			{\bf Domain} & {\bf Property} & {\bf \shop} & {\bf \siadex} \\ \hline \hline
			\multirow{3}{*}{Zeno Travel} & \multirow{2}{*} {Run Time} & 2 problems & 18 problems \\
			& & $=$ \siadex & $>$ \shop \\ \cline{2-4}
			& Scalability & $=$ \siadex & \\
			\bottomrule
			\end{tabular}
		\end{subtable}
\end{table*}

\subsubsection*{Interpretation}
\HTN~planners, especially \pHTN ones, report obscure results about their performance. For example, for \sipe~we know that it is able to handle a domain that includes up to 200 tasks, 500 objects with 10 to 15 properties per object, and a problem that includes a few thousand predicates~\cite{wilkins2001:call}. However, this information is insufficient to make statements about the how well \sipe~performs. 

\subsubsection{Applicability}
While examining the literature on \HTN~planners, we did our best to satisfy our curiosity about the application tendency of other than \shop~\HTN~planners which led us to Table~\ref{tab:apps} that contains a list of applications where each state-of-the-art \HTN~planner is applied. Particularly, \oplan~and \shop~are the most practically used planners, while \sipe~and \siadex~share the same and relatively high number of applications. \sipe~and \siadex~are applied to at least 7 domains, ranging from aircraft carrier mission~\cite{wilkins1988:extending}, oncology treatment~\cite{fdez-olivares2008:oncology,fdez-olivares2011:clinical}, e-learning~\cite{castillo2010:elearning,garrido2013:elearning} and planning tourist visits~\cite{castillo2008:touristvisit,palao2011:touristvisit}, to a construction domain~\cite{kartam1989:foundation,kartam1990:approach}. \shop~is used to at least 11 applications, ranging from evaluation of enemy threats~\cite{munoz1999:hicap,munoz2001:sin}, Web service composition~\cite{wu2003:daml,sirin2004:shop2_composition}, to a project application~\cite{nau2005:applications}. Finally, \oplan~is applied to at least 15 domains, ranging from spacecraft mission control~\cite{drummond1988:spacecraft,drabble1997:repairingplans}, crisis management~\cite{tate1996:oplanlogistics,tate1994:oplan2}, to a domain of biological pathway discovery~\cite{khan2003:biological}. Evacuation planning is tackled by \oplan, \umcp~and \shop, while Web service composition is addressed also by three planners, namely \oplan, \shop~and \siadex. From a time perspective, the majority of studies (23 in total) are performed in the decade 1990-2000, then ten studies before 2000, while in the most recent times, three applications are implemented by \siadex.

\newcommand{\dec}[2] {\begin{tikzpicture}\draw[pattern color=black,pattern=#1] (0,0) rectangle (0.5,0.2);\end{tikzpicture}#2}

\newcommand{\decC}[2] {\begin{tikzpicture}\draw[fill=#1] (0,0) rectangle (0.5,0.2);\end{tikzpicture}#2}

\begin{table*} [!ht] \footnotesize
\caption{\HTN~Planners and their Applications}\label{tab:apps}
\centering
\begin{tabular}{l|lllllll}
\toprule
{\bf Application} & {\bf \noah} & {\bf \nonlin} & {\bf \sipe} & {\bf \oplan} & {\bf \umcp} & {\bf \shop} & {\bf \siadex} \\ \hline
Aircraft Carrier Mission Planning & & & \dec{crosshatch dots}{\cite{wilkins1988:extending}} & & & & \\
Air Campaign Planning & & & \decC{black}{\cite{lee1996:military}} & \decC{black}{\cite{tate1998:qualitative}} & & & \\
Biological Pathway Discovery & & & & \dec{crosshatch}{\cite{khan2003:biological}} & & & \\
Business Process Modelling & & & & & & & \decC{black!40}{\cite{gonzalez-ferrer2013:bpm}} \\
Care Pathways & & & & & & & \decC{black!40}{\cite{sanchez-garzon2013:carepathways}}\\
Construction Planning & & & \dec{crosshatch dots}{\cite{kartam1989:foundation}} 
 & \decC{black}{\cite{currie1991:oplan}} & & & \\
Crisis Management and Logistics & & & & \decC{black}{\cite{tate1996:oplanlogistics}} 
 & & & \\
E-learning & & & & & & & \decC{black!40}{\cite{castillo2010:elearning}} 
 \\
Electricity Turbine Overhand & & \decC{white}{\cite{tate1976:project}} & & & & & \\
Engineering Tasks & & & & \decC{black}{\cite{tate1996:responsive}} & & & \\
Equipment Configuration & & & \decC{black}{\cite{agosta1995:formulation}} & & & & \\
Evacuation Planning & & & & \decC{black}{\cite{tate2000:webbased}} & \decC{black}{\cite{erol1996:formalization}} & \decC{black}{\cite{munoz1999:hicap}} 
 & \\
Evaluating Terrorist Threats & & & & & & \decC{black}{\cite{nau2005:applications}} & \\
Forest Fire Fighting & & & & & & \dec{crosshatch}{\cite{nau2005:applications}} 
 & \dec{crosshatch}{\cite{delaasuncion2005:siadex}} \\
Location-Based Services & & & & & & \decC{black}{\cite{nau2005:applications}} & \\
Material Selection in Manufacturing & & & & & & \decC{black}{\cite{nau2005:applications}} & \\
Mechanical Engineering Supervision & \decC{white}{\cite{sacerdoti1975:structure}} & & & & & & \\
Military Operations Planning & & & \decC{black}{\cite{wilkins1992:military}} & & & & \\
Naval Logistics & & \decC{white}{\cite{tate1976:project}} & & & & & \\
Oil Spill Response & & & \decC{black}{\cite{sipe2:oilspil}} & & & & \\
Oil Tanker Truck Production & & & & \dec{crosshatch dots}{\cite{alvey1987:clubs}} & & & \\
Paediatric Ontology Treatment & & & & & & & \dec{crosshatch}{\cite{fdez-olivares2008:oncology}} 
 \\
Planning Tourist Visits & & & & & & & \dec{crosshatch}{\cite{castillo2008:touristvisit}} 
\\
Production-Line Scheduling & & & \decC{black}{\cite{wilkins1991:practical}} & & & & \\
Project Planning & & & & \dec{crosshatch dots}{\cite{alvey1987:clubs}} & & \decC{black}{\cite{nau2005:applications}} & \\
Search and Rescue Coordination & & & & \decC{black}{\cite{kingston1996:commonkadsmodels}} & & & \\
Software System Integration & & & & & & \dec{crosshatch}{\cite{nau2005:applications}} & \\
Spacecraft Assembly and Integration & & & & \decC{black}{\cite{aarup1995:optimum}} & & & \\
Spacecraft Mission Planning & & & & \dec{crosshatch dots}{\cite{drummond1988:spacecraft}} 
 & & & \\
Spacecraft Platform Construction & & \decC{black}{\cite{tate1990:interfacing}} & & \decC{black}{\cite{drabble1997:repairingplans}} & & & \\
Statistical Goal Recognition & & & & & & \dec{crosshatch}{\cite{nau2005:applications}} & \\
Unmanned Autonomous Vehicle & & & &  \decC{black}{\cite{tate2003:web}} & & \decC{black}{\cite{shop2:comets,shop2:sift}} & \\
US Army Small Unit Operations & & & & \decC{black}{\cite{tate2000:usingai}} & & & \\
Web Service Composition & & & & \dec{crosshatch}{\cite{uszok2004:applying}} & & \dec{crosshatch}{\cite{wu2003:daml}} 
 & \dec{crosshatch}{\cite{fdez-olivares2007:wsc}}\\ \hline
\multicolumn{1}{r}{Total} & 1 & 3 & 7 & {\bf 15} & 1 & 10 & 7 \\
\bottomrule
\multicolumn{8}{r}{} \\
\multicolumn{1}{r}{\multirow{2}{*}{Time Line}} & \multicolumn{7}{l}{\hspace{0.15cm}\decC{white}{}\hspace{0.55cm}\dec{crosshatch dots}{}\hspace{0.52cm}\decC{black}{}\hspace{0.52cm}\dec{crosshatch}{}\hspace{0.52cm}\decC{black!40}{}} \\
\multicolumn{1}{r}{} & \multicolumn{7}{l}{<1980\hspace{0.3cm}<1990\hspace{0.3cm}<2000\hspace{0.3cm}<2010\hspace{0.3cm}<2020} \\
\end{tabular}
\end{table*}

\subsubsection*{Interpretation}
Recalling our ultimate objective for AI planning, \HTN~planning has contributed by being employed to more than 30 applications. More than half of them are tackled with \pHTN planning, and \oplan~appears to be the most applied \HTN~planner, while \shop~is the most applied \tHTN~planner.

\section{\HTN~Planning for Web Service Composition}\label{sec:wsc}
{\em Web services} are software components that provide functionalities to machines that communicate over the Web. Various companies and organisations implement and offer their business logic as Web services. Adequate selection and integration of inter-organisational Web services at runtime is real challenge for modern applications on the Web. If the request of an end-user cannot be met by a single Web service, then maybe combination of already existing services will provide value-added functionality and satisfy the user request. Thus, this idea is presented exactly with the Web service composition.

Manual composition of Web services becomes impossible due to several reasons. First, the number of available Web services increases constantly which requires considerable effort to search Web service repositories for a suitable service. Second, there is no unifying description language for Web services which implies many and different description styles that affect the identification and understanding of the meaning of Web services. Third, Web services may appear or disappear dynamically and irregularly, while other may change their description over time. This dynamicity implies uncertainty about some observed state of a Web service at a particular point in time, service unavailability, and other unexpected behaviour of Web services, especially at their execution time. Finally, the user request may involve complex conditions or some policies on the behaviour of the composition as whole or only parts of it. The request may be in form of a workflow template as well. A workflow specifies activities, their ordering, the parameters for each activity, and other details used to accomplish a particular task. A workflow template is a generalisation of a workflow, in which some of the activities are declared as abstract. An abstract activity defines preferences or constraints used to match, rank and select a concrete service at run time. For example, a preference could indicate a need for services with certified authority or a need for non-fee services. 

These observations suggest that automated Web service composition is a challenging task. The research community of Web service composition focuses on developing tools, techniques and intelligent systems able to return a composition with the best possible quality-of-service values~\cite{brian2009:wscreview}, as is the case with the Web Service Challenge~\cite{wschallenge2010,blake2006:wsc}. In the Web Service Challenge, systems, such as~\cite{aiello:2008:visualizing,aiello2009:optimal,degeler2010:concept}, generate two compositions, namely a service composition with the lowest response time, and a service composition with the highest throughput. The response time expresses the delay between the time a request is received by a Web service and the time a reply to the request is sent, and the throughput expresses the amount of requests that a Web service can handle in a given time unit. On the other hand, the AI community tries to automate the process of Web service composition by viewing the composition problem as a planning problem~\cite{aiello2002:request,sirin2004:shop2_composition,kuter2005:gathering,medjahed2005:multilevel,klusch2005:semanticwsc,sohrabi2006:wsc,paik2007:wscomposition,kaldeli2009:extended,kaldeli2011:continual}. The general assumption is that Web service descriptions can be mapped to planning operators, that is, each service can be represented by its preconditions and effects in the planning context. Among planning techniques, \HTN~planning seems particularly suitable for the purpose of Web service composition because it encourages  service modularity, can minimise a variety of failures or costs, scales well to large numbers of services, supports acquiring additional information by invoking Web services during planning, and enables user/software intervention~\cite{sirin2004:shop2_composition}.

Our objective is to assess how \HTN~planning and \HTN~planners address the challenges of Web service composition, and, at the same time, to assess whether these suitability statements can be confirmed. 

\subsection{WSC Framework}\label{sec:wscf}

The last slice of out framework pie represents the application framework. In fact, we propose a general framework for Web service composition based on AI planning. We are interested in aspects that concern the automatic composition of Web services accomplished by planning techniques. The framework provides an abstract view without requiring a particular description language, planning algorithm, or monitoring and execution approach.

\begin{figure} [!ht] 
\centering
\includegraphics[width=.6\textwidth]{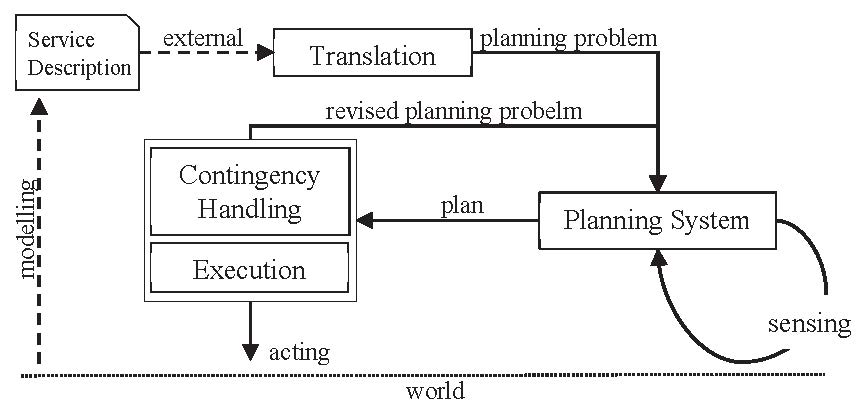}
\caption{WSC Framework}\label{fig:wsc-framework}
\end{figure}

The Web service composition framework is shown in Figure~\ref{fig:wsc-framework}. A Web service that models some business logic is described in some (external) language and provided to a translator that creates appropriate (internal) representation, that is, a planning problem. Consequently, the planning problem is given to a planner to search for a solution. If there is a solution found, that is, a plan, it is passed for execution and monitoring for potential failures. In case of a failure, appropriate actions are taken.

\begin{itemize}
\item {\bf Service description}: The description of Web services offered to the global market usually consists of three parts. The first part refers to the information about the data transformation during the execution of a service. The information is presented in form of input, output and possibly exceptions. The input contains the information required for service execution, while the output presents the information the service provides after its execution. The second part refers to when and how a service transforms the world. This part consists of preconditions, that is, requirements that must be satisfied for the service to be invoked, and postconditions, that is, physical changes to be made to the world. The last part contains the non-functional properties of a service, such as cost, reliability, and service quality.

\item {\bf Translation}: Services described in a standard Web service language appear to be hard to handle by planning systems unless they are translated into an understandable form. The translation component accepts service descriptions and converts them into formal and unambiguous encoding. The result of the translation is a planning problem. In fact, this component enables the relationship between Web service composition and AI planning.

\item {\bf Planning System}: It takes the planning problem and tries to find a solution. Many planning systems distinguish between world-altering and sensing actions. The former can change the world when executed, while the latter cannot modify the state, but only acquire additional information needed to support the planning process. The most common approach is to perform off-line planning, that is, to simulate the execution of world-altering actions, and to do sensing. The solution, if it exists, consists of world-altering actions only. Many planning systems make several assumptions while planning. These assumptions simplify the planning process, but impose restrictions about what might happen in the world and distance further from the reality. The assumptions are:
	\begin{description}
	\item[A1:] The world is static -- it can be modified only by the actions resulted from the planning process, and not by some external agent or event. All information about the world is expected to be valid till the end of the execution.
	\item[A2:] Sensing actions succeed -- the execution of a sensing action will always return the acquired information.
	\item[A3:] Sensing actions are repeatable -- the first sensed information is assumed to be valid for each to the same action (service) further in the planning process.
	\item[A4:] No changes are made to services -- service's functional properties are constant during the planning and execution process.
	\end{description}

\item {\bf Execution Monitoring and Contingency Handling}: Considering that the world is dynamic and uncertain, the execution of actions might not proceed as expected. A contingency may be inconsistent sensed information, failures of service invocations, timeouts, or unexpected change in the world. These observations suggest that the problem of Web service composition should not be tackled decoupled from the process of action execution. Monitoring of execution and contingency handling appears to be suitable to address the aforementioned issues. Execution monitoring checks the validity of off-line calculated actions when executed and, in case of contingency, reacts appropriately. For example, if the execution time of some service takes too long, then it might be possible to proceed with the execution of subsequent actions. Other types of contingency may require repair of the existing plan, or even planning from scratch. 
\end{itemize} 	

\subsection{WSC and \HTN~Planning}
We apply our general framework to \HTN~planning and approaches that employ \HTN~planning to compose Web services. All steps remain the same, except that some elements of the framework can be concreted now. That is, we choose a particular service description language upon which we define the problem of Web service composition and its corresponding \HTN~planning problem.

\subsubsection*{OWL-S}
The majority of studies that employ \HTN~planning and are examined here use OWL-S to describe Web services. OWL-S~\cite{martin2007:owl-s} is a Web ontology~\cite{horrocks2003:ontology} for Web services used to support automated discovery, enactment and composition of Web services. The OWL-S ontology has three components: service profile, process model and service grounding. The service profile indicates the purpose of a service, and comprises the elements of part one and part three described in `Service description' step in the framework. The process model indicates how to accomplish the service purpose, how to invoke the service, and what happens after the service execution. The service grounding specifies the way of interaction with the service, including a communication protocol.

OWL-S perceives services as processes. It defines three classes of processes: atomic, simple and composite. An atomic process has no sub-processes, has a grounding associated with it, and can be executed in a single step. A simple process provides an abstraction for an existing service, and has no associated grounding. A composite process consists of other processes via control constructs, such as Sequence, Split, Any-Order, etc.

\subsubsection*{WSC problem as an \HTN~Planning Problem}

The services described in OWL-S need to be encoded in corresponding \HTN~elements. Intuitively, each atomic process is translated to an operator, and each simple and composite process is translated to a method. If we consider that $\mathcal{P^W} = (s_{0},K,C)$ is a WSC problem described in OWL-S, where $s_{0}$ is an initial state of the world, $K$ is a collection of OWL-S process models, and $C$ is a composite OWL-S process defined in $K$, then the following relationship could be established (adopted from~\cite{sohrabi2013:customizing}).

\begin{mydef}[WSC Relationship to HTN Planning]
Let $\mathcal{P^W} = (s_{0},K,C)$ be an OWL-S WSC problem. Then, the sequence $p_{1},\dots,p_{n}$, where each $p_{i}$ is an atomic process defined in $K$ is a solution to $\mathcal{P^W}$ if and only if $t_{1},\dots,t_{n}$ is a solution to $\mathcal{P}= (Q,T_{p},T_{c},O,M,tn_{0},s_{0})$, where
\begin{itemize}
\item $Q,T_{p},T_{c},O,M$ are generated by an OWL-S to HTN translation for the OWL-S process models $K$,
\item $tn_{0}$ is generated by an OWL-S to HTN translation for the OWL-S process $C$, and
\item each $t_{i}$ is a primitive task that corresponds to atomic process $p_{i}$ defined by some OWL-S to HTN translation.
\end{itemize}
\end{mydef}

\subsection{Review}\label{sec:wsc-review}
Now that we have the Web service framework, and the relation between WSC OWL-S problem and \HTN~planning problem, we are able to review several studies. We group the studies into two categories. First, we analyse approaches that employ \HTN~planning exclusively in the attempt to solve the problem of Web service composition. Second, we discuss approaches that combine \HTN~planning with another technique, such as description logic and constraint satisfaction to compose services.

\subsubsection*{Approaches Based on \HTN~Planning Only}
The earliest approach to WSC uses the predecessor of OWL-S (DAML-S~\cite{burstein2002:damls}) to describe Web services~\cite{wu2003:daml}. An atomic process with effects only is translated into an operator that will simulate the effects of a world-altering Web service. An atomic process with output only is translated into an operator that has a call to a sensing Web service encoded in its precondition, and an empty delete list. A simple process is translated into a collection of methods, while a composite process is translated either into a method or a collection of methods, depending on the control construct used in the composite process. Given a composite process, a list of available processes and an incomplete initial state, the translation component creates an \HTN~planning problem which is in turn passed to the \shop~planner. Planning is performed just as described in the framework. While planning, sensing is performed with respect to the initial state meaning that the planning results would have been the same if all sensing Web services were executed first and then the planning was performed. In addition, the information provided by sensing Web services is cached in order to avoid invoking a concrete service multiple times during planning. The invocation of sensing Web services is handled by some monitor which calls an executor to execute the sensing Web service, and caches its responses. The creation of compositions is accomplished under assumptions A1-A4 which guarantee \shop. This approach is later adjusted to handle OWL-S~\cite{sirin2004:shop2_composition}. In both cases, sensing Web services may return some information or may not return at all, which in turn may block the planning process. This issue is solved in~\cite{kuter2005:gathering} by augmenting \shop~to plan continuously even when a sensing service has not provided any information yet. The algorithm starts with an incomplete initial state, senses needed information during planning, and explores alternative paths when some sensing Web services is executing. There is no need of explicit specification of sensing Web services in the initial description, but a query mechanism is used to search and select appropriate services when information is needed. 

Madhusudan and Uttamsingh~\cite{madhusudan2006:wsc} suggest a system that creates compositions of Web services by using SHOP and supports execution monitoring and replanning in case of contingency. The translation between the WSC problem and \HTN~planning problem is not clearly defined. A catalogue of declaratively described Web services is provided. The catalogue can contain a world-altering or sensing Web service, which corresponds to an \HTN~operator, and a composite Web service, which corresponds to an \HTN~method containing a sequence of \HTN~operators. The goal is declaratively described, and in fact represents a goal state. Given a catalogue of Web services, methods and operators, and a goal state, SHOP uses its methods to guide the search towards reaching the goal state. Sensing is performed during planning to acquire additional information. The final plan is a sequence of world-altering actions. The execution monitoring enables consistent execution of plan steps. It updates the catalogue with the most recent information about active Web services and functional changes of Web services. In case of contingency at execution time, it is handled by executing the remaining plan in the current state, or by replanning. The choice between these two resolving options is not clearly defined. This approach is applied at least under assumptions A1 and A2.

Fern\'{a}ndez-Olivares et al.~\cite{fdez-olivares2007:wsc} propose a middleware to interpret OWL-S descriptions by translating them into an \HTN~domain and problem descriptions, and to reason over those descriptions by using temporal \HTN~planning techniques in order to create an executable sequence of Web services. In particular, an OWL data model is translated into a PDDL data model, that is, OWL classes, properties and instances are translated into PDDL types, predicates and objects, respectively. Furthermore, an OWL-S process model is translated into an \HTN~domain model, that is, atomic processes are translated into PDDL durative actions, while the workflow pattern of each composite process is translated into a method associated to a compound task that corresponds to the composite process. A service request is translated into a compound task, while the OWL instance from the OWL data model represent the initial state. Such domain and problem descriptions are passed to an \HTN~planner. Here the \siadex~planner is employed to create a sequence of temporally annotated actions that corresponds to a composition of atomic processes. The middleware is able to monitor the execution of the composition according to the temporal information associated to the atomic processes. The execution of Web services can be performed during the planning or the plan-execution time. If a service fails or return a faulty sensed information at planning time, the planner backtracks and selects another service if possible, or tries different decomposition, if available. If some failure occurs at execution time, replanning is initiated.

The studies discussed so far use \HTN~planners to tackle functional properties of Web services being composed. Beside these functional properties, the service description may specify non-functional properties. These properties may include privacy, security, usability, reliability, cost, and performance, which can be used when deciding whether to include some service in a composition. Kuter and Golbeck~\cite{kuter2009:wsc} describe a taxonomy of such properties which is used to compute user's trust in a service composition process. The trust value of a service is approximated by using the ratings of users. This trust value is used to infer the trustworthiness of a composition, while the most trustworthy composition is chosen as a solution to a WSC problem. In addition to our definition of a WSC problem, here the problem has three more elements, namely the user that performs the composition, the set of all properties in the problem, and a set of user ratings. Such defined WSC problem is given as an input to an algorithm build on top of \shop. The assumptions made in this approach are that users' ratings are independent of each other, all services and associated ratings are given in advance, and assumption A4. 

Another perspective of desired compositions is represented by different regulations required by corporations or government that provide some Web services. End-users concerned with composing such services must enforce regulations expressed as corporate policies or government rules to different aspects of compositions. Sohrabi and McIlraith~\cite{sohrabi2009:wsc} suggest an approach to Web service composition based on \tHTN planning that considers such policies and regulations, while taking into account preferences over how a task can be decomposed, and preferences over services and data selection. The \HTN~planning problem is created from the following elements: 1) Web services are described in OWL-S and translated into \HTN~domain knowledge similarly as we described for the first study~\cite{sirin2004:shop2_composition}; 2) the user's task is translated into an initial task network; 3) user preferences are represented as soft constraints in version~3 of PDDL~\cite{gerevini2006:pddl3}, just as we described in Section~\ref{sec:expressiveness}; and 4) policies and regulations are expressed as a subset of LTL. Such planning problem is given as an input to an extended version of \shop~which selects services based on their functional and non-functional properties, while enforcing provided regulations. The extended planner performs best-first, incremental search and takes an \HTN~planning problem, a metric function and a heuristic function. The elements in the search frontier are sorted according to the heuristic function, while the metric function provides the quality value of the current plan. The planner creates a solution for the problem which represents an optimal composition for the WSC problem. The follow-up of this study enables creation of high-quality compositions through optimisation of service and data selection~\cite{sohrabi2010:wsc}. The idea is to minimise the access to unknown data and to optimise efficiently data by using the independence of actions during planning. Interesting are the cases when data optimisation can be accomplished in isolation of the generation of the composition. This can happen when data is irrelevant to the optimisation of the composition (it is not mentioned in the preferences), or when the data choice does not interfere with the flow of the composition. Here atomic processes can be either world-altering or sensing. The translation of sensing atomic processes into \HTN~operators is modified to explicitly encode caching of sensed information. Another modification to translation is done with respect to the cases of data optimisation and associated services in such a way that the execution of a sensing service is removed and the occurrences of data are replaced with placeholders. Such sensing service is executed after the computation of the composition which suggests replacing the placeholder with an appropriate choice. While planning, the assumptions A1, A2 and A3 are made. 

\subsubsection*{Approaches that Combine \HTN~Planning with Other Technique}

The I-X planning system~\cite{ix} supports services described in OWL-S and employs the \oplan's successor to generate Web service compositions~\cite{uszok2004:applying}. The study does not give details on how the translation from the Web service composition problem to the planning problem is accomplished. The input to the planning system consists of a goal to be achieved, knowledge about available services and other agents it may use, and a set of some policies. Policies constrain the behaviour of the system and include procedures for, for example, digital rights management, trust management, coalition formation etc. The planning process starts by selecting an initial plan and building OWL-S profiles for all necessary services. After a number of planning steps, the system creates a plan that will achieve the goal, be consistent with these policies, and enacted by chosen services. This idea is explained on a synthetic example, called coalition search and rescue, where the request is to save or help someone in trouble by searching medical facilities and performing rescue operations according to standard policies defined in the coalition. Planning is performed under assumption A1, that is, no local threats can happen that will prevent the rescue.

Early versions of OWL-S do not provide clear specification of preconditions and effects of Web services, which consequently affects the translation from the WSC OWL-S problem into an \HTN~planning problem. In many cases, the translation is fixed by inserting own encodings of preconditions and effects into translated tasks. However, the expressiveness of preconditions and effects in a planner has a rather different expressiveness in OWL. Later versions of OWL-S try to solve this issue by providing a default language for specifying service preconditions and effects. In order to evaluate these preconditions and effects, planners must understand the semantics of OWL. In addition, OWL-S has open world semantics, while \HTN~planners assume to have complete information about the world, thus, a close-world assumption. These issues are addressed in~\cite{sirin2004:semanticws} where a combination of \tHTN planning and Description Logic (DL) for the purpose of composing Web services is investigated. Web services are described in a syntax similar to PDDL (where preconditions and effects are written in the Semantic Web Rule Language (SWRL)), while a planning problem is encoded in OWL. Such domain and problem descriptions are given as an input to an integrated system. The integration consists of an OWL reasoner and a \tHTN planner, that is, the Pellet reasoner~\cite{pellet2003} and the Java version of SHOP planner. The planner simulates the effects of the Web services, while the reasoner takes care of the state and answers queries about the truthfulness of preconditions issued by the planner, and updates the state with simulated effects. The idea of this approach is to optimise the query answering time minimising the number of consistency checks. This approach served as a basis for the follow-up study in which compositions are created based on workflow templates~\cite{sirin2005:template}. As first, OWL-S is extended to support workflow templates in terms of ability to describe abstract functionalities and encode qualitative preferences. Then, since \tHTN planning cannot directly support such extension of OWL-S, \tHTN formalism is extended with DL knowledge about task and preference descriptions. According to the authors, the main advantage of representing services in this way is the ability to refer to profiles of the services in addition to the process descriptions. The preference descriptions about these profiles are stored in DL which means that OWL definitions can be directly used without the need for translation. Thus, the HTN-DL domain consists of a set of operators, a set of methods, and a task ontology defining the profiles and preferences related to tasks. OWL-S processes are translated into \HTN~elements which are handled as in usual \tHTN algorithm. The main difference in the proposed algorithm is the matching mechanism implemented by the DL reasoner. This mechanism returns the matching actions for a given task in the order specified by the preferences. Here too, JSHOP is used as an \tHTN planner combined with the Pellet reasoner.

Paik et al.~\cite{paik2006:framework, paik2007:wscomposition} combine an \HTN~planner with a Constraint Satisfaction Problem (CSP) solver to handle user requests that contain some additional information, such as scheduling information. A user request is translated into an \HTN~planning problem for which a solution containing a sequence of actions and some CSP set is computed. The CSP set corresponds to the temporal information specified in the user request. Such solution is passed to a CSP solver that calculates the final solution by binding variables with values that satisfy the temporal constraints. From an implementation perspective, JSHOP2 is used as an \HTN~planner, while the implementation of the CSP solver is left unspecified. From the description provided, one cannot understand how are the user requests and Web services described, how exactly the translation between the user request and \HTN~planning is performed, and what assumptions are made while producing the solutions to the \HTN~planning and CSP problem.

Lin et al.~\cite{lin2008:wsc_preferences} propose a composition of Web services described as OWL-S processes and preferences described as soft constraints in the version 3 of PDDL~\cite{gerevini2006:pddl3}. These processes and preferences are translated into an \HTN~planning formalism based on the definitions provided in~\cite{sirin2005:template} and~\cite{erol1996:formalization}. Given an initial state, an initial task network, a set of method and operators, a task ontology, user preferences and some violation-cost function, the first step of the proposed algorithm is to preprocess the initial task network. The initial task network might be modified based on its constraints and the user preferences by adding some tasks. If there are constraints that cannot be satisfied, they are removed from the task network. After the preprocessing step, the algorithm calculates a violation-cost value for the task network which is used for selecting the possible decomposition. The value indicates the violation of some constraints (preferences) into the decomposition in the current state. Based on these costs, task networks are sorted in ascending order and the first decomposition is chosen. The algorithm is built on top of the JSHOP planner extended with the Pallet reasoner, that is, on top of the approach proposed in~\cite{sirin2005:template}.

\subsection{Summary}\label{sec:wsc-summary}
We propose a general and abstract framework for classifying and comparing the automatic composition of Web services based on AI planning. Furthermore, we specify some aspects of the framework when seen from the perspective of \HTN~planning. In particular, we choose OWL-S as a description language for Web services which enables us to provide a precise definition of the problem of Web service composition and its relationship to an \HTN~planning problem. Having a framework and an \HTN~planning problem, we review several studies that employ \HTN~based techniques to solve the problem. Table~\ref{tab:wsc} summarises the studies with respect to indicators extracted from the framework. Some indicators are associated with rates. The rates range from  '\ding{72}', indicating limited focus or limited support for a respective indicator, to '\ding{72}\ding{72}\ding{72}', specifying comprehensive focus or extended support for the corresponding indicator. If a cell is empty, it denotes that we were not able to extract the information for the respective indicator from the literature. {\em Service Description} provides the language for describing Web services assumed by the study being analysed, while {\em translation} gives the dual information. First, it indicates how well and exactly the translation process is described, and second, which format is the Web service description translated to. {\em \HTN~Model} tells whether \tHTN or \pHTN planning is employed, and which \HTN~planner is used for the implementation of the taken approach. Beside the extent to which it is supported, {\em sensing} may indicate whether the execution of a sensing action blocks the planning process, and whether sensing actions are performed during planning or they may be interleaved with world-altering ones during execution. {\em Assumptions} concern the degree of assumptions made to guarantee a successful composition with respect to composing, sensing and executing actions. {\em Contingencies} refers to unexpected behaviour of a composition at execution time, including Web service failures or time outs, and events or information changes made by some external agents. Each approach is evaluated with respect to the extent to which the support is implemented, and the type of contingency the approach can handle.

\begin{table*}\footnotesize [!ht] 
\centering
\caption{Summary of \HTN-based approaches to Web service composition}\label{tab:wsc}
\begin{tabular}{L{1.2cm}C{1.7cm}C{2.4cm}C{2.5cm}C{2.4cm}C{1.5cm}C{1.8cm}}
\toprule
{\bf Study} & {\bf Web Service Description} & {\bf Translation (Representation)} & {\bf \HTN~Model (Planner)} & {\bf Sensing (Properties)} & {\bf Assumptions} & {\bf Contingencies (Types)} \\ \hline
\cite{wu2003:daml, sirin2004:shop2_composition, kuter2005:gathering}& DAML-S~\cite{burstein2002:damls}
\newline OWL-S & \ding{72}\ding{72}\ding{72} \newline (\shop) & state-based \newline (extended \shop) & \ding{72}\ding{72} \newline (blocking/non-blocking, during planning) & A1-A4 & \ding{53} \newline (not discussed) \\ \hline
\cite{madhusudan2006:wsc} &  & \ding{72} \newline (SHOP) & state-based \newline (extended SHOP) & \ding{72} \newline (interleaving) & A1,A2 & \ding{72}\ding{72} \newline (replanning for failures, time outs) \\ \hline
\cite{fdez-olivares2007:wsc} & OWL-S & \ding{72}\ding{72} \newline (HTN-PDDL) & state-based \newline (\siadex) & \ding{72}\ding{72} \newline (during planning) &  & \ding{72}\ding{72} \newline (replanning for failures, time outs)\\ \hline
\cite{kuter2009:wsc} & OWL-S & \ding{53} & state-based \newline (extended \shop) & \ding{53} \newline (not discussed) & A4, other & \ding{53} \newline (not discussed)\\ \hline
\cite{sohrabi2009:wsc,sohrabi2010:wsc} & OWL-S & \ding{72}\ding{72}\ding{72} \newline (\shop,PDDL,LTL) & state-based \newline (extended \shop) & \ding{72}\ding{72} \newline (during planning) & A1-A3 & \ding{53} \newline (not discussed) \\ \hline \hline
\cite{uszok2004:applying} & OWL-S &  & plan-based \newline (I-X) & \ding{72} \newline (during planning) & A1 & \ding{53} \newline (not discussed) \\ \hline
\cite{sirin2004:semanticws,sirin2005:template} & OWL-S & \ding{72}\ding{72}\ding{72} \newline (SHOP,DL) & state-based \newline (JSHOP + Pellet~\cite{pellet2003}) & \ding{53} \newline (not discussed) &  & \ding{53} \newline (not discussed) \\ \hline
\cite{paik2007:wscomposition} &  & \ding{72} \newline (\shop,CSP) & state-based \newline (JSHOP2 + CSP Solver) & \ding{53} \newline (not discussed) &  & \ding{53} \newline (not discussed) \\ \hline
\cite{lin2008:wsc_preferences} & OWL-S & \ding{72}\ding{72}\ding{72} \newline (SHOP,DL,PDDL) & state-based \newline (extended JSHOP + Pellet) & \ding{53} \newline (not discussed) &  & \ding{53} \newline (not discussed) \\ 
\bottomrule
\end{tabular}
\end{table*}

Most of the approaches assume OWL-S description of Web services, and provide sound translation algorithms to appropriate internal representation. With respect to the \HTN~model, all approaches but one employ state-based \HTN~planning. From the state-based \HTN~approaches, only one uses the \siadex~planner, while the rest exploit SHOP or its successor \shop. Most of the approaches give actual contribution to \HTN~planning by extending the existing algorithms or providing new algorithms on top of the existing planners. With respect to sensing, only few approaches devote appropriate attention to it and provide clear description. We can observe and conclude that sensing is done during planning and, in some cases, in a non-blocking manner. While planning, sensing and possibly executing Web services, several approaches make some of the restricting assumptions, at least those that we were able to identify from the description provided. Finally, little attention is devoted to execution monitoring and handling of contingencies at execution time.

\begin{figure} [!ht] 
\centering
\includegraphics[width=.7\textwidth]{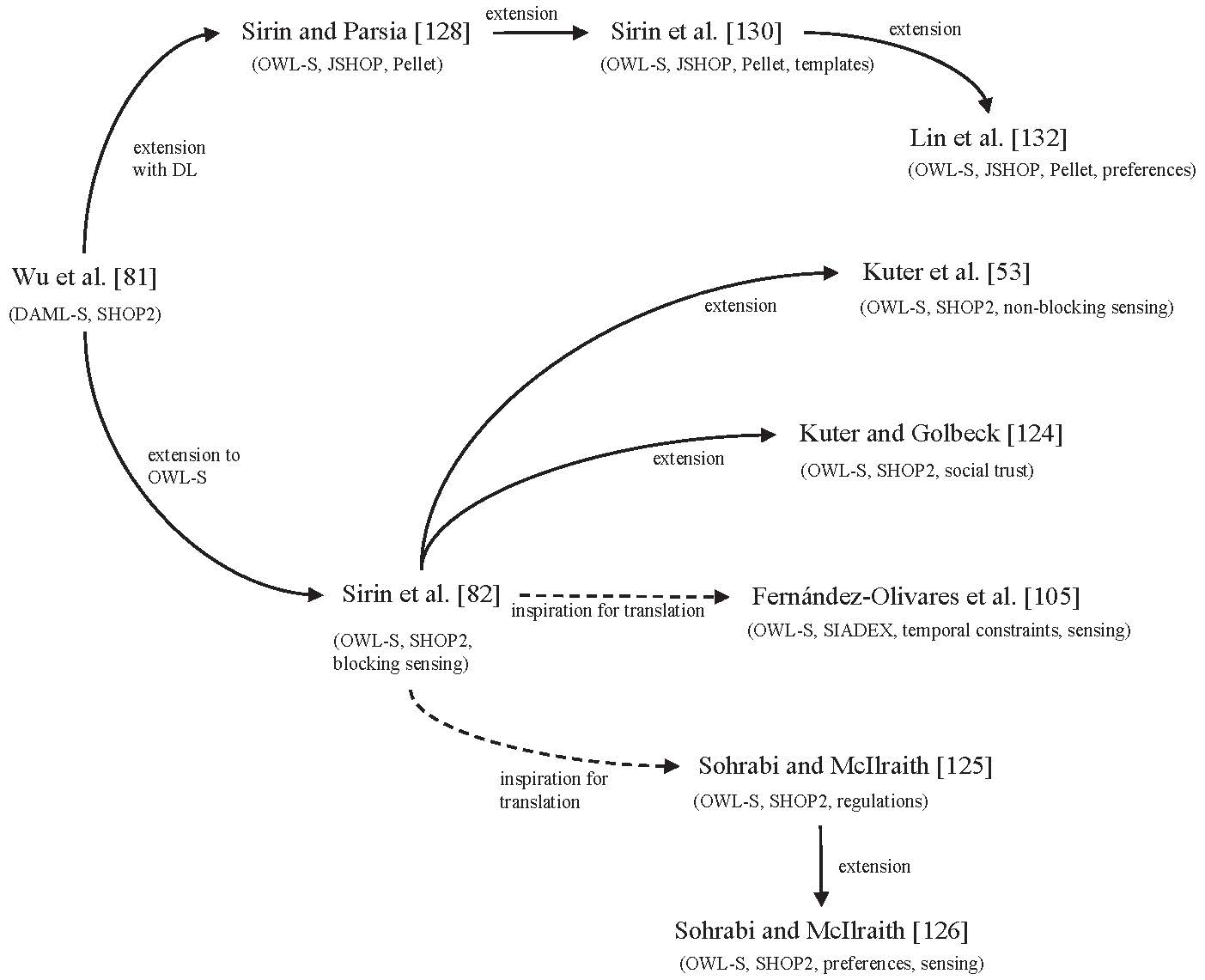}
\caption{Relations between studies that employ \HTN~planning for Web service composition}\label{fig:wsc-dependency}
\end{figure}

Figure~\ref{fig:wsc-dependency} gives another perspective of approaches that assume OWL-S description of Web services and provide clear translation to the planning-level representation. The lower part specifies the studies that employ \HTN~planning only. Sirin et al.~\cite{sirin2004:shop2_composition} appears to be the most influential and inspiring study. Two of them are a direct extension of the study, while the other two draw inspiration from the study with respect to the translation process. The upper part depicts the studies that combine \HTN~planning with DL reasoning. All studies are a continuation of the work presented in the first paper on \HTN~planning for Web service composition~\cite{wu2003:daml}.

\section{Conclusions}\label{sec:conclusions}
An observation of studies related to \HTN~planning from a period of almost 40 years is likely to result in little understanding on the current theoretical, technical and practical state of the art of \HTN~planning. First, the field encompasses studies scattered over a long period of time. Second, several \HTN~planners exist with interesting technical and practical implications, while only some of them received proper attention. Third, the modern and real-world applications pose challenges to planning techniques that might have not be principally envisioned and covered by existing \HTN~planners.

To assist in improving the outcome of such important observation, we proposed, described and evaluated a pie of frameworks whose main flavour is \HTN~planning. The pie is sliced into four frameworks. First, we define a theoretical framework upon which we derive two major formal models of \HTN~planning. Then, we design a conceptual framework which enables us to interpret differences and similarities of \HTN~planners from a perspective where the search space is the central concept. Third, we describe an analytical framework to examine in details several \HTN~planners, and to draw some conclusions more objectively. Finally, we choose Web service composition as an application domain for which we create a framework upon which we base the review and assessment of all studies that employ \HTN~planning for the problem of service composition.

We categorise \HTN~planning in \pHTN planning and \tHTN planning, while we define a formal model for each category. Based on this categorisation, we group our chosen \HTN~planners. We find that \pHTN planners need to search more complex spaces than \tHTN planners. This affects also the concepts used during the search in terms of the number of needed techniques, their technical complexity, and their interconnection. Furthermore, on an exemplifying domain, we find that \pHTN planners require smaller domain knowledge than \tHTN planners. As for the expressiveness, we know that \HTN~planning is more expressive that \strips-like planning, but that none of the \HTN~planners can achieve that expressiveness in practice. Moreover, it appears that both categories of planners have similar levels of practical expressiveness. Almost all planners use depth-first search, but only few show flexibility for incorporating other alternative mechanisms. Based on the systems of three \HTN~planners, we design a general execution monitoring and replanning architecture suitable for modern planners. We had difficulties when assessing the scalability and efficiency of \HTN~planners as little is reported on their performance. As for the applicability of \HTN~planners, we find that \oplan~is the most applied one, followed by \shop. The last part of our study shows that almost all approaches to Web service composition use a pre-processing step to convert an external service representation into a possibly complete and sound internal representation suitable for \HTN~planners. A large majority of studies employ \tHTN planning to solve the Web service composition problem, while the planners are able to address only partially the properties of Web service environments.

The topics explored herein suggest that there are still interesting research directions open for \HTN~planning. For example, theoretical contributions in terms of new models and algorithms for \HTN~planning, as recently proposed search spaces and algorithms in~\cite{alford2012:htnspaces}. Furthermore, a hot topic appears to be landmarks, that is, abstract tasks that occur in every solution found by an \HTN~planner as introduced in~\cite{elkawkagy2010:landmarks,elkawkagy2012:improving}. A common syntax and semantics for specifying \HTN~domains and problems is missing. A description language \`{a} la PDDL can stimulate improvements of research and performance evaluation of \HTN~planners, can enable direct comparison of planners on possibly standardised set of problems, and finally, can help in understanding the practical expressive power of \HTN~planners.  Another item to the list are complex goals. Rather than writing a new task or combining several tasks to accomplish a goal, or simply declaring a desired state, additional conditions may be stated in the goal, such as maintainability properties, or whether we are only interested in observing the environment or we wish to change the environment. In some situations, hybrid goals could be preferable, that is, a mixture of declarative and imperative representation of the objective, similarly to~\cite{estlin2001:complementary}. Web service composition indicates that planning under uncertainty is an inevitable necessity for most real-world environments. For example, almost every approach we reviewed executes sensing services during planning only (while making several restricting assumptions). In this context, there are few approaches that augment \HTN~planning to cope with incomplete knowledge, such as~\cite{bouguerra2004:uncertainty}. Planning under uncertainty also considers non-determinism of actions, meaning that actions may have different possible outcomes. The issue of non-determinism is discussed in~\cite{kuter2004:nondeterminism}, which may serve as initial point for further exploration. Having incomplete knowledge and actions that provide information, it is highly possible that an \HTN~planner would have to confront different types of failures and action unavailability. An appropriate reaction to failures, time-outs, external events or changes made by some agents that are beyond the control of the planner is still a real challenge for \HTN~planners. Continual planning~\cite{brenner2009:continual} investigates these problems and stimulates active sensing. Furthermore, a non-classical domain may require to distribute planning over different agents present in the surroundings. In this context, only few studies investigate planning and distribution, both, in \pHTN planning, such as distributed \noah~\cite{corkill1979:dnoah} and distributed \sipe~\cite{desJardins1999:dsipe}, and \tHTN planning, such as A-SHOP~\cite{dix2003:impacting},~\cite{amigoni2005:ambient}, and Planner9~\cite{magnenat2009:planner9}, while most of them provide minor theoretical and practical contributions. In addition, although Web service composition is a real-world challenge, we remark that there are no results on evaluation of the proposed approaches in a real setting. The approaches are tested on simple scenarios, examples or even classical problems that do not necessary lead to conclude that a particular approach will be equally successful or unfavourable in a real setting. Finally, it may seem far from the purpose of automated planning, but creating interfaces that are user-friendly and intuitive could encourage the end-user to consider even more \HTN~planning for solving real-world problems. Instead of configuring files and taking low-level steps to invoke a planner, the users could interact with a graphical interface which would enable them to model a domain, create and store a goal, follow the planning process, or examine the final solution. \sipe~and \oplan~are good examples of \HTN~planners enhanced with components for human-computer interaction.

\section*{Acknowledgement}\small
We are grateful to Luigia Carlucci Aiello and Eirini Kaldeli for the helpful comments on the paper. We would like to thank Austin Tate for valuable comments on the \nonlin~and \oplan~planning systems and Luis Castillo Vidal for the valuable feedback on the \siadex~planning system. The work is supported by the Dutch National Research Council under the NWO Smart Energy Systems program, contract no. 647.000.004.

\bibliographystyle{myIEEEtran}
\bibliography{pita}

\begin{thebibliography}{100}
\providecommand{\url}[1]{#1}
\csname url@samestyle\endcsname
\providecommand{\newblock}{\relax}
\providecommand{\bibinfo}[2]{#2}
\providecommand{\BIBentrySTDinterwordspacing}{\spaceskip=0pt\relax}
\providecommand{\BIBentryALTinterwordstretchfactor}{4}
\providecommand{\BIBentryALTinterwordspacing}{\spaceskip=\fontdimen2\font plus
\BIBentryALTinterwordstretchfactor\fontdimen3\font minus
  \fontdimen4\font\relax}
\providecommand{\BIBforeignlanguage}[2]{{%
\expandafter\ifx\csname l@#1\endcsname\relax
\typeout{** WARNING: IEEEtran.bst: No hyphenation pattern has been}%
\typeout{** loaded for the language `#1'. Using the pattern for}%
\typeout{** the default language instead.}%
\else
\language=\csname l@#1\endcsname
\fi
#2}}
\providecommand{\BIBdecl}{\relax}
\BIBdecl

\bibitem{nau2004:automated}
M.~Ghallab, D.~S. Nau, and P.~Traverso, \emph{Automated {P}lanning: {T}heory \&
  {P}ractice}.\hskip 1em plus 0.5em minus 0.4em\relax Morgan Kaufmann
  Publishers Inc., 2004.

\bibitem{nau1999:shop}
D.~S. Nau, Y.~Cao, A.~Lotem, and H.~Mu\~{n}oz Avila, ``{SHOP: Simple
  Hierarchical Ordered Planner},'' in \emph{Proceedings of the 16th
  International Joint Conference on Artificial Intelligence}, ser.
  IJCAI'99.\hskip 1em plus 0.5em minus 0.4em\relax Morgan Kaufmann Publishers
  Inc., 1999, pp. 968--975.

\bibitem{bacchus2001:aips}
F.~Bacchus, ``{The AIPS '00 Planning Competition},'' \emph{AI Magazine},
  vol.~22, no.~3, pp. 47--56, 2001.

\bibitem{nau2007:trends}
D.~S. Nau, ``{Current Trends in Automated Planning},'' \emph{AI Magazine},
  vol.~28, no.~4, pp. 43--58, 2007.

\bibitem{sacerdoti1975:structure}
E.~D. Sacerdoti, ``{A Structure for Plans and Behavior},'' Ph.D. dissertation,
  Standfor University, AI Center, 1975, aAI7605794.

\bibitem{sacerdoti1975:nonlinear}
E.~D. Sacerdoti, ``{The Nonlinear Nature of Plans},'' in \emph{Proceedings of
  the 4th international Joint Conference on Artificial Intelligence - Volume
  1}, ser. IJCAI'75.\hskip 1em plus 0.5em minus 0.4em\relax Morgan Kaufmann
  Publishers Inc., 1975, pp. 206--214.

\bibitem{tate1976:project}
A.~Tate, ``{Project Planning Using a Hierarchic Non-linear Planner},''
  Department of Artificial Intelligence, University of Edinburgh, Research
  Report~25, aug 1976.

\bibitem{tate1977:generating}
A.~Tate, ``{Generating Project Networks},'' in \emph{Proceedings of the 5th
  International Joint Conference on Artificial Intelligence - Volume 2}, ser.
  IJCAI'77.\hskip 1em plus 0.5em minus 0.4em\relax Morgan Kaufmann Publishers
  Inc., 1977, pp. 888--893.

\bibitem{wilkins1991:practical}
D.~E. Wilkins, ``{Can AI Planners Solve Practical Problems?}'' \emph{Comput.
  Intell.}, vol.~6, pp. 232--246, January 1991.

\bibitem{currie1991:oplan}
K.~Currie and A.~Tate, ``{O-Plan: The Open Planning Architecture},''
  \emph{Artificial Intelligence}, vol.~52, no.~1, pp. 49--86, Nov. 1991.

\bibitem{tate1994:oplan2}
A.~Tate, B.~Drabble, and R.~Kirby, ``{O-Plan2: An Open Architecture for
  Command, Planning and Control},'' in \emph{Intelligent Scheduling}.\hskip 1em
  plus 0.5em minus 0.4em\relax Morgan Kaufmann Publishers Inc., 1994, pp.
  213--239.

\bibitem{erol1996:formalization}
K.~Erol, ``{Hierarchical Task Network Planning: Formalization, Analysis, and
  Implementation},'' Ph.D. dissertation, Computer Science Department,
  University of Maryland, 1996.

\bibitem{nau2003:shop2}
D.~S. Nau, O.~Ilghami, U.~Kuter, J.~W. Murdock, D.~Wu, and F.~Yaman, ``{SHOP2:
  An HTN Planning System},'' \emph{Journal of Artificial Intelligence
  Research}, vol.~20, no.~1, pp. 379--404, Dec. 2003.

\bibitem{castillo2005:temporal_enhancements}
L.~A. Castillo, J.~Fern{\'a}ndez-Olivares, {\'O}.~Garc{\'i}a-P{\'e}rez, and
  F.~Palao, ``{Temporal Enhancements of an HTN Planner},'' in \emph{Proceedings
  of the 11th Conference of the Spanish Association for Artificial
  Intelligence, Current Topics in AI}.\hskip 1em plus 0.5em minus 0.4em\relax
  Springer-Verlag, 2005, pp. 429--438.

\bibitem{papazoglou2003:soc}
M.~P. Papazoglou and D.~Georgakopoulos, ``{Introduction: Service-Oriented
  Computing},'' \emph{Comm. ACM}, vol.~46, no.~10, pp. 24--28, oct 2003.

\bibitem{lazovik2004:assertions}
A.~Lazovik, M.~Aiello, and M.~Papazoglou, ``{Associating Assertions with
  Business Processes and Monitoring Their Execution},'' in \emph{Proceedings of
  the 2nd international conference on Service oriented computing}, ser. ICSOC
  '04.\hskip 1em plus 0.5em minus 0.4em\relax ACM, 2004, pp. 94--104.

\bibitem{dustdar2005:surveywsc}
S.~Dustdar and W.~Schreiner, ``{A Survey on Web Services Composition},''
  \emph{Int. J. Web Grid Serv.}, vol.~1, no.~1, pp. 1--30, aug 2005.

\bibitem{kaldeli2013:dynamicuncertain}
E.~Kaldeli, ``{Domain-Independent Planning for Services in Uncertain and
  Dynamic Environments},'' Ph.D. dissertation, Faculty of Mathematics and
  Natural Sciences, University of Groningen, 2013.

\bibitem{erol1994:umcp}
K.~Erol, J.~Hendler, and D.~S. Nau, ``{UMCP: A Sound and Complete Procedure for
  Hierarchical Task-Network Planning},'' in \emph{International Conference on
  AI Planning Systems}.\hskip 1em plus 0.5em minus 0.4em\relax Morgan Kaufmann
  Publishers Inc., 1994, pp. 249--254.

\bibitem{geier2011:decidability}
T.~Geier and P.~Bercher, ``{On the Decidability of HTN Planning with Task
  Insertion},'' in \emph{Proceedings of the Twenty-Second international joint
  conference on Artificial Intelligence - Volume Volume Three}, ser.
  IJCAI'11.\hskip 1em plus 0.5em minus 0.4em\relax AAAI Press, 2011, pp.
  1955--1961.

\bibitem{stefik1981:constraints}
M.~Stefik, ``{Planning with Constraints (MOLGEN: Part 1)},'' \emph{Artificial
  Intelligence}, vol.~16, no.~2, pp. 111--140, 1981.

\bibitem{weld1994:least_commitment}
D.~S. Weld, ``{An Introduction to Least Commitment Planning},'' \emph{AI
  Magazine}, vol.~15, no.~4, pp. 27--61, 1994.

\bibitem{tsuneto1996:commitment}
R.~Tsuneto, K.~Erol, J.~Hendler, and D.~S. Nau, ``{Commitment Strategies in
  Hierarchical Task Network Planning},'' in \emph{Proceedings of the Thirteenth
  National Conference on Artificial Intelligence - Volume 1}, ser.
  AAAI'96.\hskip 1em plus 0.5em minus 0.4em\relax AAAI, 1996, pp. 536--542.

\bibitem{foulser1992:merging}
D.~E. Foulser, M.~Li, and Q.~Yang, ``{Theory and Algorithms for Plan
  Merging},'' \emph{Artificial Intelligence}, vol.~57, pp. 143--181, October
  1992.

\bibitem{yang1992:conflict}
Q.~Yang, ``{A Theory of Conflict Resolution in Planning},'' \emph{Artificial
  Intelligence}, vol.~58, no.~1, pp. 361--392, 1992.

\bibitem{wilkins1988:extending}
D.~E. Wilkins, \emph{Practical {P}lanning: {E}xtending the {C}lassical {AI}
  {P}lanning {P}aradigm}.\hskip 1em plus 0.5em minus 0.4em\relax Morgan
  Kaufmann Publishers Inc., 1988.

\bibitem{erol1994:semanticshtn}
K.~Erol, J.~Hendler, and D.~S. Nau, ``{Semantics for Hierarchical Task-Network
  Planning},'' University of Maryland, Institute for Advanced Computer Studies,
  Tech. Rep. UMIACS-TR-94-31, 1994.

\bibitem{nareyek2005:constraints}
A.~Nareyek, E.~C. Freuder, R.~Fourer, E.~Giunchiglia, R.~P. Goldman, H.~Kautz,
  J.~Rintanen, and A.~Tate, ``{Constraints and AI Planning},'' \emph{IEEE
  Intelligent Systems}, vol.~20, pp. 62--72, March 2005.

\bibitem{georgievski2011:phantomisation}
I.~Georgievski, A.~Lazovik, and M.~Aiello, ``{Task Interaction in an HTN
  Planner},'' CoRR, Tech. Rep., 2011.

\bibitem{fikes1971:strips}
R.~E. Fikes and N.~J. Nilsson, ``{STRIPS: A New Approach to the Application of
  Theorem Proving to Problem Solving},'' in \emph{Proceedings of the 2nd
  International Joint Conference on Artificial Intelligence}, ser.
  IJCAI'71.\hskip 1em plus 0.5em minus 0.4em\relax Morgan Kaufmann Publishers
  Inc., 1971, pp. 608--620.

\bibitem{kambhampati1995:comparative}
S.~Kambhampati, ``{A Comparative Analysis of Partial Order Planning and Task
  Reduction Planning},'' \emph{SIGART Bull.}, vol.~6, pp. 16--25, January 1995.

\bibitem{collins1992:filter}
G.~Collins and L.~Pryor, ``{Achieving the Functionality of Filter Conditions in
  a Partial-Order Planner},'' in \emph{Proceedings of the 10th National
  Conference on Artificial Intelligence}, ser. AAAI'92.\hskip 1em plus 0.5em
  minus 0.4em\relax AAAI, 1992, pp. 375--380.

\bibitem{alford2012:htnspaces}
R.~Alford, V.~Shivashankar, U.~Kuter, and D.~S. Nau, ``{HTN Problem Spaces:
  Structure, Algorithms, Termination},'' in \emph{Proceedings of the Fifth
  Annual Symposium on Combinatorial Search}.\hskip 1em plus 0.5em minus
  0.4em\relax AAAI Press, 2012, pp. 2--9.

\bibitem{chapman1987:conjunctive}
D.~Chapman, ``{Planning for Conjunctive Goals},'' \emph{Artificial
  Intelligence}, vol.~32, no.~3, pp. 333--377, Jul. 1987.

\bibitem{wilkins2000:sipe2_manual}
D.~E. Wilkins, ``{Using the SIPE-2 Planning System - A Manual for SIPE-2},''
  2000.

\bibitem{tate1994:conditions}
A.~Tate, B.~Drabble, and J.~Dalton, ``{The Use of Condition Types to Restrict
  Search in an AI Planner},'' in \emph{Proceedings of the Twelfth National
  Conference on Artificial Intelligence - Volume 2}, ser. AAAI'94.\hskip 1em
  plus 0.5em minus 0.4em\relax AAAI, 1994, pp. 1129--1134.

\bibitem{tsuneto1998:external_conditions}
R.~Tsuneto, J.~Hendler, and D.~S. Nau, ``{Analyzing External Conditions to
  Improve the Efficiency of HTN Planning},'' in \emph{Proceedings of the 15th
  National/10th Conference on Artificial Intelligence/Innovative Applications
  of Artificial Intelligence}.\hskip 1em plus 0.5em minus 0.4em\relax AAAI,
  1998, pp. 913--920.

\bibitem{shop}
``{SHOP, SHOP2, JSHOP and JSHOP-2 Planners},'' {O}nline: accessed {J}an. 2014,
  http://sourceforge.net/projects/shop/.

\bibitem{python}
``{Python 3.4.0},'' {O}nline: accessed {J}an. 2014,
  http://www.python.org/getit/releases/3.4.0/.

\bibitem{wilkins2001:call}
D.~E. Wilkins and M.~Desjardins, ``{A Call for Knowledge-Based Planning},''
  \emph{AI Magazine}, vol.~22, no.~1, pp. 99--115, 2001.

\bibitem{long2003:ipc3}
D.~Long and M.~Fox, ``{The 3rd International Planning Competition: Results and
  Analysis},'' \emph{J. Artif. Intell. Res. (JAIR)}, vol.~20, pp. 1--59, 2003.

\bibitem{erol1994:expressivity}
K.~Erol, J.~A. Hendler, and D.~S. Nau, ``{HTN Planning: Complexity and
  Expressivity},'' in \emph{Proceedings of the 12th National Conference on
  Artificial Intelligence - Volume 2}, ser. AAAI'94.\hskip 1em plus 0.5em minus
  0.4em\relax AAAI, 1994, pp. 1123--1128.

\bibitem{nau1998:control}
D.~S. Nau, S.~J. Smith, and K.~Erol, ``{Control Strategies in HTN Planning:
  Theory Versus Practice},'' in \emph{Proceedings of the 15th National/10th
  conference on Artificial Intelligence/Innovative applications of Artificial
  Intelligence}, ser. AAAI'98/IAAI'98.\hskip 1em plus 0.5em minus 0.4em\relax
  AAAI, 1998, pp. 1127--1133.

\bibitem{lekavy2007:expressivity}
M.~Lekav\'{y} and P.~N\'{a}vrat, ``{Expressivity of STRIPS-Like and HTN-Like
  Planning},'' in \emph{Proceedings of the 1st KES International Symposium on
  Agent and Multi-Agent Systems: Technologies and Applications}, ser.
  KES-AMSTA'07.\hskip 1em plus 0.5em minus 0.4em\relax Springer-Verlag, 2007,
  pp. 121--130.

\bibitem{erol1995:complexity}
K.~Erol, J.~Handler, and D.~S. Nau, ``{Complexity Results for HTN Planning},''
  \emph{Annals of Mathematics and AI}, vol.~18, no.~1, pp. 69--93, 1996.

\bibitem{ghallab1998:pddl}
M.~Ghallab, C.~K. Isi, S.~Penberthy, D.~E. Smith, Y.~Sun, and D.~Weld, ``{PDDL
  - The Planning Domain Definition Language},'' CVC TR-98-003/DCS TR-1165, Yale
  Center for Computational Vision and Control, Tech. Rep., 1998.

\bibitem{fox2003:pddl21}
M.~Fox and D.~Long, ``{PDDL2.1: An Extension to PDDL for Expressing Temporal
  Planning Domains},'' \emph{Journal of Artificial Intelligence Research},
  vol.~20, no.~1, pp. 61--124, Dec. 2003.

\bibitem{gerevini2006:pddl3}
A.~Gerevini and D.~Long, ``{Preferences and Soft Constraints in PDDL3},'' in
  \emph{Proceedings of the ICAPS Workshop on Planning with Preferences and Soft
  Constraints}, 2006.

\bibitem{agosta1996:oilspill}
J.~M. Agosta, ``{Constraining Influence Diagram Structure by Generative
  Planning: An Application to the Optimization of Oil Spill Response},'' in
  \emph{Proceedings of the Twelfth international Conference on Uncertainty in
  Artificial Intelligence}, ser. UAI'96.\hskip 1em plus 0.5em minus 0.4em\relax
  Morgan Kaufmann Publishers Inc., 1996, pp. 11--19.

\bibitem{aarup1995:optimum}
M.~Aarup, M.~M. Arentoft, Y.~Parrod, I.~Stokes, H.~Vadon, and J.~Stader,
  ``{OPTIMUM-AIV: A Knowledge-Based Planning and Scheduling System for
  Spacecraft AIV},'' in \emph{Proceedings of the Conference on Knowledge Based
  Scheduling}.\hskip 1em plus 0.5em minus 0.4em\relax Morgan Kaufmann
  Publishers Inc., 1994, pp. 451--469.

\bibitem{smith1997:electrical}
S.~J.~J. Smith, K.~Hebbar, D.~S. Nau, and I.~Minis, ``{Integrating Electrical
  and Mechanical Design and Process Planning},'' 1997.

\bibitem{kaldeli2012:coorindating}
E.~Kaldeli, E.~U. Warriach, A.~Lazovik, and M.~Aiello, ``{Coordinating the Web
  of Services for a Smart Home},'' \emph{ACM Transactions on the Web}, vol.~7,
  no.~2, pp. 10:1--10:40, 2012.

\bibitem{kuter2005:gathering}
U.~Kuter, E.~Sirin, B.~Parsia, D.~Nau, and J.~Hendler, ``{Information Gathering
  during Planning for Web Service Composition},'' \emph{Web Semantic}, vol.~3,
  pp. 183--205, October 2005.

\bibitem{nau2005:applications}
D.~S. Nau, T.~C. Au, O.~Ilghami, U.~Kuter, D.~Wu, F.~Yaman, H.~Mu\~{n}oz Avila,
  and J.~W. Murdock, ``{Applications of SHOP and SHOP2},'' \emph{IEEE
  Intelligent Systems}, vol.~20, no.~2, pp. 34--41, march-april 2005.

\bibitem{schattenberg2009:hybrid}
B.~Schattenberg, ``Hybrid planning and scheduling,'' Ph.D. dissertation,
  Institute of Artificial Intelligence, Ulm University, 2009.

\bibitem{myers1996:advice}
K.~L. Myers, ``{Strategic Advice for Hierarchical Planners},'' in
  \emph{Proceedings of the 5th International Conference on Principles of
  Knowledge Representation and Reasoning}.\hskip 1em plus 0.5em minus
  0.4em\relax Morgan Kaufmann Publishers Inc., 1996, pp. 112--123.

\bibitem{myers2000:conflicting}
K.~L. Myers, ``{Planning with Conflicting Advice},'' in \emph{Proceedings of
  the 5th International Conference on Artificial Intelligence Planning
  Systems}.\hskip 1em plus 0.5em minus 0.4em\relax AAAI, 2000, pp. 355--362,
  poster Paper.

\bibitem{drabble1995:tfmanual}
B.~Drabble and A.~Tate, ``{O-Plan: Task Formalism Manual Version 2.3},'' AIAI,
  University of Edinburgh, Tech. Rep., jul 1995.

\bibitem{sohrabi2008:onpreferences}
S.~Sohrabi and S.~A. Mcilraith, ``{On Planning with Preferences in HTN},'' in
  \emph{Proceedings of the 12th International Workshop on Non-Monotonic
  Reasoning}, 2008, pp. 241--248.

\bibitem{sohrabi2009:preferences}
S.~Sohrabi, J.~A. Baier, and S.~A. McIlraith, ``{HTN Planning with
  Preferences},'' in \emph{Proceedings of the 21st International Joint
  Conference on Artifical Intelligence}, ser. IJCAI'09.\hskip 1em plus 0.5em
  minus 0.4em\relax Morgan Kaufmann Publishers Inc., 2009, pp. 1790--1797.

\bibitem{stillman1993:tachyon}
J.~Stillman, R.~Arthur, and A.~Deitsch, ``{Tachyon: A Constraint-based Temporal
  Reasoning Model and Its Implementation},'' \emph{SIGART Bull.}, vol.~4,
  no.~3, pp. 1--4, jul 1993.

\bibitem{allen1983:temporal}
J.~F. Allen, ``{Maintaining Knowledge about Temporal Intervals},'' \emph{Comm.
  ACM}, vol.~26, no.~11, pp. 832--843, 1983.

\bibitem{benthem1991:logictime}
J.~F. A. K.~v. Benthem, \emph{The {L}ogic of {T}ime : {A} {M}odel-{T}heoretic
  {I}nvestigation into the {V}arieties of {T}emporal {O}ntology and {T}emporal
  {D}iscourse}, 2nd~ed.\hskip 1em plus 0.5em minus 0.4em\relax Kluwer Academic
  Publishers Dordrecht, 1991.

\bibitem{yaman2002:timeline}
F.~Yaman and D.~S. Nau, ``{Timeline: An HTN Planner that can Reason about
  Time},'' in \emph{Proceedings of the AIPS'02 Workshop on Planning for
  Temporal Domains}, 2002, pp. 75--81.

\bibitem{goldman2006:durative}
R.~P. Goldman, ``{Durative Planning in HTNs},'' in \emph{Proceedings of the
  16th International Conference on Automated Planning and Scheduling}, ser.
  ICAPS'06.\hskip 1em plus 0.5em minus 0.4em\relax AAAI, 2006, pp. 382--385.

\bibitem{castillo2006:handling}
L.~A. Castillo, J.~Fern{\'a}ndez-Olivares, {\'O}.~Garc{\'i}a-P{\'e}rez, and
  F.~Palao, ``{Efficiently Handling Temporal Knowledge in an HTN Planner},'' in
  \emph{Proceedings of the 16th International Conference on Automated Planning
  and Scheduling}, ser. ICAPS'06.\hskip 1em plus 0.5em minus 0.4em\relax AAAI,
  2006, pp. 63--72.

\bibitem{kartam1989:foundation}
N.~A. Kartam and D.~E. Wilkins, ``{Toward a Foundation for Evaluating AI
  Planners},'' AI Center, SRI International, Tech. Rep., 1989.

\bibitem{gomes1994:comparison}
C.~Gomes, ``O-plan2 vs sipe-2 - a general comparison,'' Rome Laboratory, Tech.
  Rep., jul 1996.

\bibitem{fdez-olivares2011:tnmodeling}
J.~Fern{\'a}ndez-Olivares, I.~S\'{a}nchez-Garz\'{o}n, A.~Gonz\'{a}lez-Ferrer,
  J.~A. C\'{o}zar, A.~Fdez-Teijeiro, M.~R. Cabello, and L.~Castillo, ``{Task
  Network Based Modeling, Dynamic Generation and Adaptive Execution of
  Patient-tailored Treatment Plans Based on Smart Process Management
  Technologies},'' in \emph{Proceedings of the 3rd International Conference on
  Knowledge Representation for Health-Care}, ser. KR4HC'11.\hskip 1em plus
  0.5em minus 0.4em\relax Springer-Verlag, 2012, pp. 37--50.

\bibitem{nau2000:mshop}
D.~S. Nau, Y.~Cao, A.~Lotem, and H.~Mu\~{n}oz Avila, ``{SHOP and M-SHOP:
  Planning with Ordered Task Decomposition},'' Computer Science Department,
  University of Maryland, Tech. Rep. CS-TR-4157, jun 2000.

\bibitem{andrews1995:umtranslog}
S.~Andrews, B.~Kettler, K.~Erol, and J.~Hendler, ``{UM Translog: A Planning
  Domain for the Development and Benchmarking of Planning Systems},'' Computer
  Science Department, University of Maryland, Tech. Rep. CS-TR-3487, 1995.

\bibitem{fdez-olivares2008:oncology}
J.~Fern{\'a}ndez-Olivares, J.~A. C\'{o}zar, and L.~Castillo, ``{Automating
  Oncology Therapy Plans by means of Temporal Hierarchical Task Networks
  Planning},'' in \emph{Proceedings of the ECAI Workshop on Knowledge
  Management for Healthcare Processes}, 2008.

\bibitem{fdez-olivares2011:clinical}
J.~Fdez-Olivares, L.~Castillo, J.~A. C\'{o}zar, and O.~Garc{\'i}a~P\'{e}rez,
  ``{Supporting Clinical Processes and Decisions by Hierarchical Planning and
  Scheduling},'' \emph{Computational Intelligence}, vol.~27, no.~1, pp.
  103--122, 2011.

\bibitem{castillo2010:elearning}
L.~Castillo, L.~Morales, A.~Gonz\'{a}lez-Ferrer, J.~Fern{\'a}ndez-Olivares,
  D.~Borrajo, and E.~Onaind\'{\i}a, ``{Automatic Generation of Temporal
  Planning Domains for e-Learning Problems},'' \emph{J. of Scheduling},
  vol.~13, no.~4, pp. 347--362, aug 2010.

\bibitem{garrido2013:elearning}
A.~Garrido, S.~Fern{\'a}ndez, L.~Morales, E.~Onaindia, D.~Borrajo, and
  L.~Castillo, ``{On the Automatic Compilation of e-Learning Models to
  Planning},'' \emph{Knowledge Eng. Review}, vol.~28, no.~2, pp. 121--136,
  2013.

\bibitem{castillo2008:touristvisit}
L.~Castillo, E.~Armengol, E.~Onaind\'{\i}a, L.~Sebasti\'{a},
  J.~Gonz\'{a}lez-Boticario, A.~Rodr\'{\i}guez, S.~Fern\'{a}ndez, J.~D. Arias,
  and D.~Borrajo, ``{Samap: An User-oriented Adaptive System for Planning
  Tourist Visits},'' \emph{Expert Syst. Appl.}, vol.~34, no.~2, pp. 1318--1332,
  2008.

\bibitem{palao2011:touristvisit}
F.~Palao, L.~Castillo, J.~Fern{\'a}ndez-Olivares, and O.~Garc{\'i}a, ``{Cities
  That Offer a Customized and Personalized Tourist Experience to Each and Every
  Visitor: The Smartourism Project},'' in \emph{Proceedings of the AI for an
  Intelligent Planet}.\hskip 1em plus 0.5em minus 0.4em\relax ACM, 2011, pp.
  4:1--4:8.

\bibitem{kartam1990:approach}
N.~A. Kartam, R.~E. Levitt, and D.~E. Wilkins, ``{A Centralized Approach for
  Representing and Resolving Interactions among Multi-Agent Tasks while
  Planning Hierarchically},'' in \emph{Proceedings of the 6th Conference on
  Artificial Intelligence Applications}.\hskip 1em plus 0.5em minus 0.4em\relax
  IEEE Press, 1990, pp. 250--256.

\bibitem{munoz1999:hicap}
H.~Mu{\~n}oz-Avila, D.~W. Aha, L.~Breslow, and D.~S. Nau, ``{HICAP: An
  Interactive Case-Based Planning Architecture and its Application to
  Noncombatant Evacuation Operations},'' 1999.

\bibitem{munoz2001:sin}
H.~Mu{\~n}oz-Avila, D.~W. Aha, D.~S. Nau, R.~Weber, L.~Breslow, and F.~Yamal,
  ``{SiN: Integrating Case-Based Reasoning with Task Decomposition},'' in
  \emph{Proceedings of the 17th International Joint Conference on Artificial
  Intelligence - Volume 2}, ser. IJCAI'01.\hskip 1em plus 0.5em minus
  0.4em\relax Morgan Kaufmann Publishers Inc., 2001, pp. 999--1004.

\bibitem{wu2003:daml}
D.~Wu, B.~Parsia, E.~Sirin, J.~Hendler, and D.~S. Nau, ``{Automating DAML-S Web
  Services Composition Using SHOP2},'' in \emph{Proceedings of 2nd
  International Semantic Web Conference}, ser. ISWC'03.\hskip 1em plus 0.5em
  minus 0.4em\relax Springer-Verlag, 2003, pp. 195--210.

\bibitem{sirin2004:shop2_composition}
E.~Sirin, B.~Parsia, D.~Wu, J.~Hendler, and D.~S. Nau, ``{HTN Planning for Web
  Service Composition Using SHOP2},'' \emph{Web Semantic}, vol.~1, pp.
  377--396, October 2004.

\bibitem{drummond1988:spacecraft}
M.~E. Drummond, K.~W. Currie, and A.~Tate, ``{O-Plan Meets T-SAT: First Results
  From the Application of an AI Planner to Spacecraft Mission Sequencing},''
  Artificial Intelligence Applications Institute, University of Edinburgh,
  Tech. Rep., 1988.

\bibitem{drabble1997:repairingplans}
B.~Drabble, J.~Dalton, and A.~Tate, ``{Repairing Plans On-the-fly},'' in
  \emph{Proceedings of the NASA Workshop on P\&S for Space}, 1997.

\bibitem{tate1996:oplanlogistics}
A.~Tate, B.~Drabble, and J.~Dalton, ``{O-Plan: A Knowledge-Based Planner and
  its Application to Logistics},'' in \emph{Proceedings of ARPI}, 1996, pp.
  259--266.

\bibitem{khan2003:biological}
S.~Khan, W.~Gillis, C.~Schmidt, and K.~Decker, ``{A Multi-Agent System-driven
  AI Planning Approach to Biological Pathway Discovery},'' in \emph{Proceedings
  of the 13th International Conference on Automated Planning and Scheduling)},
  ser. ICAPS'03.\hskip 1em plus 0.5em minus 0.4em\relax AAAI, 2003, pp.
  246--255.

\bibitem{lee1996:military}
T.~J. Lee and D.~Wilkins, ``{Using SIPE-2 to Integrate Planning for Military
  Air Campaigns},'' \emph{IEEE Expert}, vol.~11, no.~6, pp. 11--12, Dec 1996.

\bibitem{tate1998:qualitative}
A.~Tate, J.~Dalton, and J.~Levine, ``{Generation of Multiple Qualitatively
  Different Plan Options},'' in \emph{Proceedings of the 4th International
  Conference on AI Planning Systems}.\hskip 1em plus 0.5em minus 0.4em\relax
  AAAI, 1998, pp. 27--35.

\bibitem{gonzalez-ferrer2013:bpm}
A.~Gonz{\'a}lez-Ferrer, J.~Fern{\'a}ndez-Olivares, and L.~Castillo, ``{From
  Business Process Models to Hierarchical Task Network Planning Domains},''
  \emph{Knowledge Eng. Review}, vol.~28, no.~2, pp. 175--193, 2013.

\bibitem{sanchez-garzon2013:carepathways}
I.~S\'{a}nchez-Garz\'{o}n, J.~Fern{\'a}ndez-Olivares, and L.~Castillo, ``{An
  Approach for Representing and Managing Medical Exceptions in Care Pathways
  Based on Temporal Hierarchical Planning Techniques},'' in \emph{Process
  Support and Knowledge Representation in Health Care}, ser. Lecture Notes in
  Computer Science, R.~Lenz, S.~Miksch, M.~Peleg, M.~Reichert, D.~Riano, and
  A.~Teije, Eds.\hskip 1em plus 0.5em minus 0.4em\relax Springer, 2013, vol.
  7738, pp. 168--182.

\bibitem{tate1996:responsive}
A.~Tate, ``{Responsive Planning and Scheduling using AI Planning Techniques -
  Optimum-AIV},'' \emph{IEEE Expert: Intelligent Systems \& their
  Applications}, vol.~11, no.~6, pp. 4--12, 1996.

\bibitem{agosta1995:formulation}
J.~M. Agosta, ``{Formulation and Implementation of an Equipment Configuration
  Problem with the (SIPE-2) Generative Planner},'' in \emph{Proceedings of the
  AAAI-95 Spring Symposium on Integrated Planning Applications}, 1995, pp.
  1--10.

\bibitem{tate2000:webbased}
A.~Tate, J.~Dalton, and J.~Levine, ``{O-Plan: A Web-Based AI Planning Agent},''
  in \emph{Proceedings of the 17th National Conference on Artificial
  Intelligence and 12th Conference on Innovative Applications of Artificial
  Intelligence}, ser. AAAI'00/IAAI'00.\hskip 1em plus 0.5em minus 0.4em\relax
  AAAI, 2000, pp. 1131--1132.

\bibitem{delaasuncion2005:siadex}
M.~de~la Asunci\'{o}n, L.~Castillo, J.~Fern\'{a}dez-Olivares,
  O.~Garc{\'i}a-P\'{e}rez, A.~Gonz\'{a}lez, and F.~Palao, ``{SIADEX: An
  Interactive Artificial Intelligence Planner for Decision Support and Training
  in Forest Fire Fighting},'' \emph{Artificial Intelligence Communications},
  vol.~18, no.~4, pp. 257--268, 2005.

\bibitem{wilkins1992:military}
D.~Wilkins and R.~V. Desimone, ``{Applying an AI Planner to Military Operations
  Planning},'' in \emph{Intelligent Scheduling}.\hskip 1em plus 0.5em minus
  0.4em\relax Morgan Kaufmann Publishers Inc., 1992, pp. 685--709.

\bibitem{sipe2:oilspil}
``{Oil Spill Response Planning in SIPE-2},'' {O}nline: accessed {J}an. 2014,
  http://www.ai.sri.com/~sipe/oil.html.

\bibitem{alvey1987:clubs}
A.~Directorate, ``{Alvey Grand Meeting of Community Clubs},'' Available through
  Institution of Electrical Engineering, Tech. Rep., 1987.

\bibitem{kingston1996:commonkadsmodels}
J.~Kingston, N.~Shadbolt, and A.~Tate, ``{CommonKADS Models for Knowledge Based
  Planning},'' in \emph{Proceedings of the National Conference on Artifical
  Intelligence}, ser. AAAI'96.\hskip 1em plus 0.5em minus 0.4em\relax AAAI,
  1996, pp. 11--6.

\bibitem{tate1990:interfacing}
A.~Tate, \emph{Interfacing a {CAD} {S}ystem to an {AI} {P}lanner}, ser.
  AIAI-TR-76.\hskip 1em plus 0.5em minus 0.4em\relax Artificial Intelligence
  Applications Institute, Univ. of Edinburgh, 1990.

\bibitem{tate2003:web}
A.~Tate and J.~Dalton, ``{O-Plan: a Common Lisp Planning Web Service},'' in
  \emph{Proceedings of the International Lisp Conference}, 2003.

\bibitem{shop2:comets}
``{Real-time coordination and control of multiple heterogeneous UAV},''
  {O}nline: accessed {J}an. 2014, http://www.comets-uavs.org/.

\bibitem{shop2:sift}
``{Smart Information Flow Technologies - Planning},'' {O}nline: accessed {J}an.
  2014, http://www.sift.net/projects/planning.

\bibitem{tate2000:usingai}
A.~Tate, J.~Levine, P.~Jarvis, and J.~Dalton, ``{Using AI Planning Technology
  for Army Small Unit Operations},'' in \emph{Proceedings of the Artificial
  Intelligence Planning and Scheduling Systems Conference}.\hskip 1em plus
  0.5em minus 0.4em\relax AAAI, 2000, pp. 379--386, poster Paper.

\bibitem{uszok2004:applying}
A.~Uszok, J.~M. Bradshaw, R.~Jeffers, A.~Tate, and J.~Dalton, ``{Applying KAoS
  Services to Ensure Policy Compliance for Semantic Web Services Workflow
  Composition and Enactment},'' in \emph{International Semantic Web
  Conference}.\hskip 1em plus 0.5em minus 0.4em\relax Springer-Verlag, 2004,
  pp. 425--440.

\bibitem{fdez-olivares2007:wsc}
J.~Fern\'{a}ndez-Olivares, T.~Garz\'{o}n, L.~Castillo, O.~Garc\'{i}a-P\'{e}rez,
  and F.~Palao, ``{A Middle-Ware for the Automated Composition and Invocation
  of Semantic Web Services Based on Temporal HTN Planning Techniques},'' in
  \emph{Current Topics in Artificial Intelligence}, ser. Lecture Notes in
  Computer Science, D.~Borrajo, L.~Castillo, and J.~Corchado, Eds.\hskip 1em
  plus 0.5em minus 0.4em\relax Springer, 2007, vol. 4788, pp. 70--79.

\bibitem{brian2009:wscreview}
S.~Bleul, T.~Weise, and K.~Geihs, ``{The Web Service Challenge -- A Review on
  Semantic Web Service Composition},'' in \emph{Service-Oriented Computing
  (SOC'2009)}, Mar 2009, pp. 1--12.

\bibitem{wschallenge2010}
``The {W}eb {S}ervices {C}hallenge,'' {O}nline: accessed {J}an. 2014,
  http://www.wschallenge.org/.

\bibitem{blake2006:wsc}
M.~B. Blake, W.~Cheung, M.~C. Jaeger, and A.~Wombacher, ``{WSC}-06: {T}he {W}eb
  {S}ervice {C}hallenge,'' in \emph{Joint Proceedings of the CEC/EEE 2006},
  2006, pp. 1--2.

\bibitem{aiello:2008:visualizing}
M.~Aiello, N.~van Benthem, and E.~el~Khoury, ``{Visualizing Compositions of
  Services from Large Repositories},'' in \emph{Joint Proceedings of the IEEE
  CEC/EEE 2008}, 2008, pp. 359--362.

\bibitem{aiello2009:optimal}
M.~Aiello, E.~el~Khoury, A.~Lazovik, and P.~Ratelband, ``{Optimal QoS-Aware Web
  Service Composition},'' in \emph{Joint Proceedings of the IEEE CEC/EEE 2009},
  2009, pp. 491--494.

\bibitem{degeler2010:concept}
V.~Degeler, I.~Georgievski, A.~Lazovik, and M.~Aiello, ``{Concept Mapping for
  Faster QoS-Aware Web Service Composition},'' in \emph{IEEE Conference on
  Service Oriented Computing and Applications}, 2010, pp. 1--4.

\bibitem{aiello2002:request}
M.~Aiello, M.~P. Papazoglou, J.~Yang, M.~Carman, M.~Pistore, L.~Serafini, and
  P.~Traverso, ``{A Request Language for Web-Services Based on Planning and
  Constraint Satisfaction},'' in \emph{Proceedings of the Third International
  Workshop on Technologies for E-Services}.\hskip 1em plus 0.5em minus
  0.4em\relax Springer-Verlag, 2002, pp. 76--85.

\bibitem{medjahed2005:multilevel}
B.~Medjahed and A.~Bouguettaya, ``{A Multilevel Composability Model for
  Semantic Web Services},'' \emph{IEEE Transactions on Knowledge and Data
  Engineering}, vol.~17, pp. 954--968, July 2005.

\bibitem{klusch2005:semanticwsc}
M.~Klusch and A.~Gerber, ``{Semantic Web Service Composition Planning with
  OWLS-XPlan},'' in \emph{Proceedings of the 1st International AAAI Fall
  Symposium on Agents and the Semantic Web}.\hskip 1em plus 0.5em minus
  0.4em\relax AAAI, 2005, pp. 55--62.

\bibitem{sohrabi2006:wsc}
S.~Sohrabi, N.~Prokoshyna, and S.~A. Mcilraith, ``{Web Service Composition via
  Generic Procedures and Customizing User Preferences},'' in \emph{Proceedings
  of the International Semantic Web Conference}, ser. ISWC'06.\hskip 1em plus
  0.5em minus 0.4em\relax Springer-Verlag, 2006, pp. 597--611.

\bibitem{paik2007:wscomposition}
I.~Paik and D.~Maruyama, ``{Automatic Web Services Composition Using Combining
  HTN and CSP},'' in \emph{Proceedings of the 7th IEEE International Conference
  on Computer and Information Technology}.\hskip 1em plus 0.5em minus
  0.4em\relax IEEE Computer Society, 2007, pp. 206--211.

\bibitem{kaldeli2009:extended}
E.~Kaldeli, A.~Lazovik, and M.~Aiello, ``{Extended Goals for Composing
  Services},'' in \emph{Proceedings of the 19th International Conference on
  Automated Planning and Scheduling}, ser. ICAPS'09.\hskip 1em plus 0.5em minus
  0.4em\relax AAAI, 2009, pp. 362--365.

\bibitem{kaldeli2011:continual}
E.~Kaldeli, A.~Lazovik, and M.~Aiello, ``{Continual Planning with Sensing for
  Web Service Composition},'' in \emph{Proceedings of the 25th AAAI Conference
  on Artificial Intelligence}.\hskip 1em plus 0.5em minus 0.4em\relax AAAI,
  2011, pp. 1198--1203.

\bibitem{martin2007:owl-s}
D.~Martin, M.~Burstein, D.~Mcdermott, S.~Mcilraith, M.~Paolucci, K.~Sycara,
  D.~L. Mcguinness, E.~Sirin, and N.~Srinivasan, ``{Bringing Semantics to Web
  Services with OWL-S},'' \emph{World Wide Web}, vol.~10, no.~3, pp. 243--277,
  2007.

\bibitem{horrocks2003:ontology}
I.~Horrocks, P.~F. Patel-Schneider, and F.~van Harmelen, ``{From \{SHIQ\} and
  \{RDF\} to OWL: the Making of a Web Ontology Language},'' \emph{Web
  Semantics: Science, Services and Agents on the World Wide Web}, vol.~1,
  no.~1, pp. 7--26, 2003.

\bibitem{sohrabi2013:customizing}
S.~Sohrabi, ``{Customizing the Composition of Web Services and Beyond},'' Ph.D.
  dissertation, Depart. of Computer Science, Univ. of Toronto, 2013.

\bibitem{burstein2002:damls}
M.~H. Burstein, J.~R. Hobbs, O.~Lassila, D.~Martin, D.~V. McDermott, S.~A.
  McIlraith, S.~Narayanan, M.~Paolucci, T.~R. Payne, and K.~P. Sycara,
  ``{DAML-S: Web Service Description for the Semantic Web},'' in
  \emph{Proceedings of the First International Semantic Web Conference on The
  Semantic Web}, ser. ISWC '02.\hskip 1em plus 0.5em minus 0.4em\relax
  Springer-Verlag, 2002, pp. 348--363.

\bibitem{madhusudan2006:wsc}
T.~Madhusudan and N.~Uttamsingh, ``{A Declarative Approach to Composing Web
  Services in Dynamic Environments},'' \emph{Decis. Support Syst.}, vol.~41,
  no.~2, pp. 325--357, jan 2006.

\bibitem{kuter2009:wsc}
U.~Kuter and J.~Golbeck, ``{Semantic Web Service Composition in Social
  Environments},'' in \emph{Proceedings of the 8th International Semantic Web
  Conference}, ser. ISWC '09.\hskip 1em plus 0.5em minus 0.4em\relax
  Springer-Verlag, 2009, pp. 344--358.

\bibitem{sohrabi2009:wsc}
S.~Sohrabi and S.~A. Mcilraith, ``{Optimizing Web Service Composition While
  Enforcing Regulations},'' in \emph{Proceedings of the 8th International
  Semantic Web Conference}, ser. ISWC '09.\hskip 1em plus 0.5em minus
  0.4em\relax Springer-Verlag, 2009, pp. 601--617.

\bibitem{sohrabi2010:wsc}
S.~Sohrabi and S.~A. McIlraith, ``{Preference-based Web Service Composition: A
  Middle Ground Between Execution and Search},'' in \emph{Proceedings of the
  9th International Semantic web Conference on The semantic web - Volume Part
  I}, ser. ISWC'10.\hskip 1em plus 0.5em minus 0.4em\relax Springer-Verlag,
  2010, pp. 713--729.

\bibitem{ix}
``{I-X Planning Architecture},'' {O}nline: accessed {J}an. 2014,
  http://www.aiai.ed.ac.uk/project/ix/.

\bibitem{sirin2004:semanticws}
E.~Sirin and B.~Parsia, ``{Planning for Semantic Web Services},'' in
  \emph{Semantic Web Services Workshop at 3rd ISWC}, 2004.

\bibitem{pellet2003}
``{Pellet: OWL DL Reasoner},'' {O}nline: accessed Jan. 2014,
  http://clarkparsia.com/pellet/.

\bibitem{sirin2005:template}
E.~Sirin, B.~Parsia, and J.~Hendler, ``{Template-Based Composition of Semantic
  Web Services},'' in \emph{Proceedings of the AAAI Fall Symposium on Agents
  and the Semantic Web}.\hskip 1em plus 0.5em minus 0.4em\relax AAAI, 2005, pp.
  85--92.

\bibitem{paik2006:framework}
I.~Paik, D.~Maruyama, and M.~N. Huhns, ``{A Framework for Intelligent Web
  Services: Combined HTN and CSP Approach},'' in \emph{Proceedings of the IEEE
  International Conference on Web Services}, ser. ICWS'06.\hskip 1em plus 0.5em
  minus 0.4em\relax IEEE Computer Society, 2006, pp. 959--962.

\bibitem{lin2008:wsc_preferences}
N.~Lin, U.~Kuter, and E.~Sirin, ``{Web service Composition with User
  Preferences},'' in \emph{Proceedings of the 5th European Semantic Web
  Conference on The Semantic Web: Research and Applications}, ser.
  ESWC'08.\hskip 1em plus 0.5em minus 0.4em\relax Springer-Verlag, 2008, pp.
  629--643.

\bibitem{elkawkagy2010:landmarks}
M.~Elkawkagy, B.~Schattenberg, and S.~Biundo, ``{Landmarks in Hierarchical
  Planning},'' in \emph{Proceedings of the 2010 Conference on ECAI 2010: 19th
  European Conference on Artificial Intelligence}.\hskip 1em plus 0.5em minus
  0.4em\relax IOS Press, 2010, pp. 229--234.

\bibitem{elkawkagy2012:improving}
M.~Elkawkagy, P.~Bercher, B.~Schattenberg, and S.~Biundo, ``{Improving
  Hierarchical Planning Performance by the Use of Landmarks},'' in
  \emph{Proceedings of the Twenty-Sixth AAAI Conference on Artificial
  Intelligence}, 2012, pp. 1763--1769.

\bibitem{estlin2001:complementary}
T.~A. Estlin, S.~A. Chien, and X.~Wang, ``{Hierarchical Task Network and
  Operator-Based Planning: Two Complementary Approaches to Real-World
  Planning},'' \emph{Journal of Experimental and Theoretical Artificial
  Intelligence}, vol.~13, no.~4, pp. 379--395, 2001.

\bibitem{bouguerra2004:uncertainty}
A.~Bouguerra and L.~Karlsson, ``{Hierarchical Task Planning under
  Uncertainty},'' in \emph{Proceedings of the 3rd Italian Workshop on Planning
  and Scheduling}, 2004.

\bibitem{kuter2004:nondeterminism}
U.~Kuter and D.~S. Nau, ``{Forward-Chaining Planning in Non-Deterministic
  Domains},'' in \emph{Proceedings of the 19th National Conference on Artifical
  intelligence}, ser. AAAI'04.\hskip 1em plus 0.5em minus 0.4em\relax AAAI,
  2004, pp. 513--518.

\bibitem{brenner2009:continual}
M.~Brenner and B.~Nebel, ``{Continual Planning and Acting in Dynamic Multiagent
  Environments},'' \emph{Autonomous Agents and Multi-Agent Systems}, vol.~19,
  no.~3, pp. 297--331, 2009.

\bibitem{corkill1979:dnoah}
D.~D. Corkill, ``{Hierarchical Planning in a Distributed Environment},'' in
  \emph{Proceedings of the 6th international joint conference on Artificial
  intelligence - Volume 1}, ser. IJCAI'79.\hskip 1em plus 0.5em minus
  0.4em\relax Morgan Kaufmann Publishers Inc., 1979, pp. 168--175.

\bibitem{desJardins1999:dsipe}
M.~desJardins and M.~Wolverton, ``{Coordinating a Distributed Planning
  System},'' \emph{AI Magazine}, vol.~20, no.~4, pp. 45--53, 1999.

\bibitem{dix2003:impacting}
J.~D. Dix, H.~Mu\~{n}oz Avila, D.~S. Nau, and L.~Zhang, ``{IMPACTing SHOP:
  Putting an AI Planner into a Multi-Agent Environment},'' \emph{Annals of
  Mathematics and AI}, vol.~37, no.~4, pp. 381--407, Apr. 2003.

\bibitem{amigoni2005:ambient}
F.~Amigoni, N.~Gatti, C.~Pinciroli, and M.~Roveri, ``{What Planner for Ambient
  Intelligence Applications?}'' \emph{IEEE Transactions on Systems, Man and
  Cybernetics, Part A}, vol.~35, no.~1, pp. 7--21, Jan. 2005.

\bibitem{magnenat2009:planner9}
S.~Magnenat, M.~Voelkle, and F.~Mondada, ``{Planner9, a HTN Planner Distributed
  on Groups of Miniature Mobile Robots},'' in \emph{Proceedings of the 2nd
  International Conference on Intelligent Robotics and Applications}, ser.
  ICIRA'09.\hskip 1em plus 0.5em minus 0.4em\relax Springer-Verlag, 2009, pp.
  1013--1022.

\end{thebibliography}

\end{document}